\documentclass{article}

\PassOptionsToPackage{numbers}{natbib}
\usepackage[eandd, final]{neurips_2026}

\usepackage[utf8]{inputenc}
\usepackage[T1]{fontenc}
\usepackage[colorlinks=true,urlcolor=blue,hypertexnames=false]{hyperref}
\usepackage{booktabs}
\usepackage{array}
\usepackage{amsmath}
\usepackage{amsfonts}
\usepackage{amssymb}
\usepackage{graphicx}
\usepackage{adjustbox}
\usepackage{float}
\usepackage{multirow}
\usepackage{listings}
\usepackage{xcolor}
\usepackage{pifont}
\usepackage{placeins}

\title{fNIRSAtlas: A Large-Scale Benchmark for Functional Near-Infrared Spectroscopy Classification}
\author{%
  Kalle Mäkelä \\
  LUT University, Finland \\
  \texttt{kalle.makela@lut.fi} \\
  \And
  Tuukka Ruotsalo \\
  LUT University, Finland \\
  University of Copenhagen, Denmark \\
  \texttt{tuukka.ruotsalo@lut.fi} \\
}

\begin{document}

\maketitle

\begin{abstract}
\noindent
    Functional near-infrared spectroscopy (fNIRS) is a promising modality for brain-computer interfaces and cognitive state decoding, but progress in machine learning for fNIRS has been limited by small, isolated datasets and inconsistent evaluation protocols.
    We introduce fNIRSAtlas, the largest open benchmark for fNIRS classification to date, comprising 23 classification tasks from 19 public data collections, spanning diverse experimental paradigms, including motor execution, motor imagery, mental arithmetic, working memory, and emotion recognition.
    fNIRSAtlas provides a fully automated pipeline from raw data to evaluation and defines standardized benchmark tasks for both cross-subject and within-subject classification, ensuring reproducible comparison of methods.
    We evaluate a variety of baseline models from fNIRS and related neuroimaging and time-series machine learning literature on the benchmark.
    Our results demonstrate that performance varies substantially across tasks and that methods validated on limited data -- especially deep learning approaches -- often fail to generalize beyond their original evaluation. fNIRSAtlas serves as an extensible, open, and reproducible benchmark to support more robust and cumulative progress in machine learning for fNIRS data.
    Our code and data can be found at \url{https://fnirsatlas.github.io}.
\end{abstract}

\section{Introduction}

Functional near-infrared spectroscopy (fNIRS) is a non-invasive neuroimaging technique
that is gaining popularity as a practical and portable tool for brain decoding
\cite{ferrari2012brief, quaresima2019mini}. Machine learning methods are increasingly
used to model the high-dimensional fNIRS signal \cite{eastmond2022deep, wang2022transformer},
but the field remains methodologically fragmented.
In particular, deep learning for fNIRS classification is still in its infancy, with most studies relying on small, single datasets and inconsistent preprocessing and evaluation settings.
This fragmentation makes it difficult to determine whether reported performance differences reflect genuinely stronger models or just differences in evaluation setup.

\begin{figure*}[t]
\centering
\begin{adjustbox}{center}
\includegraphics[width=1.05\textwidth]{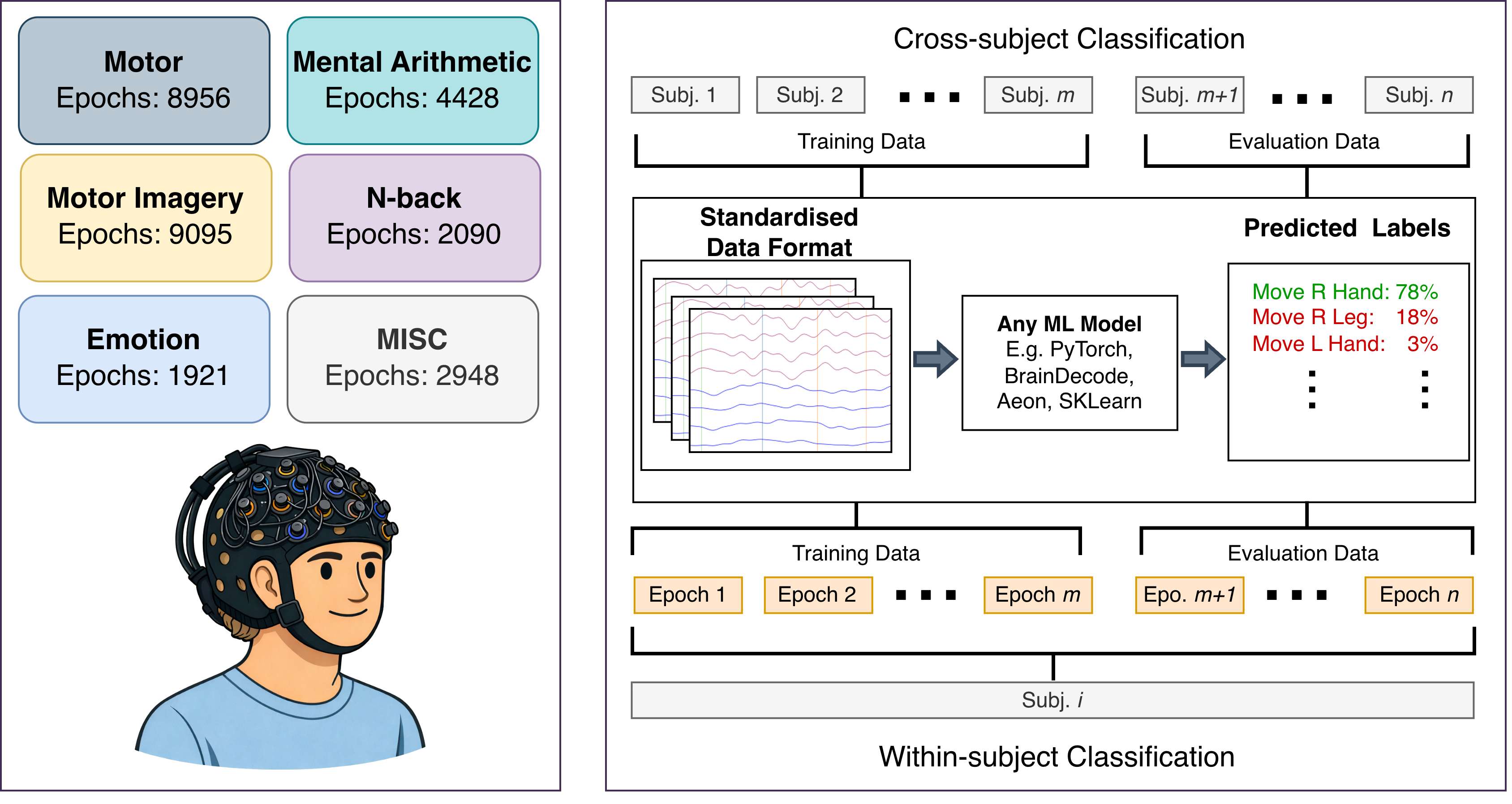}
\end{adjustbox}
\caption{Visual abstract of fNIRSAtlas. The benchmark integrates 23 classification tasks from 19 public data collections across multiple task suites into a standardized pipeline from raw recordings to benchmark tasks and model evaluation, enabling reproducible comparison of machine learning methods in both within-subject and cross-subject settings.}
\label{fig:visual_abstract}
\end{figure*}

Benchmarks and curated datasets have been essential to progress in other
machine learning domains, including computer vision \cite{deng2009imagenet,
lin2014microsoft}, natural language processing \cite{wang2018glue}, and
electroencephalography (EEG) \cite{moabb2018}.
For fNIRS, such benchmarks do not currently exist. While \citet{benerradi2023benchmarking} introduced the first multi-dataset fNIRS benchmark, it remains limited in several respects, most notably by its inclusion of only five tasks, which constrains large-scale and robust comparison across modeling approaches.

We introduce fNIRSAtlas, the largest open fNIRS benchmark to date, which integrates 23 classification tasks from 19 public data collections into a single evaluation system. The benchmark standardizes the entire workflow—from raw data handling and preprocessing to task construction and evaluation protocols—providing a foundation for future research and standard evaluation. By offering an extensible codebase, standardized data formats, and diagnostic tools for task informativeness, fNIRSAtlas facilitates reproducible comparison of both classical and state-of-the-art deep learning methods.

This shared infrastructure lowers the barrier to entry for the broader machine learning community and allows for incremental extension as more public data becomes available. Ultimately, fNIRSAtlas enables a robust assessment of methods across diverse datasets, revealing that approaches appearing strong in single-dataset evaluations often fail to generalize and identifying specific areas where current fNIRS decoding still falls short.

\section{Background and Related Work}

\paragraph*{fNIRS} is a non-invasive optical neuroimaging method that measures 
changes in oxygenated and deoxygenated haemoglobin in cortical tissue using
near-infrared light \cite{ferrari2012brief, scholkmann2014review,
quaresima2019mini}. Light is emitted from sources placed on the scalp and
detected by nearby optodes after passing through superficial tissue and the
outer cortex. Because oxyhaemoglobin and deoxyhaemoglobin absorb
near-infrared light differently, changes in detected light intensity can be
converted into estimates of haemoglobin concentrations using the
modified Beer-Lambert law \cite{mbll}.
Compared with other neuroimaging modalities, fNIRS is relatively portable,
affordable, and tolerant to moderate motion, which has supported its growing
use in applied settings and brain-computer interface research
\cite{quaresima2019mini}. 

\paragraph*{Machine learning for fNIRS classification}  has been widely studied across tasks such as motor imagery, mental workload, and affective state decoding, with the goal
of mapping high-dimensional haemodynamic responses to discrete labels
\cite{naseer2015fnirs, hong2018feature}. Classical pipelines based on
handcrafted features and linear classifiers are commonly used because fNIRS
datasets are often small, noisy, and high-dimensional, which favors simple
models with strong inductive bias \cite{hennrich2015investigating,
bauernfeind2014single, naseer2015fnirs, hong2018feature}.

More recently, deep learning models have become increasingly popular for fNIRS,
most commonly convolutional neural networks (CNNs), recurrent models such as
long short-term memory (LSTM), and, more recently, Transformer-based
architectures \cite{eastmond2022deep, wang2022transformer, huang2021tufts,
dissanayake2025deep, abdollahpour2025advancing}.
However, comparative evidence
remains limited because datasets, preprocessing pipelines, and evaluation
protocols vary substantially across studies \cite{eastmond2022deep,
benerradi2023benchmarking}.

\paragraph*{fNIRS Benchmarks} remain limited, with BenchNIRS by \citet{benerradi2023benchmarking} as the first and only existing
multi-dataset fNIRS machine learning benchmark. Compared to BenchNIRS,
fNIRSAtlas provides substantially broader coverage, increasing the dataset
count from 5 to 23, and offers a fully automated, configurable pipeline from
dataset download to classification, which lowers the barrier to running
standardized evaluations of models, even without fNIRS domain knowledge.
Another central motivation for fNIRSAtlas is to
provide a more informative benchmark for model comparison than BenchNIRS,
where limited task coverage and near-chance performance on many tasks make
rankings unstable and significant model differences difficult to detect. A more
detailed comparison with BenchNIRS, including dataset coverage,
preprocessing, evaluation protocol, and informativeness for model comparison,
is provided in Appendix~\ref{app:benchnirs_comparison}.

\section{The fNIRSAtlas Benchmark}%

\begin{table}[htbp]
\centering
\caption{Overview of the 23 classification tasks included in fNIRSAtlas.
\#Sub: number of subjects;
\#Epo: total number of epochs across all recorded conditions;
\#Ch: number of channels;
\#Cl: number of recorded conditions;
Hz: sampling rate;
Len: epoch length used in the benchmark, excluding the 2 s baseline (s).
Raw: CW: continuous wave;
OD: optical density;
Hb: oxygenated and deoxygenated haemoglobin.
Suites and dataset suffixes: MA/ma: mental arithmetic;
MI/mi: motor imagery;
wg: word generation;
empe: emotional perception;
afim: affective imagery.}
\label{tab:dataset_overview}
\footnotesize
\setlength{\tabcolsep}{4pt}
\begin{tabular}{@{}llrrrrrrlrl@{}}
\toprule
Dataset & Suite & \#Sub & \#Epo & \#Ch & \#Cl & Hz & Len & Raw & Hours & Device \\
\midrule
Buccino2016 \cite{buccino2016hybrid} & Motor & 15 & 1500 & 68 & 4 & 10.4 & 12 & CW & 6.2 & NIRx NIRScout 8-16 \\
Bak2019 \cite{bak2019open} & Motor & 30 & 2250 & 40 & 3 & 13.3 & 15 & Hb & 18.8 & Shimadzu LIGHTNIRS \\
Ortega2020 \cite{ortega2020hygrip} & Motor & 14 & 356 & 48 & 3 & 12.5 & 15 & CW & 5.3 & NIRx NIRScout \\
Luke2021 \cite{luke2021finger} & Motor & 5 & 450 & 40 & 3 & 7.8 & 15 & CW & 3.8 & NIRx NIRScout 16x24 \\
Cockx2023 \cite{cockx2023fnirs} & Motor & 24 & 1844 & 48 & 16 & 50 & 15 & OD & 26.7 & Artinis Brite24 \\
Guevara2024 \cite{guevara2024parkinson} & Motor & 20 & 395 & 44 & 2 & 50 & 15 & CW & 3.8 & Artinis Brite MKII \\
Luke2024b \cite{yucel2024fnirs} & Motor & 10 & 1081 & 128 & 4 & 8.9 & 10 & CW & 6.5 & NIRx NIRScout \\
Gao2024 \cite{gao2023short} & Motor & 12 & 1080 & 200 & 2 & 8.7 & 9 & CW & 4.9 & NIRx NIRSport2 16x16 \\
\midrule
Bnci \cite{bauernfeind2011single} & MA & 8 & 348 & 104 & 2 & 10 & 15 & Hb & 2.2 & Hitachi ETG-4000 \\
Shin2016 (ma) \cite{shin2016open} & MA & 29 & 1740 & 72 & 2 & 10 & 15 & CW & 17.3 & NIRx NIRScout \\
Shin2018b \cite{shin2018eyes} & MA & 12 & 720 & 18 & 2 & 10 & 15 & Hb & 6.5 & NIRx NIRScout \\
Shin2019 (ma) \cite{shin2018ternary} & MA & 18 & 1620 & 32 & 3 & 13.3 & 15 & Hb & 14.1 & Shimadzu LIGHTNIRS \\
\midrule
Shin2016 (mi) \cite{shin2016open} & MI & 29 & 1740 & 72 & 2 & 10 & 15 & CW & 17.4 & NIRx NIRScout \\
Shin2019 (mi) \cite{shin2018ternary} & MI & 18 & 1620 & 32 & 3 & 13.3 & 15 & Hb & 14.1 & Shimadzu LIGHTNIRS \\
Yi2025 \cite{yi2025multi} & MI & 18 & 5735 & 48 & 8 & 7.8 & 15 & CW & 34.3 & NIRx NIRScout \\
\midrule
Herff2014 \cite{herff2013classification} & N-back & 10 & 300 & 16 & 3 & 25 & 40 & Hb & 6.2 & Artinis Oxymon Mark III \\
Shin2018 (N-back) \cite{shin2018simultaneous} & N-back & 26 & 702 & 72 & 3 & 10 & 15 & Hb & 15.0 & NIRx NIRScout \\
Huang2021 \cite{huang2021tufts} & N-back & 68 & 1088 & 4 & 4 & 5.2 & 40 & Hb & 24.8 & ISS Imagent \\
\midrule
Nemo (empe) \cite{spape2023nemo} & Emotion & 31 & 1203 & 48 & 4 & 50 & 15 & OD & 12.0 & Artinis Brite24 \\
Nemo (afim) \cite{spape2023nemo} & Emotion & 31 & 718 & 48 & 4 & 50 & 30 & OD & 11.9 & Artinis Brite24 \\
\midrule
Shin2018 (wg) \cite{shin2018simultaneous} & Word Gen. & 26 & 1560 & 72 & 2 & 10 & 15 & Hb & 14.6 & NIRx NIRScout \\
Shader2021 \cite{shader2021use} & Aud-Vis & 8 & 368 & 104 & 3 & 3.9 & 15 & CW & 4.4 & NIRx NIRScout \\
Luke2024a \cite{yucel2024fnirs} & Auditory & 17 & 1020 & 50 & 3 & 5.2 & 13 & CW & 6.1 & NIRx NIRScoutX \\
\midrule
 &  & 375 & 29438 &  &  &  &  &  & 276.9 &  \\
\bottomrule
\end{tabular}
\end{table}

\paragraph*{fNIRSAtlas} is the largest open fNIRS classification benchmark, unifying 23 classification tasks from 19 publicly available data collections, with 375 unique participants,
across multiple experimental paradigms
(Table~\ref{tab:dataset_overview}). All data undergo a standard processing from raw data to classification, providing reliable and reproducible access to open fNIRS data. The benchmark is distributed as a Python package that makes it easy to run
custom models across the included tasks with flexible configuration options.
It supports both within-subject and cross-subject evaluation under a shared
protocol, enabling direct comparison across common fNIRS settings. This
also makes it easy to study cross-dataset transfer, domain
adaptation, cross-paradigm generalization, and the effect of different
pre-processing choices.

\paragraph*{Datasets} Collecting and standardizing open fNIRS datasets is challenging due to substantial differences in file organization, metadata structure, and event annotations across public releases. We therefore prioritized datasets from high-quality curated sources such as the OpenfNIRS data page\footnote{\url{https://openfnirs.org/data}}, together with datasets already used in comparative studies \cite{wang2022transformer,benerradi2023benchmarking}, and adapted partial open-source parsing implementations where available \cite{moabb2018,mne,mnebids,mnenirs}. To keep the benchmark focused on comparable classification problems, we restricted the scope to stimulus-response tasks with relatively short response windows and sufficiently structured recordings. Accordingly, we excluded long-horizon tasks such as sleep, driving, or gaming, extreme paradigms involving pain or heavy medication, and datasets for which the available files and annotations were too incomplete to support reliable epoch extraction. The current release includes datasets integrated up to the September~2025 implementation cutoff, spans six experimental suites, and provides broad coverage of classification settings in fNIRS while leaving room for future expansion.
Licenses are listed in \autoref{app:licenses}.

\paragraph*{The motor suite} covers overt movement tasks ranging from unilateral finger and
foot
tapping~\cite{bak2019open,luke2021finger,yucel2024fnirs,guevara2024parkinson}
to bimanual grip force control~\cite{ortega2020hygrip}, ball
squeezing~\cite{gao2023short}, and multi-class limb
movements~\cite{buccino2016hybrid,cockx2023fnirs}. These paradigms primarily
engage primary motor cortex and are a standard setting for fNIRS-based decoding
~\cite{naseer2015fnirs,hong2018feature,batula2014evaluating}.

\paragraph*{The mental arithmetic (MA) suite} contains workload discrimination tasks
contrasting arithmetic and baseline (BL)
conditions~\cite{bauernfeind2011single,shin2016open,shin2018eyes,shin2018ternary}.
Mental arithmetic reliably activates prefrontal cortex, which is well suited to
fNIRS measurement, making it a natural paradigm for cognitive-state
decoding~\cite{bauernfeind2011single,naseer2015fnirs,hong2018feature}.

\paragraph*{The motor imagery suite} ranges from binary tasks in Shin2016 and Shin2019 to
the eight-class upper-limb protocol of
Yi2025~\cite{shin2016open,shin2018ternary,yi2025multi}.
Because motor imagery evokes movement-related cortical activity without overt
movement, it is widely used in fNIRS BCIs for communication and motor
rehabilitation applications~\cite{naseer2015fnirs,batula2014evaluating}.

\paragraph*{The n-back suite} covers prefrontal cognitive-load classification using working
memory paradigms at varying load
levels~\cite{herff2013classification,shin2018simultaneous,huang2021tufts}. The
prefrontal cortex shows robust haemodynamic responses to working memory load,
making n-back well suited for passive cognitive-state monitoring and mental
workload BCIs~\cite{herff2014mental,huang2021tufts}.

\paragraph*{The emotion suite} is built from the NEMO dataset and includes two four-class
benchmarks derived from different stimulus subsets: emotional perception (empe) and affective imagery (afim)~\cite{spape2023nemo},
forming the primary affective decoding component of the benchmark. Tasks classify
arousal (Aro; high H / low L) crossed with valence (Val; positive P / negative N).
Affective decoding from fNIRS is motivated by the role of the prefrontal cortex in
emotional processing and its accessibility to fNIRS
measurement~\cite{bendall2016brief}. This suite is currently the smallest but
is expected to expand in the near future.

\paragraph*{The miscellaneous suite} includes word generation (WG), auditory speech perception,
and auditory-visual
conditions~\cite{shin2018simultaneous,shader2021use,yucel2024fnirs}, extending
the benchmark into language and auditory cortex decoding domains. As more
datasets are added, these paradigms will likely split into dedicated suites.

\section{Methods}

\subsection{Preprocessing}
\label{sec:preprocessing}

fNIRS recordings contain various noise sources that must be removed to ensure
reliable classification. We applied a standardized preprocessing pipeline
based on established best practices
\cite{yucel2021best}: (1) optical density (OD) conversion, (2) channel quality
assessment (SCI $< 0.2$) \cite{sci}, (3) motion artifact correction (TDDR) \cite{tddr}, (4)
 haemoglobin conversion (Beer-Lambert \cite{mbll}, partial pathlength factor (PPF) = 6), (5) band-pass filtering
 (0.01--0.1 Hz), (6) segmentation (15 s epochs by default, 2 s baseline, with longer epochs for datasets with longer stimuli and shorter epochs for datasets with event intervals too short to avoid overlap), and (7)
 downsampling to 5 Hz. These settings were used as fixed benchmark defaults to
 maintain comparability across heterogeneous datasets. The starting point depends on raw data format
(\autoref{tab:dataset_overview}): CW datasets (e.g., Buccino2016) begin at step
1, OD datasets (e.g., Cockx2023) at step 2, and HbO/HbR datasets (e.g.,
Bak2019) at step 4. Detailed descriptions are provided in
\autoref{app:preprocessing_details}.

\subsection{Baselines}

We evaluated ten baseline classifiers selected to cover the methods most relevant to fNIRS classification benchmarking, from the most established classical approach to recent fNIRS-specific models, as well as relevant models from EEG and time-series classification.
Shrinkage linear discriminant analysis on window-based summary features (LDA) \cite{hastie2009elements} was included as this classifier is the most commonly used and consistently strong in fNIRS-based BCI studies \cite{blankertz2011single, naseer2015fnirs, hong2018feature, bauernfeind2014single, benerradi2023benchmarking}.
Among earlier deep learning baselines, we included the shallow artificial neural network (ANN) from \citet{naseer2015comparison}, the CNN \cite{lecun2002gradient} baseline of \citet{trakoolwilaiwan2018convolutional}, and the LSTM \cite{hochreiter1997long} baseline of \citet{asgher2020enhanced}, as these or very similar models have continued to appear in later fNIRS comparative studies \cite{li2025metanirs, wang2022transformer, wang2023rethinking, pandey2024fnirsnet}.
We also considered the EEG-derived models EEGNet \cite{lawhern2018eegnet} and DeepConvNet \cite{dcn}, since EEG is a close relative of fNIRS in non-invasive brain decoding and both architectures have been adopted in fNIRS research \cite{kwon2021subject, huang2021tufts}.
The more recent fNIRS-specific models were fNIRS-T \cite{wang2022transformer}, the first Transformer \cite{vaswani2017attention} model in fNIRS research \cite{abdollahpour2025advancing}, fNIRSNet23 \cite{wang2023rethinking}, a compact CNN designed for delayed haemodynamic dynamics, and fNIRSNet24 \cite{pandey2024fnirsnet}, a more complex multi-view CNN, all of which reported strong comparative results in their original studies.
Finally, we included MultiRocket+Hydra \cite{dempster2023hydra}, a computationally efficient state-of-the-art time-series classifier that combines MultiRocket \cite{tan2022multirocket} and Hydra \cite{middlehurst2024bake}; earlier ROCKET-family methods \cite{dempster2019rocket} have also shown promise for fNIRS classification \cite{andreu2021single}.

Deep learning models and MRH were trained directly on the preprocessed haemoglobin
signal, whereas LDA and ANN used window-based summary features. The signal was
z-normalized separately for HbO and HbR across samples, spatial channels, and
time using statistics from the training split only, preserving relative
amplitude differences between channels. Models were trained for 100
epochs using the Adam optimizer \cite{adam}. The starting learning rate was
selected by grid search and decayed by a factor of 0.5 every 20 epochs. Batch
size was 32 for cross-subject experiments and 8 for within-subject experiments.
The loss function was categorical cross-entropy on softmax outputs. More detailed
baseline descriptions and model-specific hyperparameter search spaces are
reported in \autoref{app:baseline-details}.

\subsection{Classification tasks}

Classification tasks were defined for each dataset,
as detailed in \autoref{tab:task-informativeness}.
Most tasks classify all recorded conditions against each other,
but some use a subset or merge conditions
when this gives a more sensible classification problem.
For example, the Luke2021 task omits the rest state
and classifies left- versus right-hand finger tapping,
and the Cockx2023 task merges 16 recorded conditions,
half of them foot-tapping variants and half right-hand finger-tapping variants,
into two classes.

\subsection{Model evaluation}
Two evaluation strategies were implemented to assess model performance: cross-subject classification, which measures a model’s ability to generalize across the high physiological variability of different individuals, and within-subject classification, which determines the upper-bound decoding performance for a specific subject's neural patterns.

\paragraph*{Cross-subject classification}
evaluates model generalization to unseen subjects by training on one set of
subjects and testing on different held-out subjects. We used nested cross-validation (CV), with the outer loop for
performance evaluation and the inner loop for hyperparameter optimization. The
outer loop used grouped 5-fold CV, and hyperparameters were selected in an
inner CV loop on the training portion of each outer split using 4 folds.

\paragraph*{Within-subject classification}
evaluates model performance on each subject separately, using that
subject's data for training and testing.
Because fully nested CV was prohibitively expensive and noisy due to small sample sizes,
we instead ran a single
stratified 5-fold CV procedure per subject
(\autoref{app:blocked-cv} examines contiguous instead of shuffled folds)
for each candidate hyperparameter setting and
stored the resulting subject-level performances.
Using a leave-one-subject-out scheme, the setting with the highest
mean accuracy across the remaining subjects was selected, and the held-out
subject's performance was read directly from the already-computed scores.
This reflects a practical setting where a previously selected configuration is applied to a new subject.

To reduce variance from random partitions, all CV procedures were repeated 5
times with independently randomized splits.
\autoref{app:repeats} examines how the number of repetitions affects model comparison.
We report classification accuracy
as the primary metric. Accuracy is appropriate here because all benchmark tasks
are class-balanced.

\subsection{Statistical testing}

For multi-dataset model comparison, the methodology recommended by
\citet{demsar2006statistical} was followed for comparing multiple classifiers across
multiple datasets. The Friedman test 
\cite{friedman1937use} was used to assess whether overall performance differences among models were significant across tasks. Each non-overlapping task contributed one
mean per model to ensure statistical
independence. Specifically, Shin2019 (ma) and Shin2019 (mi) share a common
rest class, so the mental imagery partition from the multi-task
 statistical analysis was excluded. When the omnibus test revealed significant differences, the Nemenyi post-hoc tests for pairwise comparisons were conducted \cite{nemenyi1963distribution}.
Some data collections contribute multiple tasks, such as Shin2016 with separate mental arithmetic and motor imagery recordings. Although the data of these tasks do not overlap, to ensure that these collections do not bias the results, \autoref{app:collection} repeats the tests after averaging the tasks of each data collection.

\subsection{Task Informativeness}

To ensure the benchmark serves as an effective instrument for model comparison, we also include heuristic post-hoc analyses of task informativeness. In this context, we define informativeness primarily as a task's discriminative power: its ability to provide a reliable basis for distinguishing between different model architectures, estimated relative to the evaluated baselines rather than as an intrinsic property of the task. These analyses support the prioritization of tasks and the curation of compact subsets that maintain high evaluative power. Furthermore, these can help to diagnose whether the observed performance gains reflect genuine model advantages rather than differences in experimental or task derived noise.

We summarize task informativeness along three dimensions: stability of model rankings, signal decodability, and separation between model performances. All three diagnostics are derived from cross-subject Cohen's $\kappa$ \cite{cohen1960coefficient} scores, defined as
$\kappa = \frac{p_o - p_e}{1 - p_e}$
where $p_o$ is the observed agreement (accuracy), and $p_e$ is the expected agreement under independence of the true and predicted labels (chance-level).
We use $\kappa$ because it expresses performance relative to the chance
level of each task, making scores more comparable across tasks with different
numbers of classes. %

\paragraph*{Rank stability (RS)} measures the robustness of model rankings under bootstrap resampling. First, models are ranked by mean $\kappa$ across folds. We then draw $n=9999$ paired bootstrap resamples over folds, rerank the models in each resample, and average the Spearman correlations with the full-sample ranking.
\paragraph*{Top-3 kappa ($\kappa\text{@}3$)} is the mean Cohen's $\kappa$ across the three best-performing models on that task, measuring whether the task provides usable decoding signal.
\paragraph*{Model separation ratio (MSR)} is an SNR-like measure of model separation relative to within-model variation. It is computed as the standard deviation of model-wise mean $\kappa$ scores divided by the standard deviation of fold-level $\kappa$ scores after subtracting the corresponding model mean and fold mean from each score.

\textbf{Summary metric} $U_d$ is then defined as the equally-weighted
average of min-max normalized RS, $\kappa\text{@}3$, MSR. Formula definitions for all metrics are given in \autoref{app:task-informativeness-metrics}.

These diagnostics are used for two purposes. First, to define reduced benchmark subsets that allow faster model comparison and assessment of the reliability of potential new tasks to be used within the benchmark. Second, the diagnostics allow to identify importance of observed model differences given the informativeness of the tasks. To this end, we consider two complementary
selection principles: a suite-balanced informative subset containing
high-scoring tasks across task suites, and a globally informative subset
containing the most informative tasks overall. The resulting task sets are
reported in \autoref{subsec:task-informativeness-results}.

\subsection{Computational environment}

Processing of the fNIRS data was conducted using the \texttt{MNE-Python} library (version 1.11.0) \citep{mne} and \texttt{MNE-BIDS} (version 0.18.0) \citep{mnebids}. Most machine learning models were trained and evaluated on CPU, whereas GPU resources were used for some large cross-subject \mbox{fNIRS-T} and \mbox{fNIRSNet24} experiments. LDA and ANN were implemented using the \texttt{scikit-learn} library (version 1.7.2) \citep{sklearn}. Deep learning models were implemented in \texttt{PyTorch} (version 2.10.0) \citep{ansel2024pytorch}. Additional hardware specifications and aggregate compute use are reported in \autoref{app:compute}.

All reported results use fNIRSAtlas version 0.1.0. The benchmark will continue
to evolve as new datasets and models are added in later versions, but this versioned
snapshot fixes the experiment configuration needed to fully reproduce the
reported results.

\section{Results}

\subsection{Classification Performance}

\begin{table}[htbp]
\centering
\caption{Cross-subject classification accuracy for the baseline models.
Values show mean test accuracy $\pm$ standard deviation of fold-level accuracies, describing cross-validation variability.
Bold indicates the highest value in each row.
\#Cl: number of classes.}
\label{tab:cross-subject-results}
\scriptsize
\setlength{\tabcolsep}{1.5pt}
\begin{tabular*}{\textwidth}{@{}l@{\hspace{3pt}}c@{\extracolsep{\fill}}*{10}{r}@{}}
\toprule
    Dataset & \#Cl & \shortstack{LDA \\ \cite{cohen2013applied}} & \shortstack{ANN \\ \cite{naseer2015comparison}} & \shortstack{CNN \\ \cite{trakoolwilaiwan2018convolutional}} & \shortstack{LSTM \\ \cite{asgher2020enhanced}} & \shortstack{EEGNet \\ \cite{lawhern2018eegnet}} & \shortstack{DCN \\ \cite{dcn}} & \shortstack{MRH \\ \cite{dempster2023hydra}} & \shortstack{fNIRS-T \\ \cite{wang2022transformer}} & \shortstack{fNIRSNet23 \\ \cite{wang2023rethinking}} & \shortstack{fNIRSNet24 \\ \cite{pandey2024fnirsnet}} \\
\midrule
Buccino2016 & 4 & 45.2±4.8 & 41.3±3.7 & 41.6±4.2 & 41.4±4.3 & 43.8±5.4 & 44.1±4.8 & 41.3±4.4 & 42.7±3.9 & \textbf{48.5±6.0} & 38.7±4.0 \\
Bak2019 & 2 & 82.3±4.4 & 79.4±4.9 & 78.4±4.7 & 77.1±4.6 & 81.9±4.7 & 82.1±4.2 & 78.6±3.9 & 80.5±4.3 & \textbf{83.0±4.3} & 79.1±4.6 \\
Ortega2020 & 2 & 63.3±7.4 & 61.2±5.3 & 60.0±8.8 & 57.1±7.9 & 65.4±7.5 & 64.5±8.3 & 56.8±5.5 & 56.0±6.8 & \textbf{65.4±7.3} & 49.9±5.7 \\
Luke2021 & 2 & 82.3±9.2 & 81.9±9.9 & 80.5±9.8 & 79.3±10.2 & 81.0±9.0 & 79.9±8.5 & 76.1±9.3 & 79.3±11.7 & \textbf{84.2±8.9} & 75.2±11.4 \\
Cockx2023 & 2 & 84.7±4.9 & 84.2±4.4 & 82.4±5.9 & 81.7±3.2 & 84.4±5.7 & 83.8±5.4 & 84.9±5.7 & \textbf{85.1±4.9} & 83.8±4.6 & 79.8±3.1 \\
Guevara2024 & 2 & \textbf{67.6±5.9} & 65.9±5.2 & 64.9±6.1 & 64.3±6.4 & 65.8±6.1 & 66.1±5.4 & 63.5±4.3 & 62.2±4.8 & 66.3±5.8 & 60.0±5.5 \\
Luke2024b & 2 & 53.9±2.9 & 52.2±3.1 & \textbf{54.2±1.7} & 51.2±2.8 & 52.9±4.7 & 51.5±3.6 & 50.2±3.1 & 52.1±3.0 & 53.1±3.3 & 52.0±4.0 \\
Gao2024 & 2 & 67.9±9.2 & 69.1±9.0 & 68.7±8.5 & 64.2±6.9 & 69.7±9.4 & 69.0±7.4 & 65.2±7.6 & 66.7±7.2 & \textbf{71.3±8.5} & 68.6±9.1 \\
Bnci & 2 & 80.0±11.5 & 86.5±9.2 & \textbf{87.0±6.6} & 84.9±10.8 & 82.3±11.0 & 84.3±10.6 & 78.3±10.1 & 85.5±9.0 & 84.7±9.3 & 83.0±9.1 \\
Shin2016 (ma) & 2 & 68.4±3.2 & 67.6±2.6 & 63.4±3.8 & 62.7±3.9 & 68.4±2.6 & 66.3±3.2 & 68.2±3.4 & 66.6±2.7 & \textbf{69.7±3.6} & 63.3±3.0 \\
Shin2018b & 2 & 66.4±7.2 & 66.7±6.2 & 62.8±5.3 & 63.8±4.6 & 68.3±4.8 & 67.4±4.2 & 66.4±6.6 & 68.4±6.5 & \textbf{70.2±5.9} & 64.1±5.1 \\
Shin2019 (ma) & 2 & \textbf{68.3±5.0} & 65.1±4.4 & 61.4±5.5 & 60.3±2.9 & 65.1±4.6 & 64.1±5.5 & 66.0±5.9 & 64.8±5.2 & 67.9±4.9 & 61.5±4.6 \\
Shin2016 (mi) & 2 & 55.0±2.7 & 53.5±2.2 & 52.0±2.1 & 49.9±2.4 & \textbf{55.7±2.7} & 54.6±2.2 & 52.4±2.0 & 51.9±2.1 & 54.1±3.0 & 49.9±2.8 \\
Shin2019 (mi) & 2 & 54.4±4.9 & 52.8±4.8 & 52.6±4.1 & 52.9±4.6 & 54.3±4.9 & 54.5±4.4 & \textbf{55.2±5.2} & 53.9±4.4 & 55.1±4.8 & 52.0±3.7 \\
Yi2025 & 3 & \textbf{35.3±1.4} & 33.9±1.4 & 34.0±1.6 & 33.5±1.9 & 34.3±2.0 & 34.7±1.7 & 34.3±1.3 & 34.0±2.0 & 34.5±1.6 & 33.4±1.3 \\
Herff2014 & 3 & 35.5±5.2 & 34.0±5.4 & 34.9±6.7 & 36.7±6.5 & 37.2±5.2 & 35.8±4.1 & \textbf{47.5±3.6} & 36.3±5.6 & 36.4±5.1 & 35.2±6.8 \\
Shin2018 (nback) & 3 & 47.9±4.4 & 46.7±3.4 & 45.0±4.0 & 37.8±4.3 & 46.5±4.8 & 47.9±4.7 & 48.2±4.6 & 42.2±4.8 & \textbf{49.0±4.7} & 36.4±3.4 \\
Huang2021 & 4 & \textbf{34.5±2.6} & 30.1±2.6 & 30.3±3.0 & 25.6±2.8 & 33.7±2.6 & 33.3±2.4 & 32.4±3.1 & 30.3±2.4 & 34.2±2.6 & 29.2±2.6 \\
Nemo (empe) & 4 & 26.4±2.3 & 25.3±2.9 & 26.5±3.0 & 27.1±2.6 & 26.4±3.7 & 27.0±2.0 & 26.4±2.5 & \textbf{27.4±2.3} & 26.3±2.7 & 25.8±2.8 \\
Nemo (afim) & 4 & 26.9±3.3 & 26.3±2.4 & 26.1±3.0 & 24.1±2.4 & 25.0±3.2 & 25.4±2.6 & 24.3±3.1 & 26.3±3.9 & \textbf{27.2±3.4} & 25.2±3.1 \\
Shin2018 (wg) & 2 & 64.7±2.9 & 63.2±2.5 & 59.2±4.6 & 60.3±3.2 & 64.5±2.8 & 62.9±3.0 & 65.1±2.9 & 62.0±3.5 & \textbf{67.0±2.8} & 59.6±3.3 \\
Shader2021 & 3 & 48.9±7.7 & 46.7±8.4 & 47.9±6.0 & 48.9±7.6 & 50.2±8.9 & 48.4±9.2 & 47.8±9.1 & 51.6±5.4 & \textbf{51.7±7.3} & 45.8±5.2 \\
Luke2024a & 3 & \textbf{47.0±4.3} & 44.8±4.9 & 41.0±4.7 & 37.1±4.0 & 43.6±5.1 & 44.6±4.1 & 43.3±4.8 & 43.9±4.0 & 45.7±5.0 & 37.5±4.2 \\
\midrule
Mean &  & 56.8 & 55.6 & 54.6 & 53.1 & 56.5 & 56.2 & 55.3 & 55.2 & \textbf{57.8} & 52.4 \\
\bottomrule
\end{tabular*}

\end{table}

\begin{figure}[htbp]
  \centering
    \includegraphics[width=0.99\textwidth]{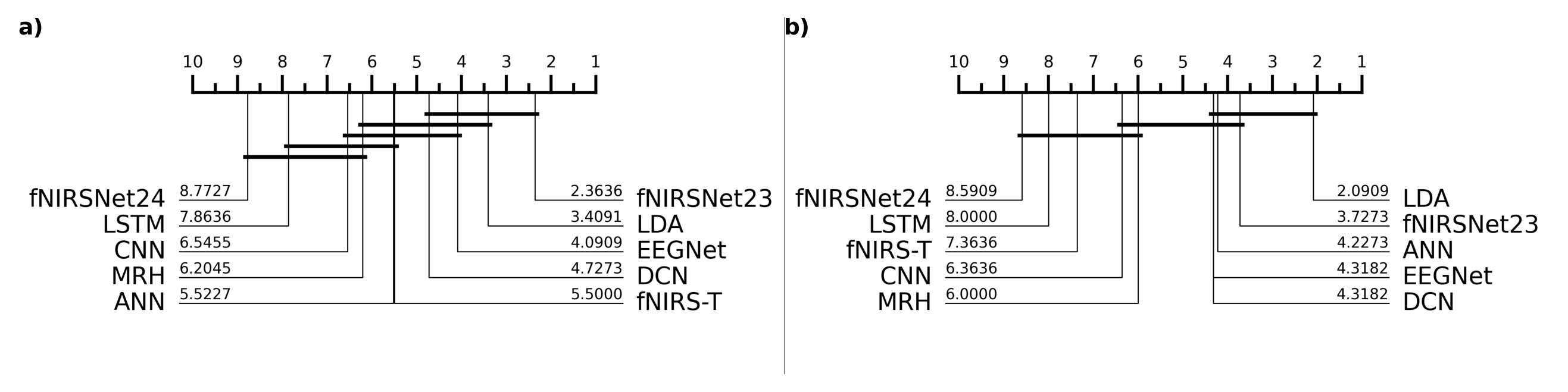}
\caption{
Critical difference diagrams \cite{demsar2006statistical} for cross-subject (a) and within-subject (b) accuracy ranks. Models are ordered by average rank across tasks, with lower rank indicating better performance. Horizontal bars connect models that are not significantly different under the Nemenyi post-hoc test ($\alpha = 0.05$).
    }
  \label{fig:cd-diagram}
\end{figure}

\autoref{tab:cross-subject-results} presents cross-subject classification
accuracy for all tasks, with fNIRSNet23 giving the strongest
overall results. The Friedman test detected a significant difference among
models ($\chi^2=83.252$, $p<0.001$; 22 tasks), and the Nemenyi post-hoc test
showed that LDA, EEGNet, and DCN were not significantly different from
fNIRSNet23, as visualized in \autoref{fig:cd-diagram}. In contrast, the more
complex models performed markedly worse: fNIRS-T ranked below the strongest
baselines, while LSTM and fNIRSNet24 were among the weakest methods overall.
Taken together, the results over the fNIRSAtlas show that in this benchmark many recent models that are based on more complex architectures were not showing performance improvements.
\autoref{app:learning-curve} examines the cross-subject performance of all models with an increasing number of training subjects.
\autoref{app:benchmark-size} examines the model ranking and significant model differences with an increasing number of tasks.

\begin{table}[htbp]
\centering
\caption{Within-subject classification accuracy for the baseline models.
Values show mean test accuracy $\pm$ standard deviation across subjects.
Bold indicates the highest value in each row.
Abbreviations as in \autoref{tab:cross-subject-results}.}
\label{tab:within-subject-results}
\scriptsize
\setlength{\tabcolsep}{1.5pt}
\begin{tabular*}{\textwidth}{@{}l@{\hspace{3pt}}c@{\extracolsep{\fill}}*{10}{r}@{}}
\toprule
        Dataset & \#Cl & \shortstack{LDA \\ \cite{cohen2013applied}} & \shortstack{ANN \\ \cite{naseer2015comparison}} & \shortstack{CNN \\ \cite{trakoolwilaiwan2018convolutional}} & \shortstack{LSTM \\ \cite{asgher2020enhanced}} & \shortstack{EEGNet \\ \cite{lawhern2018eegnet}} & \shortstack{DCN \\ \cite{dcn}} & \shortstack{MRH \\ \cite{dempster2023hydra}} & \shortstack{fNIRS-T \\ \cite{wang2022transformer}} & \shortstack{fNIRSNet23 \\ \cite{wang2023rethinking}} & \shortstack{fNIRSNet24 \\ \cite{pandey2024fnirsnet}} \\
\midrule
Buccino2016 & 4 & \textbf{71.8±12.7} & 64.6±10.3 & 57.3±10.2 & 50.8±10.1 & 63.2±11.6 & 64.2±11.4 & 61.2±12.2 & 59.6±10.1 & 65.9±12.8 & 49.6±9.5 \\
Bak2019 & 2 & \textbf{83.5±15.4} & 79.7±16.6 & 78.1±16.8 & 76.0±17.3 & 82.6±15.6 & 82.4±16.5 & 72.8±14.9 & 75.7±15.9 & 83.3±14.4 & 67.5±17.6 \\
Ortega2020 & 2 & 62.6±16.5 & 56.0±16.3 & 61.1±16.2 & 59.9±15.0 & \textbf{70.5±17.3} & 68.3±15.7 & 54.8±15.5 & 56.9±13.9 & 63.2±16.3 & 59.1±14.1 \\
Luke2021 & 2 & \textbf{90.3±14.0} & 78.0±15.0 & 80.8±18.5 & 67.0±17.7 & 83.8±13.8 & 85.4±14.5 & 77.1±15.8 & 78.1±17.1 & 87.7±14.8 & 80.1±14.5 \\
Cockx2023 & 2 & 95.9±5.2 & 93.5±5.1 & 90.1±6.2 & 86.6±6.2 & 93.2±5.1 & 93.2±5.8 & \textbf{96.5±3.1} & 91.8±5.9 & 95.4±5.2 & 87.2±7.2 \\
Guevara2024 & 2 & 59.5±13.3 & \textbf{64.6±15.4} & 56.9±14.8 & 58.3±14.2 & 64.3±16.5 & 61.5±16.5 & 56.3±13.0 & 54.3±14.9 & 61.5±15.0 & 55.4±11.7 \\
Luke2024b & 2 & 72.4±10.3 & 61.5±10.5 & 60.8±11.9 & 57.4±10.4 & 58.7±8.1 & 62.3±11.0 & \textbf{72.6±9.4} & 61.8±10.6 & 62.0±8.8 & 67.6±11.4 \\
Gao2024 & 2 & \textbf{77.7±13.2} & 70.6±12.5 & 66.7±12.7 & 62.3±11.0 & 69.0±13.8 & 71.0±14.0 & 65.4±10.8 & 66.5±15.0 & 73.9±13.0 & 64.9±12.4 \\
Bnci & 2 & \textbf{98.6±2.0} & 95.2±4.8 & 94.6±5.1 & 93.3±6.4 & 96.1±3.1 & 95.0±4.1 & 87.9±10.9 & 90.4±5.3 & 97.8±2.8 & 76.9±12.9 \\
Shin2016 (ma) & 2 & \textbf{84.2±7.8} & 83.3±7.6 & 78.4±7.9 & 76.3±8.0 & 82.3±8.0 & 82.5±7.7 & 75.7±9.1 & 76.7±8.6 & 83.2±7.8 & 67.5±9.6 \\
Shin2018b & 2 & \textbf{80.1±10.9} & 77.2±8.1 & 74.6±8.5 & 72.3±8.4 & 77.4±10.3 & 77.9±9.7 & 74.2±10.9 & 73.6±8.4 & 77.2±10.3 & 68.2±9.2 \\
Shin2019 (ma) & 2 & \textbf{78.5±9.5} & 77.2±8.1 & 74.4±8.9 & 72.4±8.5 & 78.0±10.3 & 76.9±9.5 & 74.6±10.2 & 72.0±9.0 & 76.4±10.4 & 62.9±10.1 \\
Shin2016 (mi) & 2 & \textbf{59.0±8.7} & 58.5±9.2 & 56.0±9.1 & 54.0±7.7 & 58.3±9.0 & 56.6±9.3 & 52.2±8.1 & 53.4±7.8 & 58.2±8.8 & 50.6±6.7 \\
Shin2019 (mi) & 2 & \textbf{68.7±8.4} & 67.7±9.2 & 63.7±9.6 & 60.8±8.1 & 66.1±9.1 & 66.3±9.1 & 65.4±9.0 & 62.6±8.0 & 67.4±8.5 & 57.8±7.9 \\
Yi2025 & 3 & 39.5±5.1 & \textbf{39.6±5.1} & 36.3±4.5 & 34.5±3.6 & 37.6±5.5 & 38.8±5.4 & 36.8±4.9 & 35.7±4.1 & 39.1±5.1 & 35.1±4.0 \\
Herff2014 & 3 & \textbf{48.1±9.8} & 44.6±9.8 & 41.1±10.2 & 37.8±9.6 & 45.2±9.9 & 44.5±10.5 & 45.1±7.4 & 42.7±10.0 & 41.7±8.6 & 39.5±8.6 \\
Shin2018 (nback) & 3 & \textbf{49.4±12.1} & 45.3±10.3 & 40.9±12.1 & 38.1±10.4 & 40.7±11.1 & 42.3±11.6 & 42.9±12.3 & 35.7±10.5 & 42.6±10.3 & 35.9±12.5 \\
Huang2021 & 4 & 25.9±12.7 & 28.2±14.2 & 27.0±13.1 & 25.8±10.6 & 26.2±13.5 & 24.2±10.6 & \textbf{29.4±13.4} & 27.6±12.5 & 27.7±12.0 & 26.5±12.2 \\
Nemo (empe) & 4 & \textbf{35.1±10.3} & 32.9±9.1 & 28.5±8.5 & 30.5±10.2 & 34.5±11.3 & 33.6±9.3 & 29.7±8.4 & 30.2±8.5 & 32.5±9.0 & 26.3±7.4 \\
Nemo (afim) & 4 & 22.0±10.0 & 21.2±8.8 & 24.4±10.9 & \textbf{25.0±10.3} & 21.8±8.1 & 22.9±8.7 & 23.0±11.1 & 23.8±9.2 & 22.1±8.2 & 23.9±11.6 \\
Shin2018 (wg) & 2 & \textbf{75.0±10.8} & 72.1±9.6 & 68.7±10.6 & 64.4±9.9 & 71.3±11.6 & 70.6±11.2 & 67.7±10.3 & 66.1±10.6 & 72.2±11.8 & 57.2±8.3 \\
Shader2021 & 3 & \textbf{62.8±14.6} & 59.5±13.1 & 55.4±14.6 & 53.1±14.4 & 55.4±17.2 & 57.4±16.5 & 55.5±11.0 & 52.8±13.9 & 52.9±14.2 & 46.6±11.7 \\
Luke2024a & 3 & \textbf{52.2±15.9} & 44.7±9.7 & 45.2±10.3 & 39.6±9.2 & 46.7±12.4 & 46.5±12.3 & 47.6±14.2 & 45.5±11.8 & 47.8±12.5 & 38.4±10.0 \\
\midrule
Mean &  & \textbf{64.9} & 61.5 & 59.2 & 56.3 & 61.9 & 61.9 & 59.3 & 58.0 & 62.2 & 54.1 \\
\bottomrule
\end{tabular*}

\end{table}

\autoref{tab:within-subject-results} presents within-subject classification
accuracy for all tasks, with LDA giving the strongest overall
results. As in the cross-subject setting, the same five models occupy the top
half of the comparison: LDA, fNIRSNet23, ANN, EEGNet, and DCN. The Friedman
test again detected a significant difference among models ($\chi^2=94.681$,
$p<0.001$; 22 tasks), and the Nemenyi post-hoc test showed that none of these
four alternatives differed significantly from LDA, as visualized in
\autoref{fig:cd-diagram}. The separation between the stronger and weaker
methods was clearer in the within-subject setting, with MultiRocketHydra, CNN,
fNIRS-T, LSTM, and fNIRSNet24 falling well behind the leading baselines.
Overall, the within-subject results on the fNIRSAtlas benchmark also show that higher-capacity architectures do not outperform classical linear models.
Additional classical baselines evaluated under the same protocol are reported in \autoref{app:classical}.%

\subsection{Task Informativeness}
\label{subsec:task-informativeness-results}

\begin{table}[htbp]
\centering
\caption{Task informativeness diagnostics and reduced-partition membership.
RS is rank stability,
$\kappa\text{@}3$ is the mean Cohen's $\kappa$ across the three best-performing models,
MSR is the model separation ratio,
and $U_d$ is their normalized average.
Checkmarks indicate inclusion in fNIRSAtlas-Balanced (Bal) and fNIRSAtlas-HighInfo (HI).
Tasks: BL: baseline;
MA: mental arithmetic;
WG: word generation;
MI: motor imagery;
LMI/RMI: left/right motor imagery;
L/R: left/right;
Aro: arousal (H: high, L: low);
Val: valence (P: positive, N: negative).}
    \label{tab:task-informativeness}
\small
\begin{tabular}{llrrrrrr}
\toprule
Dataset & Task & RS\% & $\kappa\text{@}3$\% & MSR & U$_d$\% & Bal & HI \\
\midrule
Bak2019 & FingerTap[L|R] & 97.3 & 64.9 & 1.23 & 93.5 & \ding{51} & \ding{51} \\
Cockx2023 & Tapping[Finger|Foot] & 90.2 & 69.8 & 0.66 & 67.8 & \ding{51} & \ding{51} \\
Ortega2020 & HandGrip[L|R] & 95.7 & 30.2 & 0.96 & 66.1 & \ding{51} & \ding{51} \\
Gao2024 & HandSqueeze[L|R] & 93.1 & 40.1 & 0.73 & 60.1 & \ding{51} & \ding{51} \\
Luke2021 & FingerTap[L|R] & 88.8 & 65.6 & 0.49 & 58.7 & \ding{51} & \ding{51} \\
Buccino2016 & Side[R|L]\texttimes{}Joint[Arm|Hand] & 93.7 & 27.9 & 0.83 & 58.4 & \ding{51} & \ding{51} \\
Guevara2024 & FingerTap[L|R] & 90.1 & 33.1 & 0.56 & 47.1 & \ding{51} & \ding{51} \\
Luke2024b & FingerTap[L|R] & 87.1 & 5.3 & 0.38 & 24.1 & \ding{55} & \ding{55} \\
\midrule
Shin2016 (ma) & BL|MA & 95.1 & 37.7 & 1.00 & 70.2 & \ding{51} & \ding{51} \\
Shin2019 (ma) & BL|MA & 95.2 & 34.8 & 1.00 & 68.9 & \ding{51} & \ding{51} \\
Bnci & BL|MA & 90.8 & 72.7 & 0.49 & 65.0 & \ding{51} & \ding{51} \\
Shin2018b & BL|MA & 91.6 & 37.9 & 0.57 & 52.0 & \ding{51} & \ding{51} \\
\midrule
Shin2016 (mi) & LMI|RMI & 96.6 & 10.2 & 0.93 & 57.2 & \ding{51} & \ding{51} \\
Shin2019 (mi) & MI|Rest & 85.9 & 9.9 & 0.37 & 24.5 & \ding{51} & \ding{55} \\
Yi2025 & Limb[Shoulder|Elbow|Wrist] & 86.4 & 2.2 & 0.40 & 22.4 & \ding{51} & \ding{55} \\
\midrule
Shin2018 (nback) & 0-back|2-back|3-back & 94.9 & 22.6 & 1.32 & 72.9 & \ding{51} & \ding{51} \\
Huang2021 & 0-back|1-back|2-back|3-back & 95.9 & 12.2 & 1.14 & 63.5 & \ding{51} & \ding{51} \\
Herff2014 & 1-back|2-back|3-back & 76.4 & 10.7 & 0.79 & 24.6 & \ding{55} & \ding{55} \\
\midrule
Nemo (afim) & Aro[H|L]\texttimes{}Val[P|N] & 86.4 & 2.4 & 0.36 & 21.4 & \ding{51} & \ding{55} \\
Nemo (empe) & Aro[H|L]\texttimes{}Val[P|N] & 73.6 & 2.9 & 0.26 & 0.3 & \ding{55} & \ding{55} \\
\midrule
Shin2018 (wg) & BL|WG & 97.2 & 31.2 & 1.09 & 72.9 & \ding{51} & \ding{51} \\
Luke2024a & Control|Noise|Speech & 95.4 & 18.7 & 1.14 & 66.1 & \ding{51} & \ding{51} \\
Shader2021 & Audio|Control|Video & 82.4 & 23.7 & 0.35 & 25.4 & \ding{51} & \ding{55} \\
\bottomrule
\end{tabular}
\end{table}

\autoref{tab:task-informativeness} shows that task informativeness varied substantially across the benchmark, reflecting differences in how strongly tasks supported model comparison. Motor execution, mental arithmetic, and word generation contained many of the most informative tasks, combining usable decoding signal with stable model rankings. Lower informativeness in motor imagery and emotion recognition was mainly driven by high task difficulty, which left little separation between models. Within otherwise stronger task groups, Luke2024b and Herff2014 were clear low-informativeness outliers; Luke2024b had much lower $\kappa\text{@}3$ than the other motor execution tasks, while Herff2014 combined weak decoding signal with a small sample size. Overall, these analyses identify which tasks contributed most to model comparison under the current configuration. Low-informativeness tasks remain valuable as difficult decoding settings, but were less useful for distinguishing between the tested models.

We define two optional reduced benchmark partitions from the task informativeness scores: fNIRSAtlas-Balanced (Bal) and fNIRSAtlas-HighInfo (HI).
Inclusion indicators for the tasks are shown in \autoref{tab:task-informativeness}.
Balanced excludes tasks with $U_d$ below half of the median $U_d$ within their task suite, preserving task diversity while removing clear low-informativeness outliers. All miscellaneous tasks are included in Balanced, since they do not belong to a larger task suite for within-suite thresholding. HighInfo excludes tasks with $U_d$ below half of the benchmark-wide median $U_d$, providing a benchmark-wide higher-informativeness partition with stronger model-comparison signal. These partitions provide higher-informativeness alternatives for focused model comparison, while the full benchmark and complete task suites remain useful for broader evaluation.

\section{Discussion}
We present fNIRSAtlas, a benchmark for systematically comparing fNIRS classification models and enabling more robust evaluation of consistency and generalization across studies. fNIRSAtlas provides a practical basis for more reproducible evaluation and cumulative progress in fNIRS machine learning by bringing datasets, tasks, and methods into a common benchmarking framework. We release all code and processed data to ensure full reproducibility, and we hope that fNIRSAtlas can serve as a standard and continuously maintained evaluation benchmark for the community.

Under the standardized evaluation used in fNIRSAtlas, fNIRSNet23 achieves the strongest cross-subject performance, while LDA on window-based features remains the strongest method in within-subject evaluation. In our benchmark, larger deep learning models do not provide clear advantages for within-subject personalized decoding, but they show improved generalization in the cross-subject setting. This suggests that recent architectures have potential to learn more transferable representations, although their benefits are not yet consistent across evaluation regimes. At the same time, the persistent gap between within-subject and cross-subject performance highlights subject variability as a central challenge for fNIRS decoding.

Earlier studies report similar trends on a smaller scale.
BenchNIRS \cite{benerradi2023benchmarking} evaluated five tasks and found LDA strongest on four within-subject and three cross-subject tasks,
but did not evaluate the fNIRS-specific CNNs that were competitive in this benchmark.
In the cross-subject setting, compact CNNs outperformed heavier models like Transformers in the model comparison of \citet{dissanayake2025deep} and in the original fNIRSNet23 study \cite{wang2023rethinking},
while \citet{kwon2021subject} reported a compact CNN outperforming LDA.
The strength of LDA in standardized evaluations contrasts with the 26 of 32 deep learning studies reviewed by \citet{eastmond2022deep} that reported outperforming traditional machine learning,
highlighting the value of standardized evaluations.

fNIRSAtlas is the largest open fNIRS benchmark,
but it is still limited by the public datasets currently available
and remains modest in absolute scale compared with mature neuroimaging resources.
Some paradigms are still less represented
than others, especially emotion and other non-motor cognitive tasks, although
we expect this to improve as the benchmark grows and new public datasets are
added.
Participants are predominantly young adults (\autoref{app:demographics}),
so the results may not generalize to other age groups.
The unified preprocessing and evaluation pipeline improves comparability
across datasets,
but results and model rankings are specific to these fixed defaults,
which may not be optimal for every dataset or model.
We also note that our empirical comparison is limited to only
10 baselines. This limitation is due to limited code sharing practices in the
fNIRS deep learning literature, which we also expect to improve in the future.
Furthermore, the benchmark currently focuses only on supervised classification in
within-subject and cross-subject settings, and does not yet include approaches
such as transfer learning, self-supervised learning, data augmentation, or
fine-tuning-based adaptation. These settings are important for practical fNIRS
modeling and are natural additions for future versions of the benchmark,
together with new datasets and models.
Continued expansion should also help identify the dataset properties that influence decoding performance.

\paragraph{Code and Data.} Our code and data are publicly available and can be found at \url{https://fnirsatlas.github.io}. The repository provides a fully automated, reproducible pipeline for fNIRS data analysis, including dataset download, preprocessing, classification, and evaluation. Datasets across multiple paradigms come with dependency-free data format, and a configurable nested cross-validation setup for within- and cross-subject experiments. The codebase includes built-in baseline models and support for custom models (e.g., via PyTorch and scikit-learn), statistical comparison tools, and ready-to-use Jupyter notebooks and examples for plug-and-play experimentation and extension. New datasets and tasks can be contributed through pull requests following a contribution tutorial, and dataset requests and corrections can be submitted through structured issue templates.

\begin{ack}
The authors wish to acknowledge CSC -- IT Center for Science, Finland, for computational resources.
The research was partially funded by Business Finland and the PHP Säätiö.
The authors declare no competing interests.
\end{ack}

\bibliographystyle{plainnat}
\bibliography{references}

\appendix
\let\appendixsection\section
\renewcommand{\section}{\FloatBarrier\appendixsection}

\section{Task Informativeness Metrics}
\label{app:task-informativeness-metrics}

For a single task, model, and fold, let $n_{ij}$ denote the number of examples with true class $i$ and predicted class $j$, where $i,j=1,\ldots,C$, and let $N=\sum_{i=1}^{C}\sum_{j=1}^{C}n_{ij}$. Cohen's $\kappa$ is
\begin{equation}
\kappa = \frac{p_o-p_e}{1-p_e},
\end{equation}
where the observed agreement is
\begin{equation}
p_o = \frac{1}{N}\sum_{c=1}^{C}n_{cc},
\end{equation}
and the expected agreement under independence of the true and predicted labels is
\begin{equation}
p_e = \frac{1}{N^2}\sum_{c=1}^{C}\left(\sum_{j=1}^{C}n_{cj}\right)\left(\sum_{i=1}^{C}n_{ic}\right).
\end{equation}

Let $\kappa_{dmf}$ denote the resulting Cohen's $\kappa$ score for task $d$, model $m$, and fold $f$, with $m=1,\ldots,M$ and $f=1,\ldots,F$. Let
\begin{equation}
\bar{\kappa}_{dm\cdot}=\frac{1}{F}\sum_{f=1}^{F}\kappa_{dmf},\quad
\bar{\kappa}_{d\cdot f}=\frac{1}{M}\sum_{m=1}^{M}\kappa_{dmf},\quad
\bar{\kappa}_{d\cdot\cdot}=\frac{1}{MF}\sum_{m=1}^{M}\sum_{f=1}^{F}\kappa_{dmf}.
\end{equation}

Let $T_d$ denote the three models with the highest $\bar{\kappa}_{dm\cdot}$ for task $d$. $\kappa\text{@}3$ is the average fold-level $\kappa$ across these models and folds:
\begin{equation}
\kappa\text{@}3_d = \frac{1}{3F}\sum_{m\in T_d}\sum_{f=1}^{F}\kappa_{dmf}.
\end{equation}

Model separation ratio (MSR) compares the spread of model-wise mean performance to the fold-level variation remaining after removing model and fold means:
\begin{equation}
\mathrm{MSR}_d =
\frac{\operatorname{sd}_{m}\left(\bar{\kappa}_{dm\cdot}\right)}
{\operatorname{sd}_{m,f}\left(\kappa_{dmf}-\bar{\kappa}_{dm\cdot}-\bar{\kappa}_{d\cdot f}+\bar{\kappa}_{d\cdot\cdot}\right)}.
\end{equation}

Rank stability (RS) is computed by paired bootstrap resampling over folds with $B=9999$ resamples. Let $r_d$ be the model ranking induced by $\bar{\kappa}_{dm\cdot}$. For bootstrap sample $b=1,\ldots,B$, sample folds with replacement, recompute model-wise mean scores, and let $r_d^{(b)}$ be the resulting ranking. Rank stability is
\begin{equation}
\mathrm{RS}_d = \frac{1}{B}\sum_{b=1}^{B}\rho_S\left(r_d,r_d^{(b)}\right),
\end{equation}
where $\rho_S$ is Spearman's rank correlation.

The combined informativeness score $U_d$ is the equally weighted average of min--max normalized metrics across tasks. For any metric $x_d$, define
\begin{equation}
\tilde{x}_d = \frac{x_d - \min_{d'} x_{d'}}{\max_{d'} x_{d'} - \min_{d'} x_{d'}}.
\end{equation}
Then
\begin{equation}
U_d = \frac{1}{3}\left(\widetilde{\mathrm{RS}}_d + \widetilde{\kappa\text{@}3}_d + \widetilde{\mathrm{MSR}}_d\right).
\end{equation}

\section{Participant Demographics}
\label{app:demographics}

\autoref{tab:demographics} summarizes the available demographics of the 19 data collections.
Participants were predominantly young adults,
with mean ages of 22 to 33 years.
The exception is Guevara2024, which consists of older adults with a mean age of 65.6 years.
The youngest participant is 17 years old (Shin2018),
and no collection is reported to include children.

\begin{table}[ht]
\footnotesize
\centering
\caption{Participant demographics per data collection. Ages are reported as range and/or mean $\pm$ standard deviation. NR: not reported; M: male; F: female; R: right-handed; L: left-handed.}
\label{tab:demographics}
\setlength{\tabcolsep}{3pt}
\begin{tabular}{lrlll}
\toprule
Collection & N & Age (years) & Sex/gender & Handedness \\
\midrule
Buccino2016 & 15 & 23--54; $27.4 \pm 7.7$ & 15 M & All R \\
Bak2019 & 30 & $23.4 \pm 2.5$ & 17 M, 13 F & 29 R, 1 L \\
Ortega2020 & 14 & NR & NR & All R \\
Luke2021 & 5 & 20--53; 33.0 & 2 M, 3 F & NR \\
Cockx2023 & 24 & 20--29; $24.0 \pm 2.5$ & 10 M, 14 F & All R \\
Guevara2024 & 20 & $65.6 \pm 9.9$ & 9 M, 11 F & All R \\
Luke2024b & 10 & 20--35 & NR & All R \\
Gao2024 & 12 & 18--37; $23.2 \pm 5.2$ & 6 M, 6 F & NR \\
BNCI & 8 & $26 \pm 2.8$ & 3 M, 5 F & NR \\
Shin2016 (MA, MI) & 29 & $28.5 \pm 3.7$ & 14 M, 15 F & 28 R, 1 L \\
Shin2018b & 12 & $26.7 \pm 3.7$ & 5 M, 7 F & NR \\
Shin2019 (MA, MI) & 18 & $23.8 \pm 2.5$ & 10 M, 8 F & NR \\
Yi2025 & 18 & 22--27 & 8 M, 10 F & All R \\
Herff2014 & 10 & Mean 22 & 6 M, 4 F & 8 R, 2 L \\
Shin2018 (n-back, WG) & 26 & 17--33; $26.1 \pm 3.5$ & 9 M, 17 F & All R \\
Huang2021 & 68 & 18--44; $21.71 \pm 4.01$ & 26 M, 39 F, 2 other, 1 declined & 64 R, 3 L, 1 unknown \\
NEMO (EMPE, AFIM) & 31 & 21--52; $31.7 \pm 7.5$ & 18 M, 12 F, 1 non-binary & 27 R, 4 L \\
Shader2021 & 8 & 24--34; $29.8 \pm 3.1$ & 4 M, 4 F & NR \\
Luke2024a & 17 & 22--40 & 7 M, 10 F & NR \\
\bottomrule
\end{tabular}
\end{table}

\section{Dataset Licenses}
\label{app:licenses}

Table~\ref{tab:licenses} summarizes the licensing and access conditions for the datasets included in fNIRSAtlas. Datasets marked as \texttt{REQUEST} require a custom license agreement: the user follows a link to a request form, submits the signed form, and then automatically receives the dataset by email.

\begin{table}[htbp]
\centering
\caption{Licenses and access conditions for the datasets included in fNIRSAtlas.}
\label{tab:licenses}
\small
\begin{tabular}{lll}
\toprule
Suite & Dataset & License \\
\midrule
\multirow[t]{8}{*}{Motor} & Bak2019 \cite{bak2019open} & CC BY 4.0 \\
& Buccino2016 \cite{buccino2016hybrid} & CC BY 4.0 \\
& Cockx2023 \cite{cockx2023fnirs} & CC-BY-SA-4.0 \\
& Gao2024 \cite{gao2023short} & CC0 \\
& Guevara2024 \cite{guevara2024parkinson} & CC-BY-4.0 \\
& Luke2021 \cite{luke2021finger} & Other (Open) \\
& Luke2024b \cite{yucel2024fnirs} & CC-BY-4.0 \\
& Ortega2020 \cite{ortega2020hygrip} & MIT \\
\multirow[t]{4}{*}{MA} & Bnci \cite{bauernfeind2011single} & CC BY-ND 4.0 \\
& Shin2016 (ma) \cite{shin2016open} & GPL-3.0 \\
& Shin2018b \cite{shin2018eyes} & CC-BY-4.0 \\
& Shin2019 (ma) \cite{shin2018ternary} & CC-BY-4.0 \\
\multirow[t]{3}{*}{MI} & Shin2016 (mi) \cite{shin2016open} & GPL-3.0 \\
& Shin2019 (mi) \cite{shin2018ternary} & CC-BY-4.0 \\
& Yi2025 \cite{yi2025multi} & CC-BY-4.0 \\
\multirow[t]{3}{*}{N-back} & Herff2014 \cite{herff2013classification} &
REQUEST \\
& Huang2021 \cite{huang2021tufts} & CC-BY-4.0 \\
& Shin2018 (N-back) \cite{shin2018simultaneous} & CC-BY-4.0 \\
\multirow[t]{2}{*}{Emotion} & Nemo (afim) \cite{spape2023nemo} & CC-BY-NC-4.0
\\
& Nemo (empe) \cite{spape2023nemo} & CC-BY-NC-4.0 \\
\multirow[t]{3}{*}{MISC} & Luke2024a \cite{yucel2024fnirs} & CC-BY-4.0 \\
& Shader2021 \cite{shader2021use} & CC-BY-4.0 \\
& Shin2018 (wg) \cite{shin2018simultaneous} & CC-BY-4.0 \\
\bottomrule
\end{tabular}

\end{table}

\section{Preprocessing Details}
\label{app:preprocessing_details}

This section provides a detailed description of each preprocessing step applied to the fNIRS data.

The raw data of datasets vary, as shown in \autoref{tab:dataset_overview}, and since the preprocessing methods are specific to data types, the preprocessing starts at different points for different datasets.
The datasets that have continuous wave (CW) as the raw format go through all preprocessing steps, those with optical density (OD) skip step 1, and haemoglobin concentration (HbO/HbR) datasets start at step 4.

\paragraph*{Step 1: Optical Density Conversion}

Raw continuous wave fNIRS data is converted to optical density (OD) using the negative logarithm of the light intensity ratio: $OD = -\log(I/I_0)$, where $I$ is the measured intensity and $I_0$ is the baseline intensity.

\paragraph*{Step 2: Channel Quality Assessment}

Poor-quality channels are excluded using the scalp coupling index (SCI) \cite{sci}, which calculates the negative correlation within the heartbeat frequency range (0.7--1.5 Hz) between channels that measure activity at different wavelengths in the same location.
A low SCI value indicates weak coupling, usually due to poor connection with the scalp.
Channel pairs with SCI values lower than 0.2 are marked as bad and interpolated from the closest neighbouring channel of the same wavelength.
This relatively low threshold was selected to avoid discarding large portions of data in sessions where overall coupling was weaker, improving retention across heterogeneous public datasets.
\autoref{tab:sci_summary} shows that most datasets retain the large majority of channels at the benchmark threshold of SCI $<0.2$, with substantially larger rejection rates only at more conservative thresholds.
We therefore aimed to keep eliminations minimal whenever possible, since the benchmark prioritizes broad data coverage and downstream classifiers can in principle learn to downweight persistently noisy channels.

The pipeline aims to rely as much as possible on automated quality control (SCI) rather than manual exclusion to keep the pipeline reproducible and new datasets easy to add.
However, signals were still visually checked during data parsing for the expected format, large artifacts, and responses aligned with event timings.
The only manual exclusion was in Yi2025, where subject 10, block 3 was removed because all SCI values were very close to $-1$. Since SCI measures the correlation between the two wavelengths at the same location in the cardiac band, this pattern strongly suggests a recording error.

\paragraph*{Step 3: Motion Artifact Correction}

Motion artifacts caused by head movements are removed using the temporal derivative distribution repair (TDDR) method \cite{tddr}, which effectively eliminates spike artifacts by repairing large deviations in the temporal derivative of the signal.

\paragraph*{Step 4: Conversion to Haemoglobin Concentrations}

Optical density signals are converted into oxyhemoglobin (HbO) and deoxyhemoglobin (HbR) concentrations using the modified Beer-Lambert law (MBLL) \cite{mbll} with a partial pathlength factor of 6.
This conversion estimates changes in chromophore concentrations based on changes in optical density while accounting for how near-infrared light is absorbed and scattered in biological tissue.

\paragraph*{Step 5: Band-pass Filtering}

A finite impulse response band-pass filter is applied with a low cutoff frequency of 0.01 Hz and a high cutoff frequency of 0.1 Hz.
The high-pass component (0.01 Hz) eliminates slow signal drifts, while the low-pass component (0.1 Hz) removes physiological noise, particularly from the heartbeat.

\paragraph*{Step 6: Segmentation}

Data are segmented into epochs, each containing 15 seconds of post-stimulus recording by default, preceded by a 2-second baseline period.
The baseline period is used for baseline correction, which isolates the stimulus-related response.
The 15-second interval corresponds with the typical haemodynamic response time course \cite{glover2011overview}. Some datasets use longer epochs because their stimuli last longer,
and datasets with shorter event intervals use shorter epochs to prevent overlap between epochs.
A shorter interval could truncate later response components, whereas a longer interval would increase overlap risk and reduce the number of usable trials in denser designs.

\paragraph*{Step 7: Downsampling}

Finally, the signal is downsampled to 5 Hz.
Since the filtered signal is very smooth after band-pass filtering, minimal information is lost during this downsampling process. This also reduces storage and the temporal input length for downstream models.

\begin{figure}[htbp]
  \centering
  \includegraphics[width=0.99\textwidth]{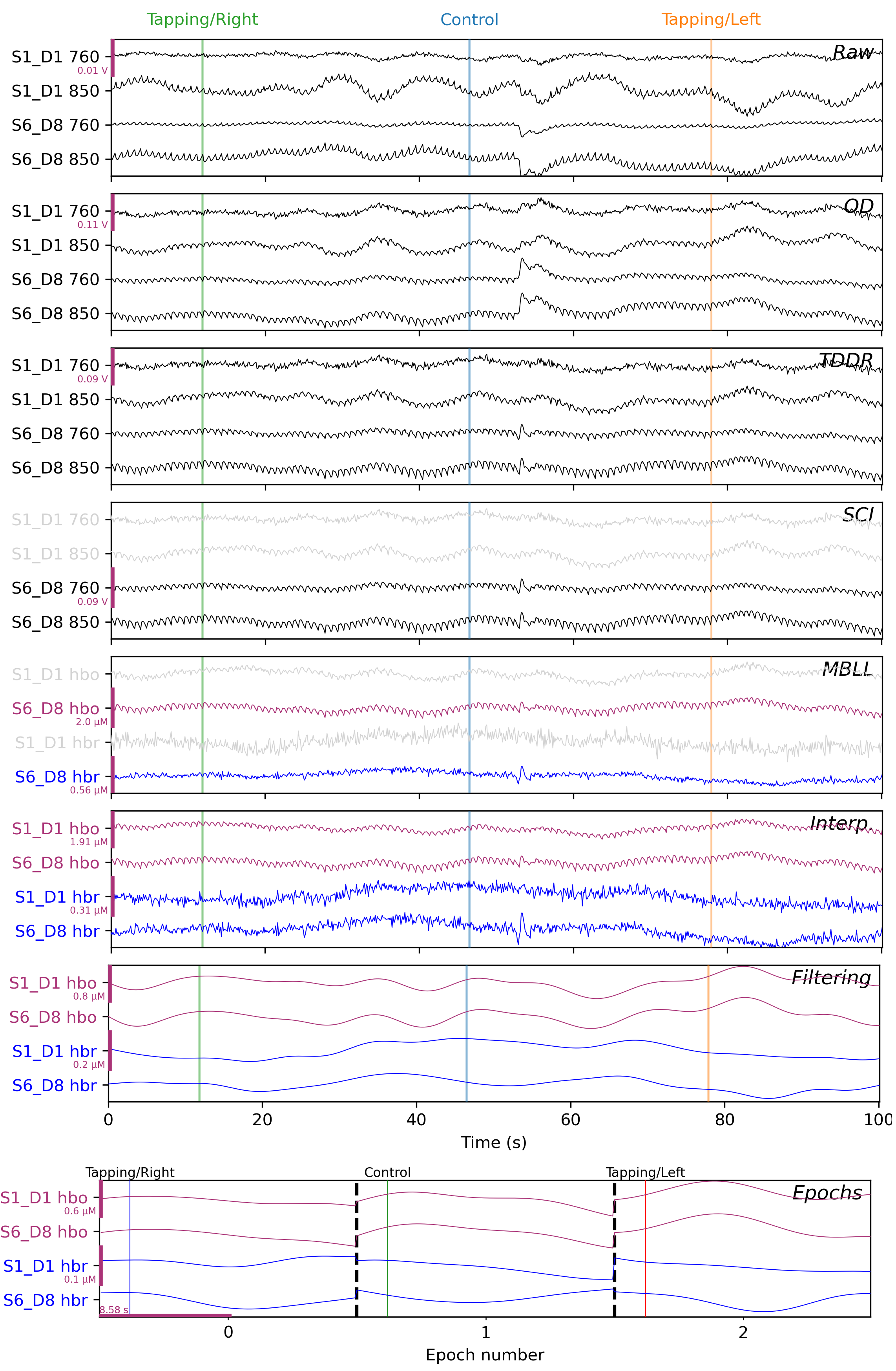}
  \caption{
  Overview of the fNIRS preprocessing steps.
  }
  \label{fig:preprocessing}
\end{figure}

\subsection*{Data Quality Analysis}

\autoref{tab:sci_summary} reports the distribution of SCI values in each dataset together with the proportion of channel pairs falling below several possible rejection thresholds. These summaries are intended to provide a descriptive overview of data quality variation across the benchmark.
As far as we are aware, there is no established channel-quality assessment method for HbO/HbR data. We therefore approximated SCI for HbO/HbR-only datasets by converting the data to OD by inverting the MBLL, and then computed SCI in the same way as for the other datasets.
We excluded Huang2021 from this analysis as its data is recorded with a custom frequency-domain device, and as frequency-domain devices do not use the MBLL, we can not reliably convert it back to OD.
Because this is not an established channel-quality assessment practice, this method was not used in the standard preprocessing pipeline for channel elimination.

\begin{table}[htbp]
\centering
\caption{SCI summary for all datasets.
Elim\%(\textless{}x) denotes the percentage of channel pairs below threshold x.}
\label{tab:sci_summary}
\small
\begin{tabular}{lrrrrrr}
\toprule
Dataset & \#Ch & Mean SCI & Median SCI & Elim\%(<0.2) & Elim\%(<0.5) & Elim\%(<0.7) \\
\midrule
Buccino2016 & 68 & 0.578 & 0.655 & 13.7 & 35.3 & 56.1 \\
Bak2019 & 40 & 0.618 & 0.715 & 14.2 & 32.2 & 48.3 \\
Ortega2020 & 48 & 0.824 & 0.914 & 5.1 & 9.5 & 16.4 \\
Luke2021 & 56 & 0.789 & 0.902 & 5.7 & 16.4 & 23.6 \\
Cockx2023 & 72 & 0.947 & 0.978 & 0.2 & 1.4 & 3.4 \\
Guevara2024 & 44 & 0.986 & 1.000 & 0.9 & 1.1 & 1.6 \\
Luke2024b & 136 & 0.821 & 0.981 & 7.1 & 13.3 & 16.8 \\
Gao2024 & 200 & 0.805 & 0.947 & 6.2 & 15.6 & 23.7 \\
Bnci & 104 & 0.807 & 0.932 & 6.7 & 11.3 & 16.7 \\
Shin2016 (ma) & 72 & 0.790 & 0.956 & 9.5 & 18.4 & 25.0 \\
Shin2018b & 18 & 0.913 & 0.978 & 3.7 & 4.6 & 6.5 \\
Shin2019 (ma) & 32 & 0.898 & 0.963 & 1.1 & 3.3 & 8.5 \\
Shin2016 (mi) & 72 & 0.793 & 0.959 & 9.3 & 18.4 & 24.8 \\
Shin2019 (mi) & 32 & 0.898 & 0.963 & 1.1 & 3.3 & 8.5 \\
Yi2025 & 48 & 0.509 & 0.508 & 24.5 & 49.4 & 66.6 \\
Herff2014 & 16 & 0.929 & 0.938 & 0.0 & 0.0 & 0.0 \\
Shin2018 (nback) & 72 & 0.756 & 0.912 & 9.1 & 21.6 & 30.8 \\
Nemo (empe) & 48 & 0.834 & 0.970 & 6.7 & 14.4 & 18.5 \\
Nemo (afim) & 48 & 0.834 & 0.970 & 6.7 & 14.4 & 18.5 \\
Shin2018 (wg) & 72 & 0.751 & 0.908 & 9.1 & 21.8 & 30.6 \\
Shader2021 & 104 & 0.852 & 0.941 & 2.6 & 8.7 & 16.1 \\
Luke2024a & 66 & 0.780 & 0.921 & 9.6 & 16.9 & 22.6 \\
\bottomrule
\end{tabular}

\end{table}

Most datasets showed broadly reasonable SCI profiles, without obvious signs of widespread channel-quality problems, although a small number of datasets stood out from this overall pattern. Buccino2016 and Yi2025 showed comparatively weaker SCI profiles, which may reflect their hybrid EEG-fNIRS recording setup. Bak2019 also showed comparatively weak SCI values, which should be interpreted cautiously because the SCI values were computed after approximate inversion to optical density.

\section{Comparison with BenchNIRS}
\label{app:benchnirs_comparison}

BenchNIRS \citep{benerradi2023benchmarking} is the closest prior benchmark to
fNIRSAtlas and established the need for more rigorous multi-dataset evaluation
in fNIRS machine learning. The two resources share the goal of improving
reproducibility, but differ in scale and benchmark design.

BenchNIRS evaluates five tasks, whereas fNIRSAtlas includes 23 tasks
spanning motor execution, mental arithmetic, motor imagery, n-back, emotion,
and other cognitive tasks. This larger benchmark covers a broader range of
devices, channel counts, sampling rates, raw data formats, and task designs.

The preprocessing and evaluation setups also differ. BenchNIRS uses strong input harmonization, including spatial aggregation \citep{benerradi2023benchmarking},
whereas fNIRSAtlas uses a unified pipeline that retains more of the native dataset structure.
fNIRSAtlas reaches higher baseline accuracy on four of the five shared tasks (\autoref{tab:benchnirs_shared_accuracy}),
which we suspect is due to the spatial aggregation of channels in BenchNIRS.
We investigate the impact of spatial aggregation in \autoref{app:resolution},
which shows a clear decline in LDA accuracy as the degree of aggregation increases.

\begin{table}[htbp]
\centering
\caption{Cross-subject LDA accuracy (\%) on tasks shared with BenchNIRS.}
\label{tab:benchnirs_shared_accuracy}
\small
\begin{tabular}{lcc}
\toprule
Dataset & BenchNIRS & Ours \\
\midrule
Shin2018 (nback) & 36.0 & 49.2 \\
Shin2018 (wg)    & 56.8 & 63.7 \\
Shin2016 (ma)    & 56.4 & 68.8 \\
Bak2019*         & 42.7 & 64.8 \\
Herff2014        & 38.5 & 37.6 \\
\bottomrule
\end{tabular}
\end{table}

*Bak2019 is evaluated here as a three-class task (left-hand, right-hand, and
foot tapping), which is why it differs from Table~\ref{tab:dataset_overview}.

Overall, BenchNIRS introduced the first rigorous multi-dataset fNIRS benchmark,
and fNIRSAtlas extends it to a larger benchmark with stronger baseline results.

\section{Baseline Model Details}
\label{app:baseline-details}

This appendix briefly describes the baseline models used in the benchmark and the
model-specific hyperparameter search spaces.

\paragraph*{Linear Discriminant Analysis (LDA)} \cite{hastie2009elements} is the most commonly used
classification method in fNIRS-based brain--computer interface (BCI) studies due
to its solid performance in high-dimensional, low-sample settings, fast
training and inference times, and the lack of hyperparameters to tune
\cite{blankertz2011single, naseer2015fnirs, hong2018feature,
bauernfeind2014single, benerradi2023benchmarking}. We use shrinkage
regularization because it improves covariance estimation in high-dimensional
settings. The shrinkage strength is determined analytically using the Ledoit
and Wolf lemma \cite{ledoitwolflemma}. The input features were selected with
grid search over the number of non-overlapping windows (1, 3, 5, 15) and
feature type (mean, slope, standard deviation, mean+slope). The
features were z-normalized independently across samples using statistics from
the training split only. These representations are commonly used and have shown
strong performance in comparative fNIRS studies \cite{li2021fugl, fernandez2019machine,
batula2014evaluating, hong2018feature, naseer2015fnirs, bauernfeind2014single,
treacy2015designing}.

\paragraph*{Artificial Neural Network (ANN)} \cite{naseer2015comparison} is included as an early ANN application for fNIRS classification, which has also been used as a baseline in later studies \cite{li2025metanirs}.
Their implemetation used a single layer with 20 hidden units, a tanh activation function, and mean and slope for each channel as input.
We select hyperarameters with grid search over the number of windows in feature extraction (1, 50), activation function (ReLU, tanh), hidden layer size (20, 256), and initial learning rate (0.001, 0.01).

\paragraph*{Convolutional Neural Network (CNN)} \cite{trakoolwilaiwan2018convolutional} is included as an early fNIRS adaptation of a CNN baseline, which has also been used as a baseline in later fNIRS studies \cite{li2025metanirs}.
In the original paper, it outperformed SVM and ANN baselines for within-subject fNIRS motor classification on a private dataset.
The architecture applies three 1D convolutional blocks with ReLU activations, max pooling, and dropout to the haemoglobin time series, using filters that span all input channels while convolving along time and thus capture spatio-temporal structure.
These convolutional blocks are followed by two fully connected layers and a softmax classification head.
Hyperparameters were selected with grid search over starting learning rate (0.01, 0.001), dropout (0.25, 0.5), number of filters (16, 32), and kernel size (3, 10).

\paragraph*{Long short-term memory (LSTM)} \cite{asgher2020enhanced} is included as an early fNIRS adaptation of an LSTM baseline, which has also been used as a baseline in later fNIRS studies \cite{li2025metanirs}.
In the original paper, it outperformed CNN, ANN, SVM, and k-NN baselines for within-subject four-class mental workload classification on a self-collected dataset.
The architecture uses four stacked LSTM layers with hidden size 64 to model temporal dependencies in the haemoglobin time series.
Hyperparameters were selected with grid search over starting learning rate (0.01, 0.001), hidden size (32, 64), number of layers (2, 4), and dropout (0.25, 0.5).

\paragraph*{EEGNet} \cite{lawhern2018eegnet} is a lightweight convolutional neural
network tailored for EEG-based BCI, but it is commonly used in fNIRS research
as well \cite{kwon2021subject, huang2021tufts} as a baseline against custom deep learning (DL)
models. The architecture consists of a temporal, a depthwise and a separable
convolutional layer, followed by a single linear layer as the classifcation
head. These convolutional layes allow EEGNet to capture temporal and spatial
patterns in the fNIRS data with very few trainable parameters, which makes it
less prone to overfitting and suitable for datasets with limited samples. Hyperparameters
were selected with grid search over starting learning rate (0.01, 0.001),
dropout (0.25, 0.5), number of temporal filters (4, 16), and temporal kernel
size (3, 10).

\paragraph*{DeepConvNet} (DCN) \cite{dcn} is a deeper convolutional neural network
originally proposed for EEG-based BCI and later adopted in fNIRS research
\cite{huang2021tufts}. It consists of an initial temporal and spatial
convolution followed by four temporal convolutional blocks with batch
normalization, exponential linear unit (ELU), and max pooling, and ends with a convolutional
classification head. Hyperparameters were selected
with grid search over starting learning rate (0.01, 0.001), dropout (0.25,
0.5), number of filters in the first block (4, 25), and kernel size (3, 10).

\paragraph*{MultiRocket+Hydra} (MRH) \cite{dempster2023hydra} is a computationally efficient state-of-the-art method for general time-series classification that combines MultiRocket \cite{tan2022multirocket} and Hydra \cite{middlehurst2024bake}.
Older model variants from the ROCKET family \cite{dempster2019rocket} have been applied in fNIRS classification with promising performance \cite{andreu2021single}, making MRH a fitting baseline for this study.
MultiRocket applies random convolutional kernels to the raw and first-order differenced signal and summarizes the resulting responses with simple pooling statistics, such as percentage of positive values \cite{tan2022multirocket}.
Hydra also uses random convolutional kernels, but organizes them into \(g\) groups of \(k\) kernels and, at each time point, counts which kernel in each group produces the strongest response, producing \(g \times k\) group-wise count features.
Finally, the features from MultiRocket and Hydra are concatenated and used to train a linear classifier.
Hyperparameters were selected with grid search over the number of kernels per group (2, 4, 8, 32) and the number of groups (8, 16, 64, 128).

\paragraph*{fNIRS-T} \cite{wang2022transformer} is the first and most widely used
Transformer model in fNIRS research \cite{abdollahpour2025advancing}. In the
original paper, it also reported very strong results on the BNCI, Bak2019, and
Shin2016 (ma) datasets included in this benchmark. It uses a dual-tower
embedding structure that processes channel-level and patch-level information
separately, with convolutional layers in both towers. Hyperparameters were
selected with grid search over starting learning rate (0.01, 0.001), embedding
dimension (64, 128), number of layers (2, 6), and dropout (0.0, 0.3).

\paragraph*{fNIRSNet23} \cite{wang2023rethinking} is an fNIRS-specific convolutional neural network that was proposed to better match the delayed haemodynamic dynamics of fNIRS signals compared to EEGNet.
It has also been used as a baseline in later fNIRS studies \cite{li2025metanirs}.
In the original paper, fNIRSNet23 outperformed CNN, temporal CNN, LSTM, and fNIRS-T baselines on the reported within-session and cross-subject experiments on the Bak2019 and Shin2016 datasets.
The model uses a compact convolutional architecture built around large temporal kernels, depthwise separable convolutions, and activation functions with saturated negative values to capture delayed haemodynamic responses while keeping the network simple.
Hyperparameters were selected with grid search over starting learning rate (0.01, 0.001), number of filters in the first convolutional layer (2, 4, 8), and number of filters in the final convolutional layer (8, 16, 32).

\paragraph*{fNIRSNet24} \cite{pandey2024fnirsnet} is an fNIRS-specific convolutional neural network that uses a multi-view spatio-temporal design for classification.
Relative to simpler other CNN baselines, this architecture is a step-up in complexity as it has a spatio-temporal and temporal-spatial feature extraction branches that are combined in a fusion module.
In the original paper, fNIRSNet24 outperformed the reported baselines, including SLDA, DCN, LSTM, and many other models, on a private auditory oddball dataset.
Hyperparameters were selected with grid search over starting learning rate (0.01, 0.001), dropout (0.25, 0.5), number of temporal filters (4, 20), and number of spatial filters (4, 60).

\section{Classical Baselines}
\label{app:classical}

As a robustness check of whether the main conclusions remain unchanged with additional classical baselines, we evaluated logistic regression (LR), support vector machine with radial basis function kernel (RBF-SVM), random forest (RF), and k-nearest neighbors (kNN) under the same protocol as the main baselines.

All four models use the scikit-learn implementations and take as input the standardized channel-wise mean over non-overlapping windows, with the number of windows selected from (1, 5).
Hyperparameters were selected with grid search over the following values.
LR: penalty (L1, L2) and inverse regularization strength $C$ ($10^{-5}$, $10^{-2}$, 1, 100), using the SAGA solver.
RBF-SVM: $C$ ($10^{-4}$, $10^{-3}$, $10^{-2}$, $10^{-1}$, 1, 10, 100, 1000), with the kernel width set to scikit-learn's \texttt{scale} default.
RF: number of trees (10, 50, 100, 1000) and minimum number of samples per leaf (1, 2).
kNN: number of neighbors (1, 3, 5, 7) and neighbor weighting (uniform, inverse distance).

\autoref{tab:classical-cross} and \autoref{tab:classical-within} show the results.
Ranked by mean accuracy together with the main baselines, LR was second in both settings, with 57.2\% cross-subject and 62.6\% within-subject accuracy.
Of the 14 models, RBF-SVM ranked 7th cross-subject and 9th within-subject, RF ranked 11th in both settings, and kNN ranked last in both settings.
fNIRSNet23 and LDA remained the strongest models in the cross-subject and within-subject settings, respectively.
These results align with previous evaluations of these models in BenchNIRS \cite{benerradi2023benchmarking} and NEMO \cite{spape2023nemo}.
The strong performance of LDA therefore extends to LR, but not to the non-linear classical models.

\begin{table}[htbp]
\centering
\caption{Cross-subject classification accuracy for the additional classical baselines.
Values show mean test accuracy $\pm$ standard deviation of fold-level accuracies, describing cross-validation variability.
Bold indicates the highest value in each row.
\#Cl: number of classes;
\#Ep: number of epochs.
Task abbreviations as in \autoref{tab:task-informativeness}.}
\label{tab:classical-cross}
\footnotesize
\setlength{\tabcolsep}{4pt}
\begin{tabular}{l l c c *{4}{c}}
\toprule
Dataset & Task & \#Cl & \#Ep & LR & RBF-SVM & RF & kNN \\
\midrule
Buccino2016 & Side[R|L]×Joint[Arm|Hand] & 4 & 1500 & 42.7±4.5 & 42.5±3.1 & \textbf{43.1±3.2} & 30.9±1.5 \\
Bak2019 & FingerTap[L|R] & 2 & 1500 & \textbf{81.9±4.4} & 79.2±4.3 & 80.3±3.7 & 69.9±3.9 \\
Ortega2020 & HandGrip[L|R] & 2 & 328 & \textbf{63.4±6.7} & 58.4±7.7 & 55.0±4.8 & 49.5±5.6 \\
Luke2021 & FingerTap[L|R] & 2 & 300 & \textbf{84.1±10.4} & 80.3±12.0 & 70.7±12.3 & 63.0±9.5 \\
Cockx2023 & Tapping[Finger|Foot] & 2 & 1844 & \textbf{84.7±4.1} & 84.4±4.3 & 84.1±3.5 & 73.9±6.0 \\
Guevara2024 & FingerTap[L|R] & 2 & 395 & \textbf{65.7±5.5} & 65.1±4.9 & 60.1±4.8 & 55.9±5.2 \\
Luke2024b & FingerTap[L|R] & 2 & 1081 & \textbf{54.8±2.4} & 52.6±3.6 & 52.2±3.9 & 52.8±3.2 \\
Gao2024 & HandSqueeze[L|R] & 2 & 1080 & \textbf{68.3±8.0} & 65.5±7.8 & 62.8±6.5 & 52.2±4.4 \\
Bnci & BL|MA & 2 & 348 & 86.6±7.9 & 84.8±10.1 & \textbf{86.7±8.4} & 78.5±7.8 \\
Shin2016 (ma) & BL|MA & 2 & 1740 & \textbf{70.6±3.1} & 68.2±3.4 & 67.7±3.4 & 57.7±2.5 \\
Shin2018b & BL|MA & 2 & 720 & \textbf{68.7±6.7} & 66.4±5.7 & 67.2±8.2 & 60.9±4.0 \\
Shin2019 (ma) & BL|MA & 2 & 1080 & \textbf{67.7±4.4} & 65.2±4.5 & 65.6±4.5 & 59.0±5.0 \\
Shin2016 (mi) & LMI|RMI & 2 & 1740 & \textbf{54.5±2.3} & 53.4±2.6 & 53.7±2.4 & 52.7±2.0 \\
Shin2019 (mi) & MI|Rest & 2 & 1080 & 54.7±5.1 & 54.8±5.6 & \textbf{55.0±4.4} & 50.8±4.3 \\
Yi2025 & Limb[Shoulder|Elbow|Wrist] & 3 & 4305 & 34.7±1.5 & 34.5±1.7 & \textbf{34.7±1.6} & 33.1±1.7 \\
Herff2014 & 1-back|2-back|3-back & 3 & 300 & \textbf{34.9±6.8} & 32.6±5.3 & 33.1±6.7 & 31.6±4.6 \\
Shin2018 (nback) & 0-back|2-back|3-back & 3 & 702 & \textbf{48.7±4.6} & 44.7±3.6 & 42.1±4.5 & 35.8±3.8 \\
Huang2021 & 0-back|1-back|2-back|3-back & 4 & 1088 & \textbf{33.1±3.0} & 31.5±2.8 & 30.3±3.2 & 28.2±2.0 \\
Nemo (empe) & Aro[H|L]×Val[P|N] & 4 & 1203 & 26.2±2.5 & 25.8±2.4 & \textbf{26.7±2.3} & 26.3±2.7 \\
Nemo (afim) & Aro[H|L]×Val[P|N] & 4 & 718 & 27.2±2.5 & 27.8±3.0 & 25.9±3.6 & \textbf{27.8±3.4} \\
Shin2018 (wg) & BL|WG & 2 & 1560 & \textbf{65.8±2.3} & 64.1±2.7 & 64.8±2.9 & 58.2±3.1 \\
Shader2021 & Audio|Control|Video & 3 & 368 & \textbf{51.7±7.3} & 48.3±5.9 & 46.5±6.6 & 42.0±6.6 \\
Luke2024a & Control|Noise|Speech & 3 & 1020 & \textbf{44.6±3.3} & 43.5±3.8 & 40.1±4.4 & 38.5±3.8 \\
\midrule
Mean &  &  &  & \textbf{57.2} & 55.4 & 54.3 & 49.1 \\
\bottomrule
\end{tabular}

\end{table}

\begin{table}[htbp]
\centering
\caption{Within-subject classification accuracy for the additional classical baselines.
Values show mean test accuracy $\pm$ standard deviation across subjects.
Bold indicates the highest value in each row.
\#Cl: number of classes;
\#Ep: number of epochs.
Task abbreviations as in \autoref{tab:task-informativeness}.}
\label{tab:classical-within}
\footnotesize
\setlength{\tabcolsep}{4pt}
\begin{tabular}{l l c c *{4}{c}}
\toprule
Dataset & Task & \#Cl & \#Ep & LR & RBF-SVM & RF & kNN \\
\midrule
Buccino2016 & Side[R|L]×Joint[Arm|Hand] & 4 & 1500 & \textbf{66.2±10.1} & 60.1±9.7 & 56.4±8.7 & 45.2±9.4 \\
Bak2019 & FingerTap[L|R] & 2 & 1500 & \textbf{82.3±15.2} & 74.1±17.3 & 72.7±17.3 & 65.3±18.6 \\
Ortega2020 & HandGrip[L|R] & 2 & 328 & \textbf{60.8±16.8} & 55.8±15.2 & 57.9±14.6 & 53.1±11.4 \\
Luke2021 & FingerTap[L|R] & 2 & 300 & \textbf{86.7±15.0} & 73.8±17.9 & 75.3±16.1 & 65.9±17.1 \\
Cockx2023 & Tapping[Finger|Foot] & 2 & 1844 & \textbf{94.5±5.5} & 93.6±4.9 & 91.3±4.6 & 86.5±7.4 \\
Guevara2024 & FingerTap[L|R] & 2 & 395 & \textbf{59.2±13.6} & 55.3±14.6 & 56.1±13.5 & 55.8±13.5 \\
Luke2024b & FingerTap[L|R] & 2 & 1081 & 60.6±8.3 & \textbf{63.4±10.6} & 62.6±10.0 & 58.7±11.1 \\
Gao2024 & HandSqueeze[L|R] & 2 & 1080 & \textbf{75.3±13.6} & 63.8±11.8 & 63.9±13.5 & 56.4±7.9 \\
Bnci & BL|MA & 2 & 348 & \textbf{97.3±2.7} & 95.0±4.5 & 93.4±4.2 & 88.3±10.8 \\
Shin2016 (ma) & BL|MA & 2 & 1740 & \textbf{83.1±7.8} & 78.4±8.1 & 76.2±9.5 & 68.0±9.6 \\
Shin2018b & BL|MA & 2 & 720 & \textbf{75.7±9.8} & 71.3±7.9 & 71.7±7.1 & 67.2±8.2 \\
Shin2019 (ma) & BL|MA & 2 & 1080 & \textbf{77.8±9.2} & 73.5±9.8 & 71.9±8.5 & 65.6±8.9 \\
Shin2016 (mi) & LMI|RMI & 2 & 1740 & \textbf{58.5±8.2} & 52.6±9.2 & 51.2±8.9 & 51.5±8.8 \\
Shin2019 (mi) & MI|Rest & 2 & 1080 & \textbf{67.2±9.2} & 64.5±8.9 & 61.4±8.2 & 58.5±8.3 \\
Yi2025 & Limb[Shoulder|Elbow|Wrist] & 3 & 4305 & \textbf{38.9±5.1} & 38.0±4.5 & 37.3±3.7 & 35.7±3.7 \\
Herff2014 & 1-back|2-back|3-back & 3 & 300 & \textbf{50.1±10.6} & 44.3±11.6 & 43.3±13.4 & 38.5±12.3 \\
Shin2018 (nback) & 0-back|2-back|3-back & 3 & 702 & \textbf{44.1±12.7} & 40.3±12.7 & 38.3±12.3 & 37.4±10.2 \\
Huang2021 & 0-back|1-back|2-back|3-back & 4 & 1088 & 25.2±12.7 & 22.7±11.9 & \textbf{28.3±12.4} & 23.8±10.8 \\
Nemo (empe) & Aro[H|L]×Val[P|N] & 4 & 1203 & \textbf{33.6±10.5} & 28.6±9.4 & 29.8±9.1 & 28.7±8.0 \\
Nemo (afim) & Aro[H|L]×Val[P|N] & 4 & 718 & \textbf{24.6±7.9} & 21.3±10.0 & 23.5±10.3 & 22.7±9.5 \\
Shin2018 (wg) & BL|WG & 2 & 1560 & \textbf{74.3±10.7} & 68.7±10.9 & 66.4±9.3 & 61.3±9.7 \\
Shader2021 & Audio|Control|Video & 3 & 368 & 58.2±16.1 & 56.5±14.0 & \textbf{58.3±14.7} & 49.8±11.8 \\
Luke2024a & Control|Noise|Speech & 3 & 1020 & \textbf{46.8±11.0} & 42.1±10.7 & 43.5±13.0 & 37.3±9.3 \\
\midrule
Mean &  &  &  & \textbf{62.6} & 58.2 & 57.9 & 53.1 \\
\bottomrule
\end{tabular}

\end{table}

\section{Collection-Level Statistical Comparison}
\label{app:collection}

Multiple tasks derived from the same data collection, as in Shin2016, Shin2018, and NEMO, could influence the statistical comparison results.
To check this, we repeated the Friedman and Nemenyi tests at the level of data collections, averaging the task-level mean accuracies of each model within each of the 19 collections.
Shin2016 (ma, mi), Shin2018 (n-back, WG), Shin2019 (ma, mi), and NEMO (EMPE, AFIM) were each merged into one collection, and Shin2018b was kept separate.

The Friedman test detected significant differences among models in both the cross-subject ($\chi^2=79.823$, $p<0.001$) and within-subject ($\chi^2=95.704$, $p<0.001$) settings.
The highest-ranked model and the group of models not significantly different from it were the same as in the task-level analysis.
In the cross-subject setting, two of the 45 pairwise decisions changed: LDA became significantly better than MRH, and LDA was no longer significantly better than CNN.
In the within-subject setting, no pairwise decision changed.
Averaging tasks within data collections therefore does not change the main conclusions.

\section{Learning Curves}
\label{app:learning-curve}

To examine how cross-subject performance scales with the number of training subjects,
we evaluated all models with an increasing number of subjects in the training set.

Each model was trained on 1, 2, 4, 8, and 16 subjects and tested on the same held-out subjects.
The held-out subjects were the test folds of a 5-fold cross-subject cross-validation,
so differences between training sizes are not caused by different test subjects.
In every fold, 16 randomly drawn training subjects were split into 16 subsets of 1 subject, 8 of 2, 4 of 4, 2 of 8, and 1 of 16,
so every training size uses all 16 subjects once.
The procedure was repeated with 5 seeds, each with different folds and subsets.

Tasks were selected by two criteria.
First, drawing 16 training subjects from 4 of 5 folds requires at least 20 subjects.
Second, to obtain a representative sample,
we selected from fNIRSAtlas-Balanced (\autoref{tab:task-informativeness}), which retains the most informative tasks of every suite.
Nine tasks meet both:
Bak2019, Cockx2023, Guevara2024, Shin2016 (ma), Shin2016 (mi), Shin2018 (n-back), Shin2018 (wg), Huang2021, and NEMO (afim).
Since these tasks have different numbers of classes, we report Cohen's $\kappa$, which corrects for chance level.
$\kappa$ was averaged within each task over seeds, folds, and subsets,
and then across tasks, so that each task contributes equally.

To limit computational cost, each model used one hyperparameter setting for all training sizes,
the one with the highest mean $\kappa$ over these tasks in the nested cross-subject search (\autoref{app:hparams}).
As this setting was selected on the same tasks, we interpret only changes in performance with training size, not absolute performance.

\begin{figure}[htbp]
  \centering
  \begin{minipage}{0.49\textwidth}
    \centering
    \includegraphics[width=\textwidth]{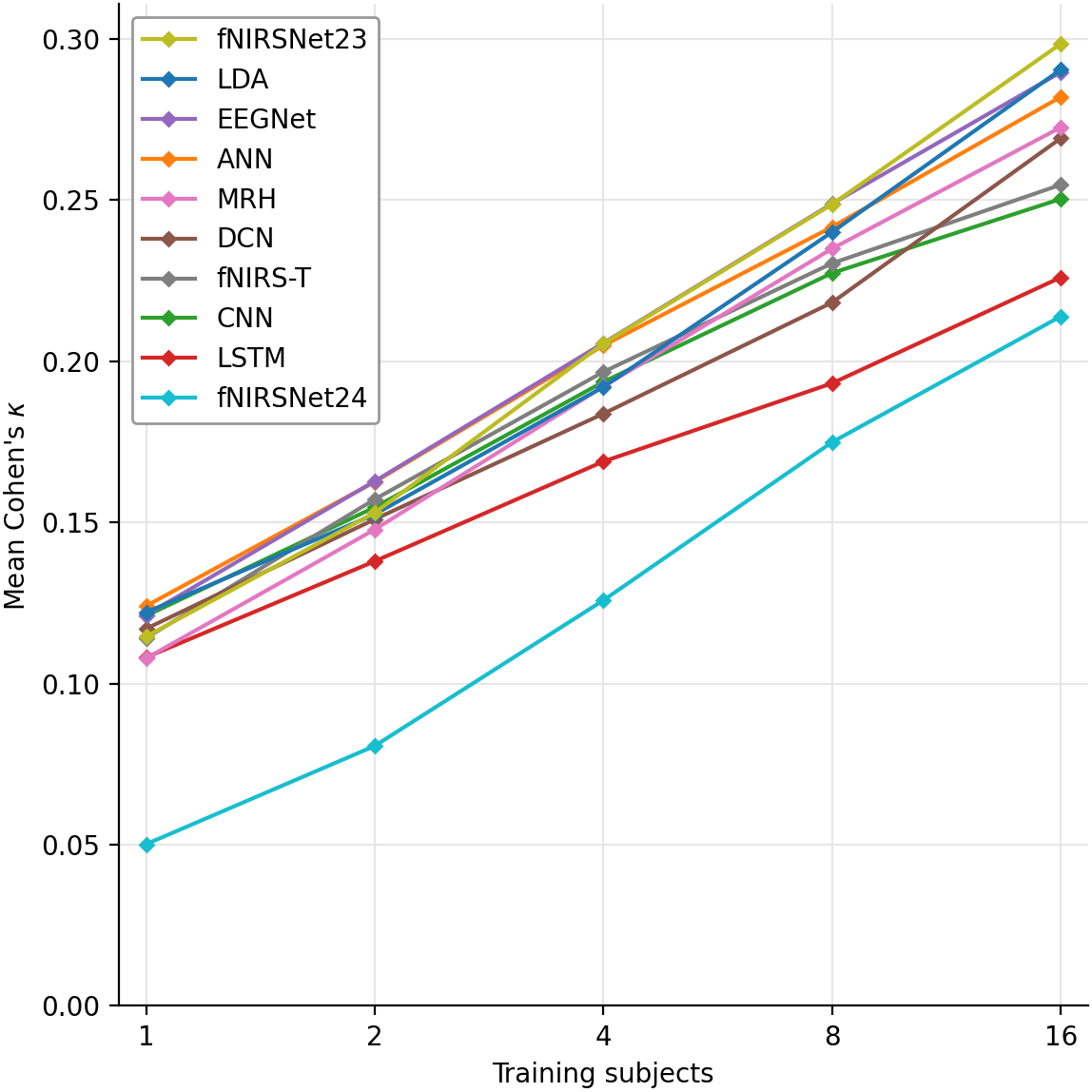}\\
    (a)
  \end{minipage}
  \hfill
  \begin{minipage}{0.49\textwidth}
    \centering
    \includegraphics[width=\textwidth]{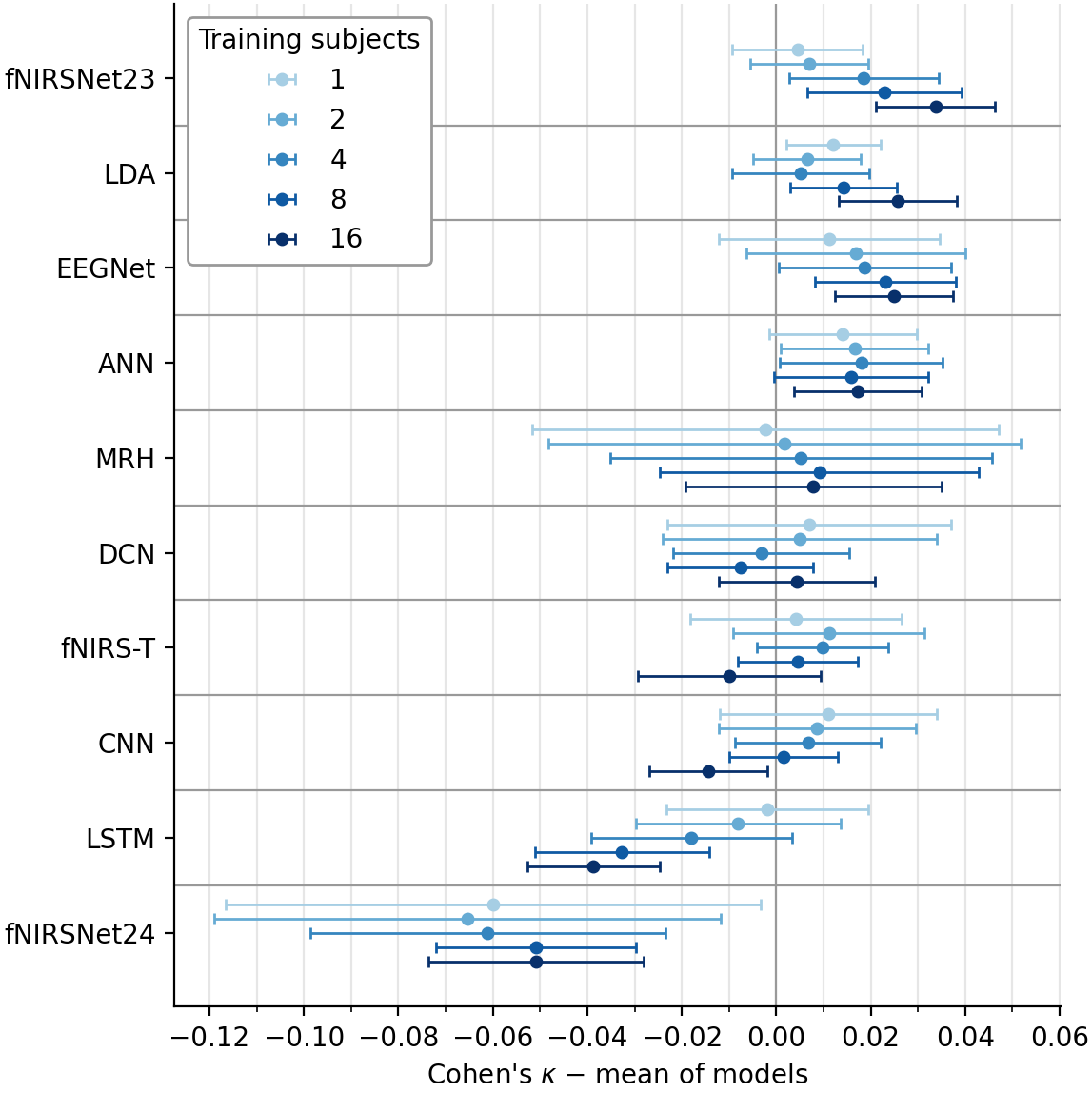}\\
    (b)
  \end{minipage}
  \caption{Learning curves averaged over 9 tasks. (a) Mean $\kappa$ of each model against the number of training subjects (log$_2$ scale). (b) Each model's $\kappa$ minus the mean $\kappa$ of all models on the same task and training size. Error bars show 95\% t-based confidence intervals of the mean across the 9 tasks. Point shade indicates the number of training subjects, from 1 (light) to 16 (dark).}
  \label{fig:lc}
\end{figure}

\autoref{fig:lc}a shows the mean $\kappa$ of each model as a function of the number of training subjects,
and \autoref{tab:lc-kappa} lists the corresponding values together with the gain over 1 subject.
Every model improved with each doubling of training subjects,
on average by about 0.04 $\kappa$,
and on average this improvement did not slow down up to 16 subjects.

Model differences are seen more clearly in \autoref{fig:lc}b, which shows each model's $\kappa$ minus the mean of all models on the same task and training size.
With 1 subject, the models performed similarly,
apart from fNIRSNet24, the weakest model at every size.
As the number of subjects increased, the differences between models became more pronounced,
with fNIRSNet23, LDA, and EEGNet gaining the most,
and fNIRS-T, CNN, and LSTM the least.

\begin{table}[htbp]
\centering
\caption{Mean $\kappa$ ($\times 100$) of each model by number of training subjects, averaged over 9 tasks. Parentheses give the gain over 1 subject. Bold marks the best model per training size.}
\label{tab:lc-kappa}
\small
\begin{tabular}{lrrrrr}
\toprule
Training subjects & 1 & 2 & 4 & 8 & 16 \\
\midrule
fNIRSNet23 & 11.5 & 15.3 (+3.8) & 20.5 (+9.1) & \textbf{24.9} (+13.4) & \textbf{29.8} (+18.4) \\
LDA & 12.2 & 15.2 (+3.0) & 19.2 (+7.0) & 24.0 (+11.8) & 29.0 (+16.8) \\
EEGNet & 12.1 & \textbf{16.3} (+4.2) & \textbf{20.6} (+8.4) & \textbf{24.9} (+12.8) & 29.0 (+16.8) \\
ANN & \textbf{12.4} & \textbf{16.3} (+3.9) & 20.5 (+8.1) & 24.2 (+11.8) & 28.2 (+15.8) \\
MRH & 10.8 & 14.8 (+4.0) & 19.2 (+8.4) & 23.5 (+12.7) & 27.3 (+16.5) \\
DCN & 11.7 & 15.1 (+3.4) & 18.4 (+6.7) & 21.8 (+10.1) & 26.9 (+15.2) \\
fNIRS-T & 11.4 & 15.7 (+4.3) & 19.7 (+8.3) & 23.0 (+11.6) & 25.5 (+14.1) \\
CNN & 12.1 & 15.5 (+3.4) & 19.4 (+7.3) & 22.7 (+10.6) & 25.0 (+12.9) \\
LSTM & 10.8 & 13.8 (+3.0) & 16.9 (+6.1) & 19.3 (+8.5) & 22.6 (+11.8) \\
fNIRSNet24 & 5.0 & 8.1 (+3.0) & 12.6 (+7.6) & 17.5 (+12.5) & 21.4 (+16.4) \\
\bottomrule
\end{tabular}

\end{table}

\section{Ranking Stability and Number of Tasks}
\label{app:benchmark-size}

This appendix analyses how closely the model ranking matches the full-benchmark ranking,
and how many model differences are significant,
as the number of benchmark tasks grows.
We used the cross-subject accuracy of the 10 models, averaged over 5 seeds and folds,
on the 22 independent tasks of the main model comparison (Shin2019 (mi) excluded).
For each subset size $k$ from 2 to 22,
we used all $k$-task subsets if there were at most 1000, and otherwise 1000 random subsets
(e.g., all 231 subsets for $k = 2$, and 1000 of the 26,334 subsets for $k = 5$).
In each subset, we ranked the models by their mean rank across tasks, as in the main results,
and computed Spearman's $\rho$ between this ranking and the full ranking.
We report the median and the 95\% interval (2.5th to 97.5th percentile) of $\rho$ across subsets.
To measure how many significant differences a subset can detect,
we also counted in each subset the model pairs whose mean ranks differed by more than the Nemenyi critical difference (CD, $\alpha = 0.05$).
Additionally, we computed these metrics on the 5 tasks shared with BenchNIRS~\cite{benerradi2023benchmarking}
for a concrete comparison between the two benchmarks.

\begin{figure}[htbp]
  \centering
  \includegraphics[width=\textwidth]{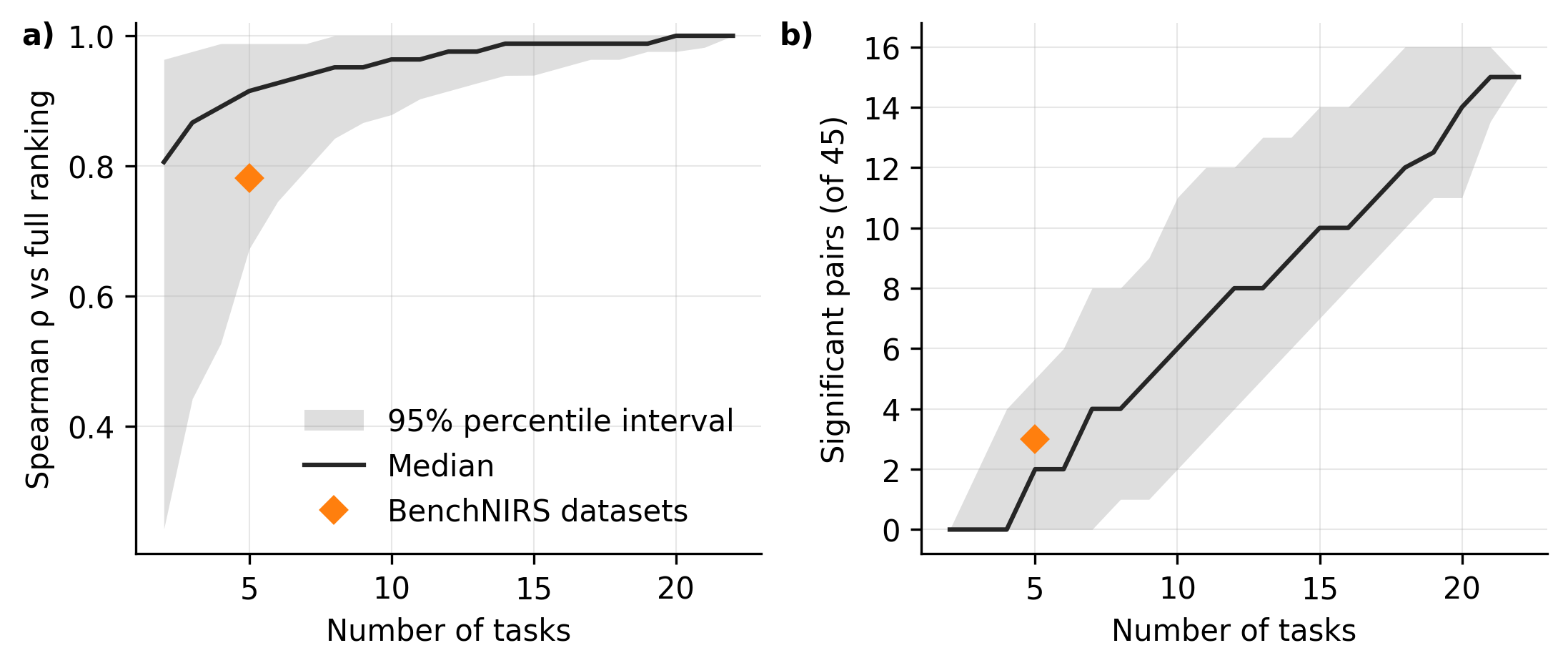}
  \caption{Model ranking and significant model differences against the number of tasks, over random task subsets. (a) Spearman's $\rho$ between the subset and full rankings. (b) Number of significantly different model pairs (Nemenyi, $\alpha = 0.05$). Legend applies to both panels.}
  \label{fig:benchmark-size}
\end{figure}

\autoref{fig:benchmark-size} shows both metrics against the number of tasks.
The median correlation with the full ranking was high for all subset sizes,
but for small subsets it varied widely with the tasks chosen,
e.g., from 0.67 to 0.99 (95\% interval) with 5 tasks around a median of 0.92.
The median number of significant pairs increased steadily with the number of tasks,
from 2 of 45 with 5 tasks to 15 of 45 with all 22 tasks.
On the 5 tasks shared with BenchNIRS,
the ranking correlated with the full ranking at $\rho = 0.78$,
and 3 of 45 model pairs differed significantly.
Thus, the main limitation of a few tasks is that few pairwise model comparisons are significant.
In addition, the model ranking can change substantially depending on which tasks are chosen.

\section{Blocked Within-Subject Cross-Validation}
\label{app:blocked-cv}

Within-subject results use stratified 5-fold cross-validation with shuffled trials,
which means that test trials can be recorded close in time to training trials.
A model could then exploit temporally correlated signals unrelated to the task, such as physiological state,
which could inflate accuracy.
Our preprocessing reduces such correlations, for example:
band-pass filtering removes drift and heart rate,
and baseline correction removes differences in signal level between trials.
However, it is still possible that the models can exploit some remaining temporal leakage.
To investigate this,
we repeated the within-subject evaluation with contiguous (blocked) splits.

We split each subject's trials into 5 consecutive blocks.
Each block was used once as the test set, with the model trained on the other four.
All other steps were the same as in the main within-subject evaluation.
Accuracy was averaged within each task over 5 seeds, subjects and folds,
and then across tasks.
As in the main tests, Shin2019 (mi) was left out of the significance tests,
because it overlaps with Shin2019 (ma).

\autoref{tab:blocked-cv} lists the mean accuracy of each task, averaged over models, under shuffled and blocked splits.
Blocked splits lowered mean accuracy from 59.9\% to 58.6\%,
and accuracy dropped in 16 of 23 tasks
(Wilcoxon signed-rank test, $W = 70$, $n = 22$, $p = 0.068$).
Four outlier tasks accounted for most of the drop,
Luke2024b (12.3 points), Herff2014 (5.8), Cockx2023 (4.2) and Ortega2020 (3.5),
compared with at most 2.1 in any other task.
Without them, the mean drop was 0.3 instead of 1.3 points,
so in most tasks the choice of split had little effect on accuracy.

In these four outlier tasks, contiguous splitting also caused the highest class imbalance in the test folds.
We measured imbalance as the total variation distance between the class distribution of each test fold and that of all the subject's trials,
averaged over folds and subjects,
which was 0.18 to 0.44 in these tasks and at most 0.15 in all others.
The largest drops coincided with the largest imbalance, which suggests that imbalance is a factor in the drop.
In Luke2024b and Cockx2023, the imbalance arises because class balance between separate recording sessions varies drastically,
which could let a model trained on shuffled splits separate the classes by differences between sessions.
In these two tasks, temporal leakage may therefore have inflated shuffled accuracy.
However, this issue is not avoided by blocked splits,
as the class differences are confounded with the sessions by the recording design,
so within-subject results on tasks with such session-dependent class balance should be interpreted with caution.

Mean accuracy of every model, averaged over tasks, dropped by 0.3 to 2.5 points,
and the size of the drop did not differ significantly between models (Friedman test, $\chi^2(9, N = 22) = 13.76$, $p = 0.13$).
The ranking of models by mean rank under blocked splits matched the main ranking
except for one swap between MRH and CNN at 6th and 7th place (Kendall's $\tau = 0.94$).
LDA remained the top-ranked model,
and the group of models not significantly different from it (Nemenyi test) was the same under both split types,
with all significant differences between models in the main results remaining significant.

Overall, changing to blocked splits lowered within-subject accuracy by about one point,
most of it in four outlier tasks,
and left the model ranking almost unchanged, with the same top group of models.
As neither shuffled nor blocked splits are better in every respect, both are available in the fNIRSAtlas repository.
Shuffled splits keep each subject's class balance in every test set and can be redrawn for each seed,
but place test trials close in time to training trials.
Blocked splits separate training and test trials in time,
but can leave test sets unbalanced and give a single partition per subject.

\begin{table}[htbp]
\centering
\caption{Mean accuracy ($\times 100$) $\pm$ standard deviation across subjects of each task under shuffled and blocked within-subject cross-validation, averaged over 10 models and 5 seeds.
The last column gives the per-subject blocked difference from shuffled.
Tasks are sorted by this difference.
$\dagger$ marks tasks with strongly session-dependent class balance.}
\label{tab:blocked-cv}
\small
\begin{tabular}{lrrr}
\toprule
Dataset & Shuffled & Blocked & Difference \\
\midrule
Luke2024b$^\dagger$ & 63.7 $\pm$ 9.1 & 51.4 $\pm$ 9.7 & $-12.3$ $\pm$ 3.6 \\
Herff2014 & 43.0 $\pm$ 5.6 & 37.3 $\pm$ 6.3 & $-5.8$ $\pm$ 3.3 \\
Cockx2023$^\dagger$ & 92.3 $\pm$ 4.7 & 88.1 $\pm$ 7.7 & $-4.2$ $\pm$ 3.4 \\
Ortega2020 & 61.2 $\pm$ 11.7 & 57.7 $\pm$ 11.3 & $-3.5$ $\pm$ 3.8 \\
Luke2024a & 45.4 $\pm$ 10.1 & 43.3 $\pm$ 10.3 & $-2.1$ $\pm$ 2.1 \\
Shin2016 (mi) & 55.7 $\pm$ 5.9 & 54.1 $\pm$ 6.8 & $-1.5$ $\pm$ 3.8 \\
Shin2019 (mi) & 64.6 $\pm$ 6.3 & 63.2 $\pm$ 7.1 & $-1.4$ $\pm$ 2.6 \\
Shader2021 & 55.1 $\pm$ 12.6 & 53.8 $\pm$ 10.8 & $-1.4$ $\pm$ 3.0 \\
Shin2018b & 75.3 $\pm$ 7.9 & 74.0 $\pm$ 8.2 & $-1.3$ $\pm$ 2.2 \\
Buccino2016 & 60.8 $\pm$ 10.0 & 59.6 $\pm$ 10.0 & $-1.2$ $\pm$ 1.7 \\
Bnci & 92.6 $\pm$ 4.4 & 91.7 $\pm$ 5.5 & $-0.9$ $\pm$ 1.2 \\
Yi2025 & 37.3 $\pm$ 3.7 & 36.4 $\pm$ 3.6 & $-0.9$ $\pm$ 1.0 \\
Shin2016 (ma) & 79.0 $\pm$ 6.6 & 78.1 $\pm$ 7.3 & $-0.9$ $\pm$ 1.6 \\
Nemo (empe) & 31.4 $\pm$ 5.6 & 30.6 $\pm$ 6.2 & $-0.8$ $\pm$ 2.8 \\
Bak2019 & 78.2 $\pm$ 14.9 & 77.6 $\pm$ 16.0 & $-0.6$ $\pm$ 2.5 \\
Shin2019 (ma) & 74.3 $\pm$ 8.1 & 73.8 $\pm$ 8.3 & $-0.6$ $\pm$ 2.2 \\
Shin2018 (wg) & 68.5 $\pm$ 9.0 & 68.6 $\pm$ 9.3 & 0.1 $\pm$ 2.5 \\
Luke2021 & 80.8 $\pm$ 14.6 & 81.6 $\pm$ 14.5 & 0.8 $\pm$ 1.2 \\
Guevara2024 & 59.3 $\pm$ 10.3 & 60.1 $\pm$ 11.8 & 0.8 $\pm$ 4.0 \\
Shin2018 (nback) & 41.4 $\pm$ 6.9 & 42.4 $\pm$ 7.7 & 1.0 $\pm$ 3.0 \\
Huang2021 & 26.8 $\pm$ 7.2 & 28.8 $\pm$ 7.4 & 2.0 $\pm$ 3.5 \\
Nemo (afim) & 23.0 $\pm$ 4.2 & 25.0 $\pm$ 5.0 & 2.0 $\pm$ 2.5 \\
Gao2024 & 68.8 $\pm$ 11.7 & 71.0 $\pm$ 11.3 & 2.2 $\pm$ 1.2 \\
\midrule
Mean & 59.9 & 58.6 & $-1.3$ \\
\bottomrule
\end{tabular}

\end{table}

\section{Compute Resources}
\label{app:compute}

Our benchmarking experiments were run on the Puhti supercomputer at CSC -- IT Center for Science, Finland. Because CPU resources were more readily available, we ran most jobs on CPU and used GPUs only for a subset of the cross-subject deep-learning runs where they were needed to keep runtimes practical. CPU runs used nodes with two 20-core Intel Xeon Gold 6230 (Cascade Lake) processors and 192--384\,GiB RAM, while GPU runs used nodes with four NVIDIA V100 GPUs (32\,GiB each) and 384\,GiB RAM. In total, the reported paper experiments used about 30.9k CPU core-hours and 0.9k GPU-hours. By model family, this included about 0.15k hours for LDA, 0.49k hours for ANN, 0.43k hours for CNN, 3.7k CPU core-hours for LSTM, 0.63k hours for EEGNet, 2.9k CPU core-hours for DCN, 0.15k hours for MRH, 16.5k CPU core-hours plus 0.9k GPU-hours for fNIRS-T, 0.38k hours for fNIRSNet23, and 5.6k CPU core-hours plus 0.04k GPU-hours for fNIRSNet24.

\begin{table}[htbp]
\centering
\caption{Compute usage for the paper models.}
\label{tab:compute_usage}
\small
\begin{tabular}{llrr}
\toprule
Resource & Model & Jobs & Hours \\
\midrule
CPU & LDA & 690 & 146 \\
CPU & ANN & 690 & 486 \\
CPU & CNN & 690 & 425 \\
CPU & LSTM & 690 & 3733 \\
CPU & EEGNet & 690 & 634 \\
CPU & DCN & 690 & 2891 \\
CPU & MRH & 690 & 148 \\
CPU & fNIRS-T & 490 & 16450 \\
GPU & fNIRS-T & 200 & 889 \\
CPU & fNIRSNet23 & 690 & 376 \\
CPU & fNIRSNet24 & 665 & 5577 \\
GPU & fNIRSNet24 & 25 & 37 \\
\midrule
\multicolumn{2}{l}{Total jobs} & 6900 & \\
\multicolumn{3}{l}{Total CPU core-hours} & 30867 \\
\multicolumn{3}{l}{Total GPU-hours} & 925 \\
\bottomrule
\end{tabular}
\end{table}

\section{Impact of Temporal and Spatial Aggregation}
\label{app:resolution}

\begin{figure}[htbp]
\centering
\includegraphics[width=\textwidth]{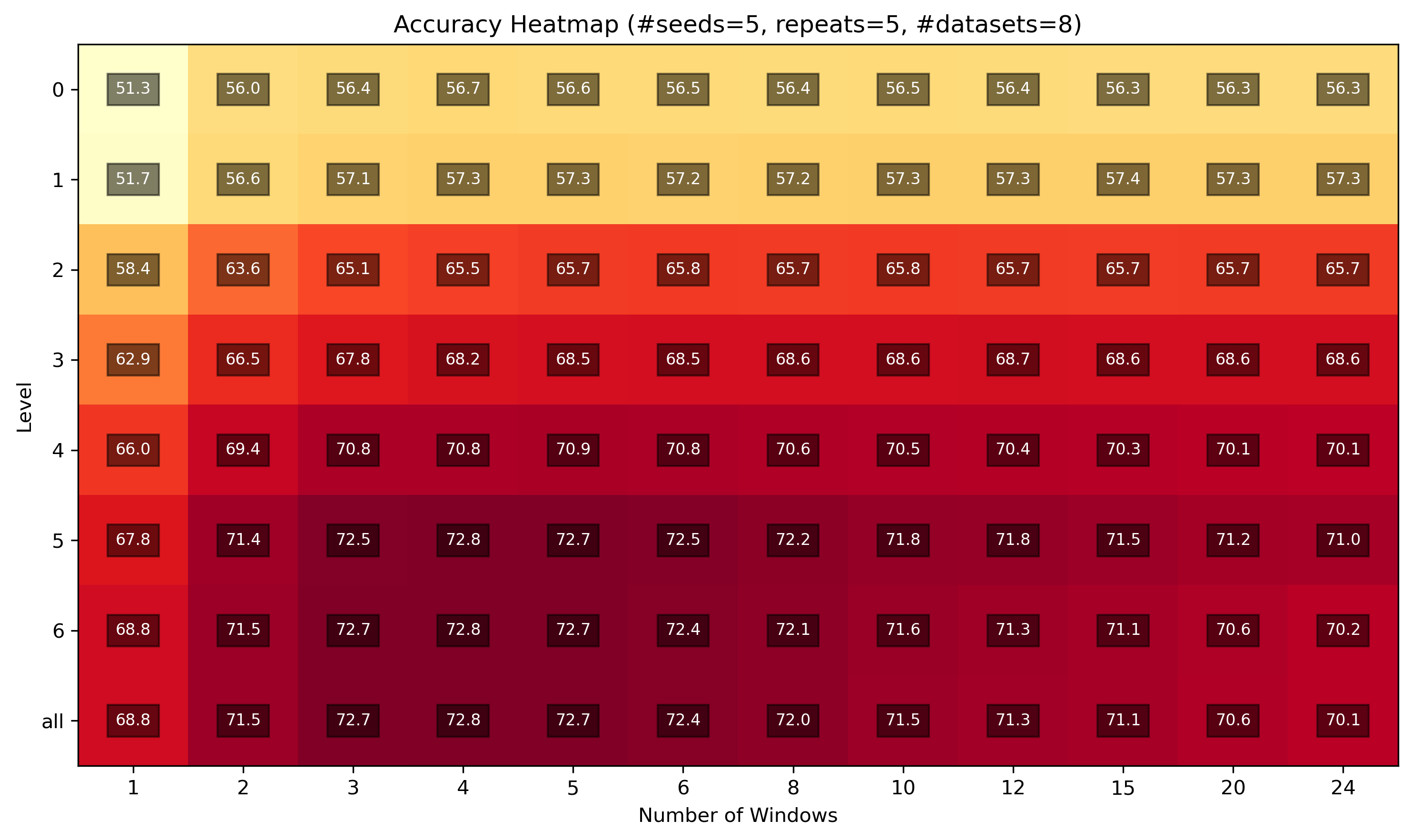}
\caption{Heatmap of cross-subject cross-validation accuracy across aggregation levels and number of windows for fNIRS datasets.}
\label{fig:resolution}
\end{figure}

\autoref{fig:resolution} illustrates the impact of temporal and spatial aggregation on cross-subject classification accuracy. A clear trend emerges with spatial aggregation: as aggregation increases, accuracy decreases. For temporal aggregation, accuracy peaks at four windows across most spatial aggregation levels.

Spatial aggregation levels include:
\begin{itemize}
  \item 0 (all channels grouped together),
  \item 1 (grouped by channel type),
  \item 2--6 (spatial aggregation)
  \item `all' (no aggregation), in practice this is almost identical to level 6 for most tasks.
\end{itemize}
The analysis was conducted on the most high-performing tasks (bnci, bak2019, luke2021, buccino2016, shin2016--ma, shin2019--ma, shin2018b, cockx2023) using LDA.
Results were averaged over 5 seeds and 5 repeats. 

\begin{figure}[htbp]
\centering
\includegraphics[width=\textwidth]{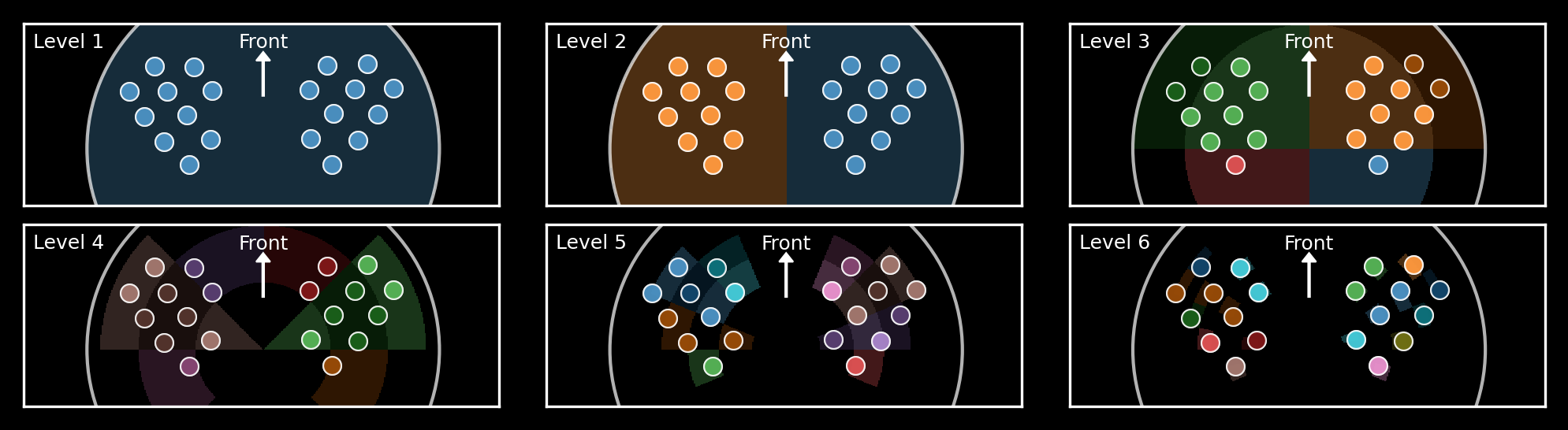}
    \caption{Illustration of spatial aggregation.}
\label{fig:spatial_aggregation_levels}
\end{figure}

\section{Repetitions}
\label{app:repeats}

\begin{figure}[htbp]
\centering
\includegraphics[width=0.99\textwidth]{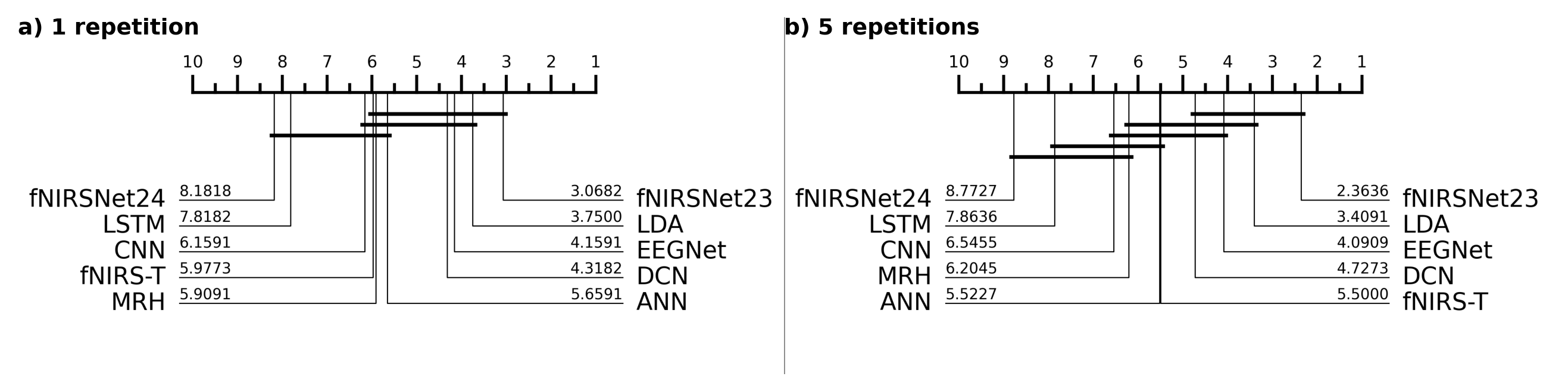}
\caption{
    Cross-subject critical difference diagrams from only the first CV repetition (a) and averaged over 5 repetitions (b). Horizontal bars connect models that are not significantly different under the Nemenyi post-hoc test ($\alpha = 0.05$).
}
\label{fig:repeats-cd}
\end{figure}

To obtain a stable model comparison in the main results, we repeated CV 5 times with independently randomized splits and model initializations.
\autoref{fig:repeats-cd} compares the cross-subject critical difference diagram from a single repetition with the one averaged over 5 repetitions. With 5 repetitions, the average ranks spread further apart and the models separate into smaller groups, while the order of the highest-ranked models is unchanged.
\autoref{tab:repeats-pairs} shows that, over all combinations of repetitions, the number of significantly different model pairs increases with the number of repetitions in the cross-subject setting and depends less on which repetitions are used in both settings.
As most of this gain is reached with 3 repetitions, we recommend averaging over at least 3 CV repetitions when comparing models on fNIRSAtlas.

\begin{table}[htbp]
\centering
\caption{
    Number of significantly different model pairs (Nemenyi, $\alpha = 0.05$) out of 45, as a function of the number of CV repetitions. Values are the mean (min--max) over all combinations of that many repetitions out of 5.
}
\label{tab:repeats-pairs}
\small
\begin{tabular}{ccc}
\toprule
Repetitions & Cross-subject & Within-subject \\
\midrule
1 & 11.6 (10--13) & 16.2 (13--19) \\
2 & 13.6 (12--16) & 16.6 (15--18) \\
3 & 14.7 (13--16) & 16.3 (15--18) \\
4 & 15.0 (14--16) & 16.4 (15--18) \\
5 & 15 & 17 \\
\bottomrule
\end{tabular}
\end{table}

\section{Validation Kappa for All Hyperparameters}
\label{app:hparams}

This appendix shows how validation $\kappa$ changes across the hyperparameter grids.
Each heatmap shows the full grid-search landscape of one model.
Each cell shows validation $\kappa \times 100$ for one task and hyperparameter setting, and the bottom row shows the mean across tasks.
The top row shows the number of parameters (\#Params) of each setting, counted on a synthetic input with the shape of Bak2019 (1500 samples, 2 classes, 40 channels, 76 time points).
The value in parentheses below \#Params is the measured storage size of these parameters.

Each heatmap is followed by an ordinary least squares fit of validation $\kappa$ on the hyperparameters, each entered as a categorical predictor, pooled over tasks and CV repetitions.
The intercept is the fitted validation $\kappa \times 100$ at the reference level of every hyperparameter, and each other coefficient is the difference in mean validation $\kappa \times 100$ between a level and the reference level of its hyperparameter.

\clearpage
\begin{figure}[H]
\centering
    \includegraphics[width=0.95\textwidth]{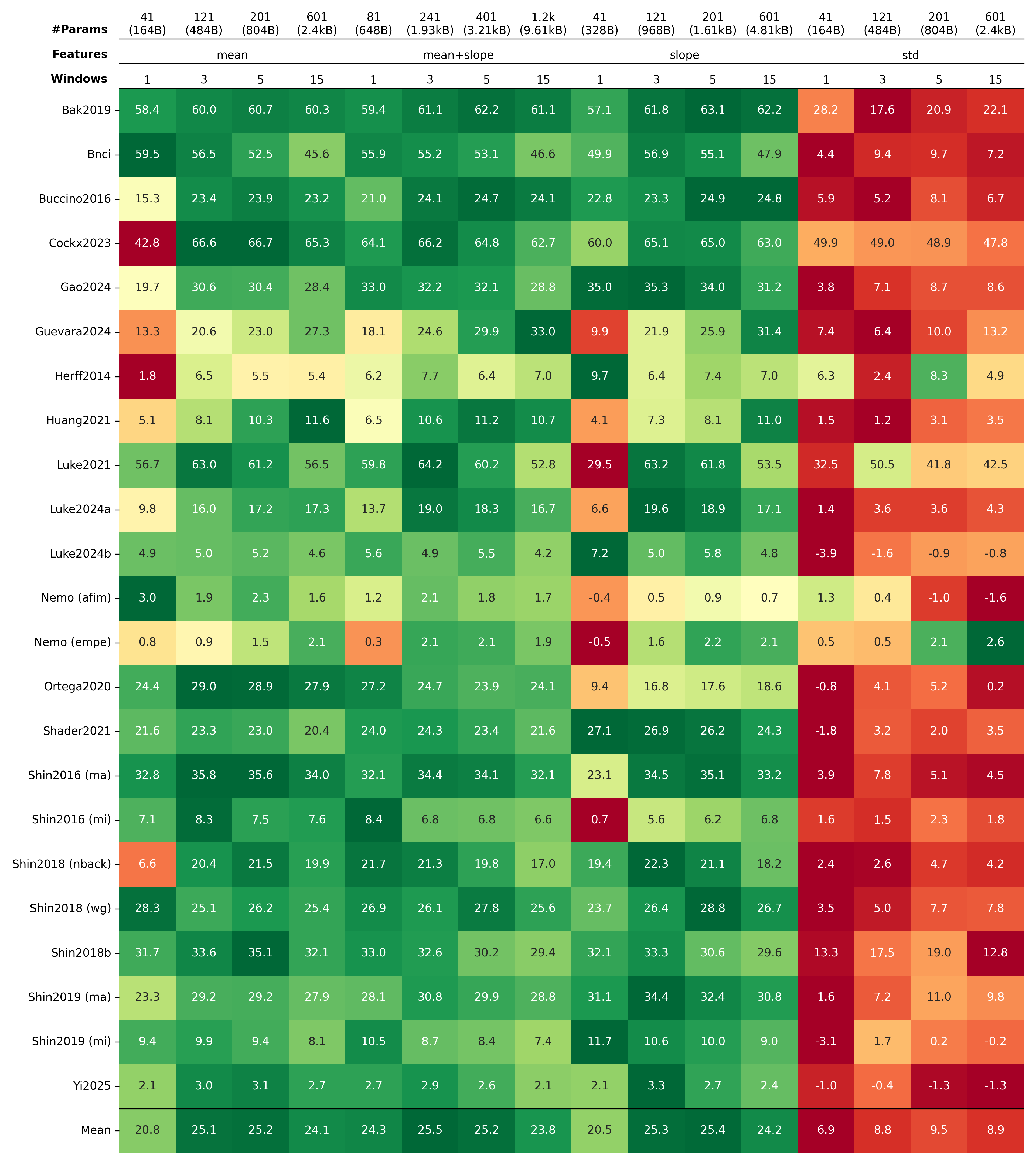}
\caption{Cross-subject validation $\kappa$ across the full hyperparameter grid for LDA.}
    \label{fig:lda_paramsearch_cross}
\end{figure}
\begin{table}[htbp]
\centering
\caption{Coefficient summary for the LDA hyperparameter analysis (cross-subject).}
\label{tab:hparam_lm_cross_lda}
\small
\begin{tabular}{l c}
\toprule
Term & Coef. \\
\midrule
Intercept (Features = mean, Windows = 1) & 21.7 \\
Features = mean+slope & 0.9 \\
Features = slope & 0.0 \\
Features = std & -15.3 \\
Windows = 3 & 3.0 \\
Windows = 5 & 3.2 \\
Windows = 15 & 2.1 \\
\bottomrule
\end{tabular}
\end{table}

\clearpage
\begin{figure}[H]
\centering
    \includegraphics[width=0.95\textwidth]{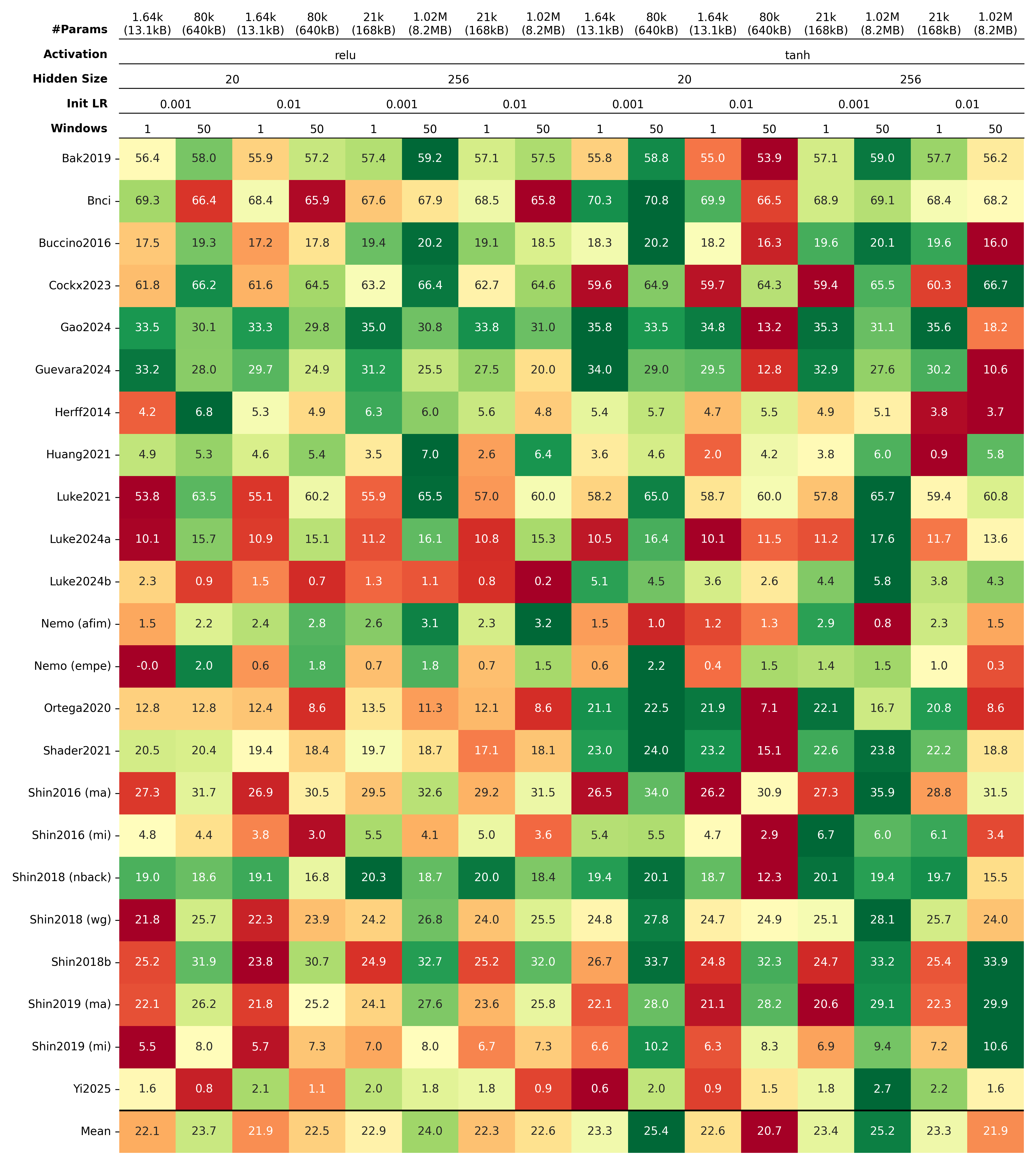}
\caption{Cross-subject validation $\kappa$ across the full hyperparameter grid for ANN.}
    \label{fig:ann_paramsearch_cross}
\end{figure}
\begin{table}[htbp]
\centering
\caption{Coefficient summary for the ANN hyperparameter analysis (cross-subject).}
\label{tab:hparam_lm_cross_ann}
\small
\begin{tabular}{l c}
\toprule
Term & Coef. \\
\midrule
Intercept (Activation = relu, Hidden Size = 20, Init LR = 0.001, Windows = 1) & 23.0 \\
Activation = tanh & 0.5 \\
Hidden Size = 256 & 0.4 \\
Init LR = 0.01 & -1.5 \\
Windows = 50 & 0.5 \\
\bottomrule
\end{tabular}
\end{table}

\clearpage
\begin{figure}[H]
\centering
    \includegraphics[width=0.95\textwidth]{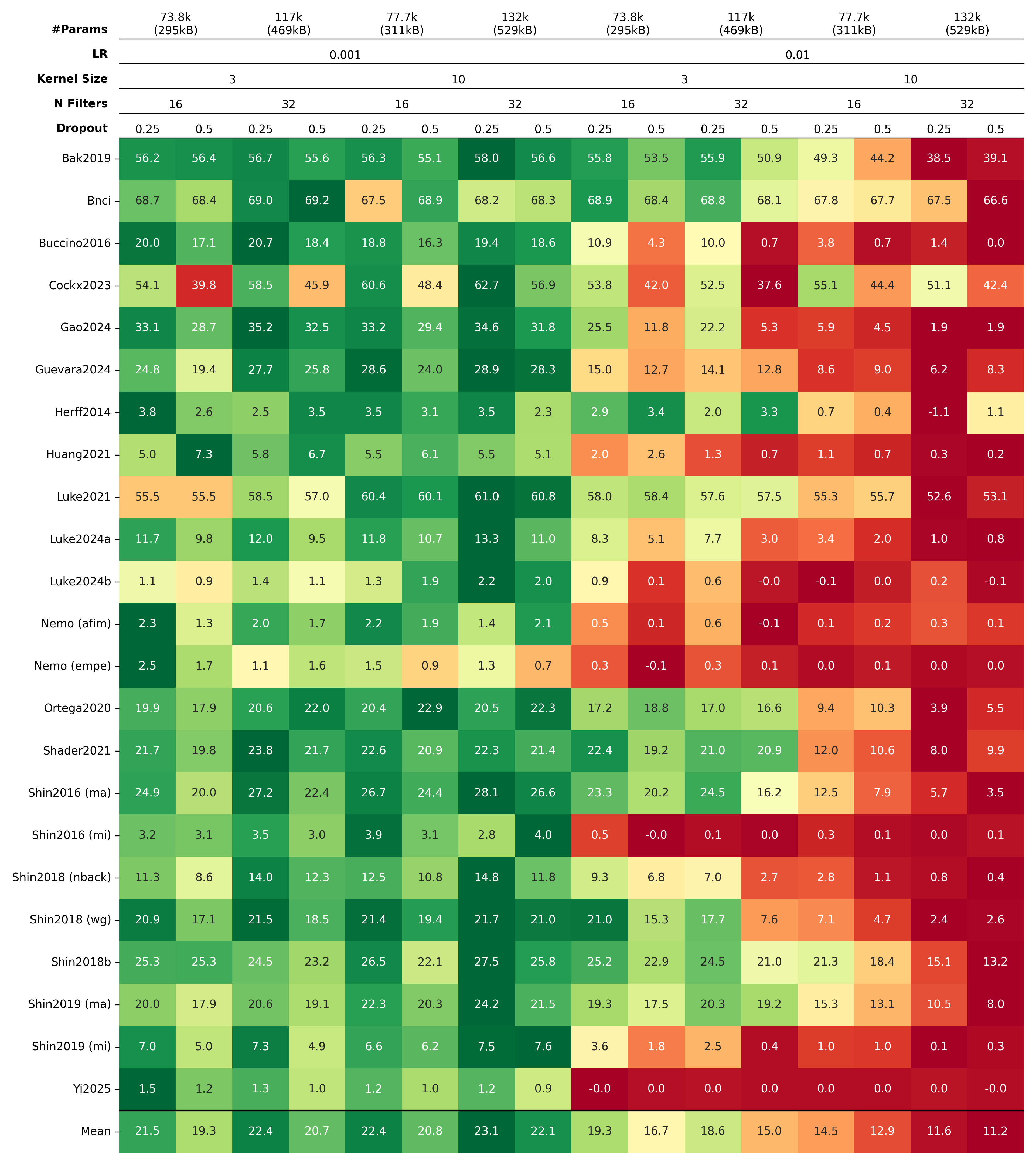}
\caption{Cross-subject validation $\kappa$ across the full hyperparameter grid for CNN.}
    \label{fig:cnn_paramsearch_cross}
\end{figure}
\begin{table}[htbp]
\centering
\caption{Coefficient summary for the CNN hyperparameter analysis (cross-subject).}
\label{tab:hparam_lm_cross_cnn2a}
\small
\begin{tabular}{l c}
\toprule
Term & Coef. \\
\midrule
Intercept (Kernel Size = 3, Learning Rate = 0.001, N Filters = 16, Dropout = 0.25) & 23.6 \\
Kernel Size = 10 & -1.9 \\
Learning Rate = 0.01 & -6.6 \\
N Filters = 32 & -0.4 \\
Dropout = 0.5 & -1.8 \\
\bottomrule
\end{tabular}
\end{table}

\clearpage
\begin{figure}[H]
\centering
    \includegraphics[width=0.95\textwidth]{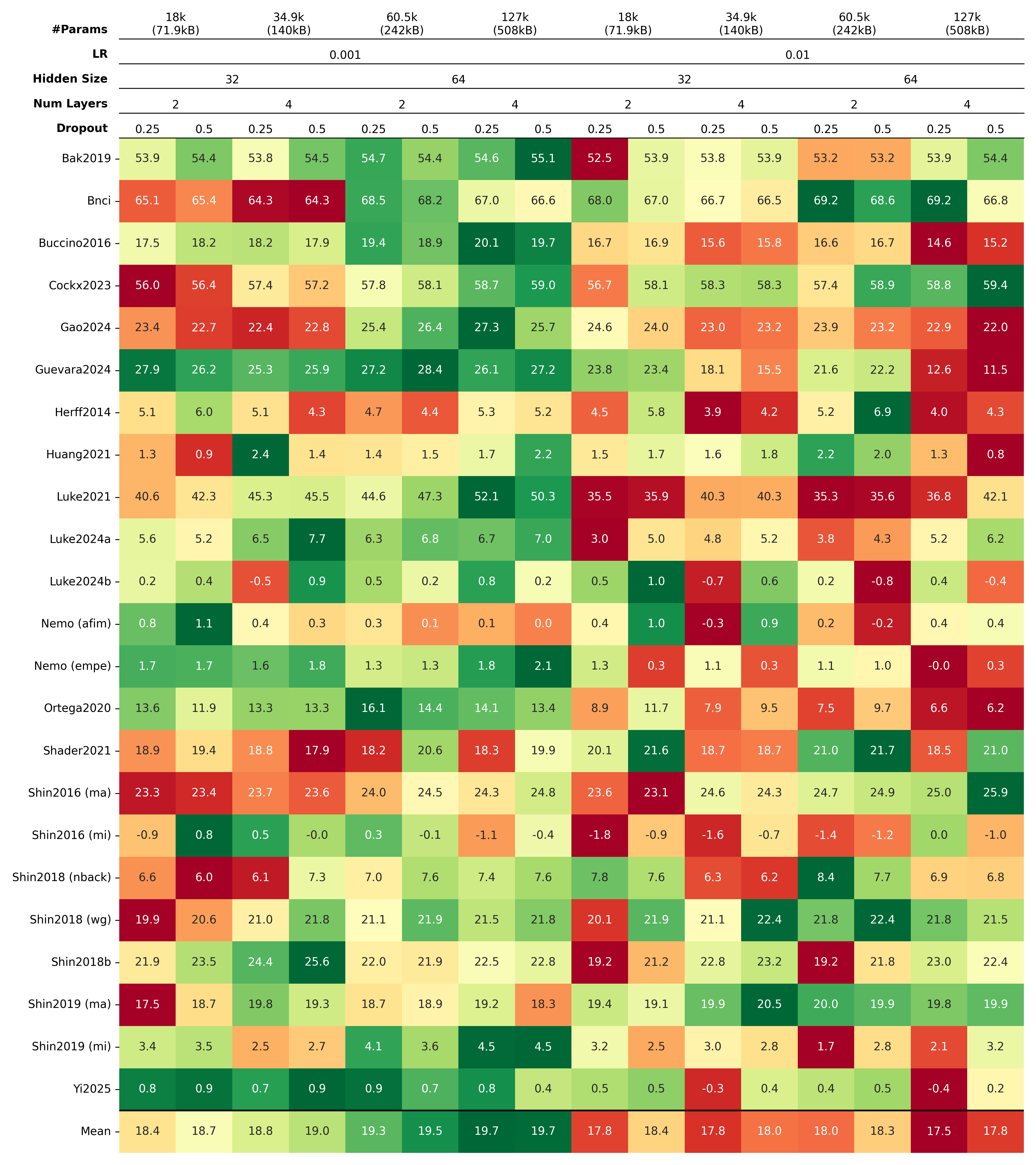}
\caption{Cross-subject validation $\kappa$ across the full hyperparameter grid for LSTM.}
    \label{fig:lstm_paramsearch_cross}
\end{figure}
\begin{table}[htbp]
\centering
\caption{Coefficient summary for the LSTM hyperparameter analysis (cross-subject).}
\label{tab:hparam_lm_cross_lstm}
\small
\begin{tabular}{l c}
\toprule
Term & Coef. \\
\midrule
Intercept (Hidden Size = 32, Learning Rate = 0.001, Num Layers = 2, Dropout = 0.25) & 18.8 \\
Hidden Size = 64 & 0.4 \\
Learning Rate = 0.01 & -1.2 \\
Num Layers = 4 & 0.0 \\
Dropout = 0.5 & 0.2 \\
\bottomrule
\end{tabular}
\end{table}

\clearpage
\begin{figure}[H]
\centering
    \includegraphics[width=0.95\textwidth]{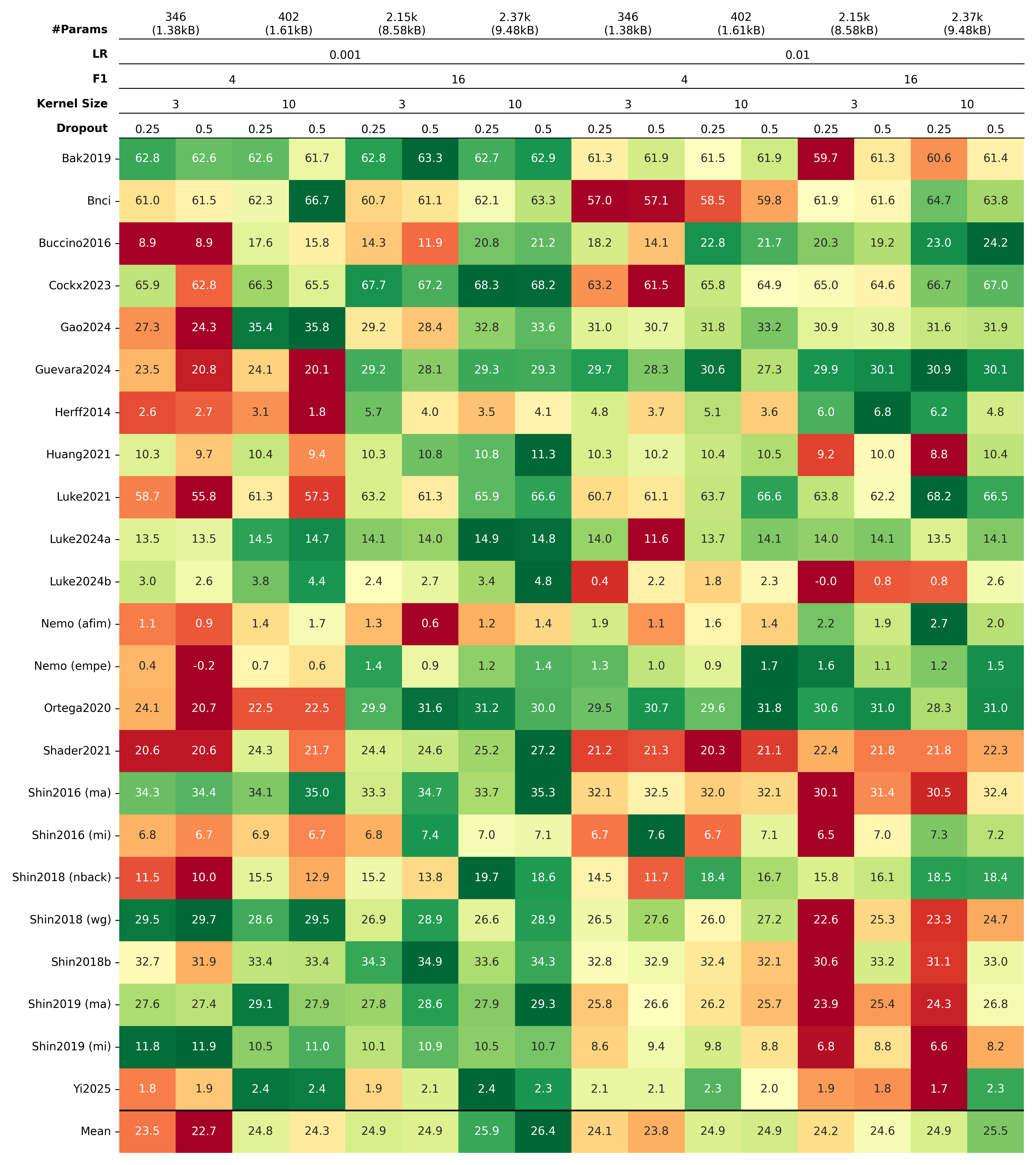}
\caption{Cross-subject validation $\kappa$ across the full hyperparameter grid for EEGNet.}
    \label{fig:eegnet_paramsearch_cross}
\end{figure}
\begin{table}[htbp]
\centering
\caption{Coefficient summary for the EEGNet hyperparameter analysis (cross-subject).}
\label{tab:hparam_lm_cross_eegnet}
\small
\begin{tabular}{l c}
\toprule
Term & Coef. \\
\midrule
Intercept (F1 = 4, Kernel Size = 3, Learning Rate = 0.001, Dropout = 0.25) & 23.6 \\
F1 = 16 & 1.0 \\
Kernel Size = 10 & 1.1 \\
Learning Rate = 0.01 & -0.1 \\
Dropout = 0.5 & 0.0 \\
\bottomrule
\end{tabular}
\end{table}

\clearpage
\begin{figure}[H]
\centering
    \includegraphics[width=0.95\textwidth]{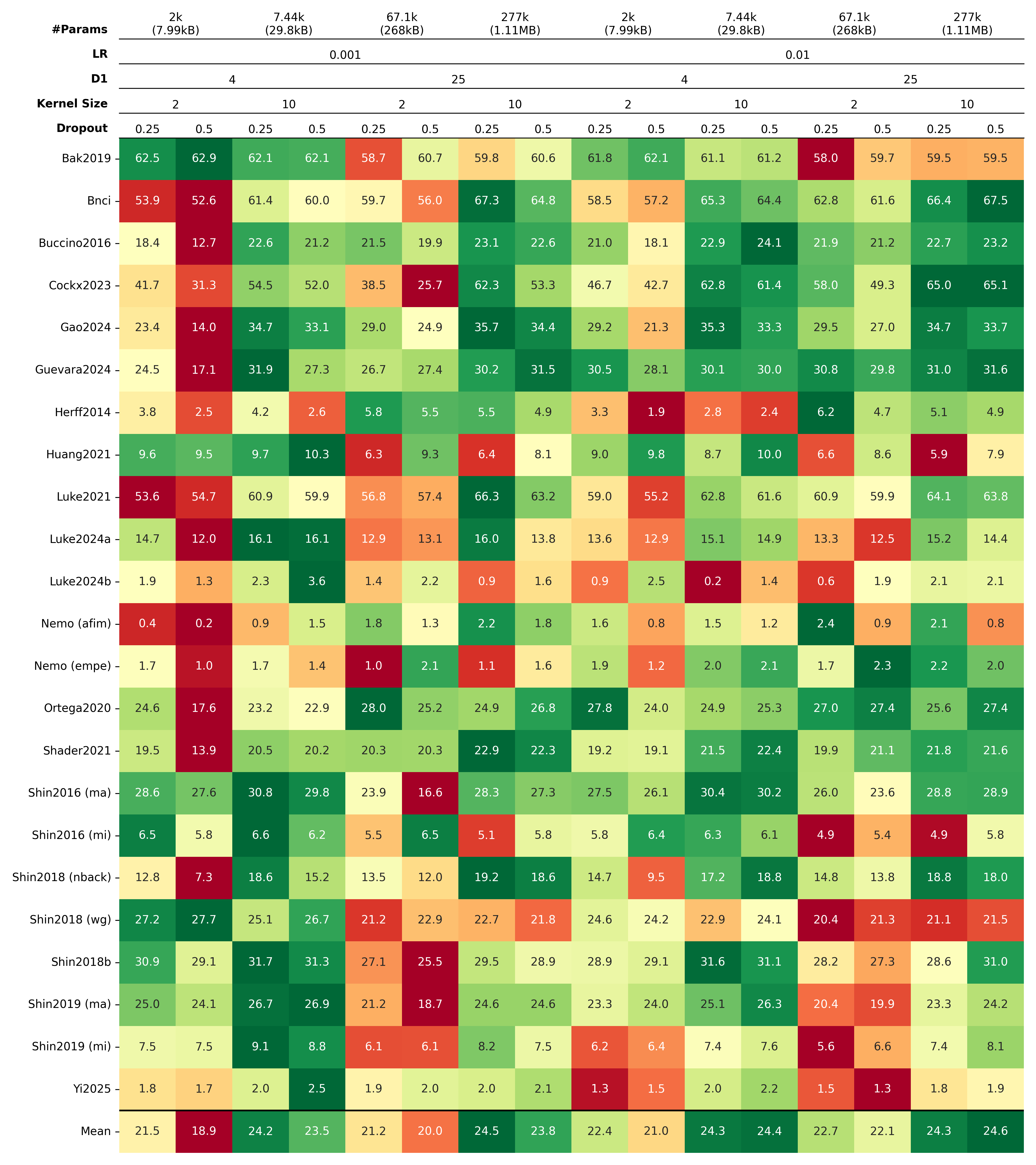}
\caption{Cross-subject validation $\kappa$ across the full hyperparameter grid for DCN.}
    \label{fig:dcn_paramsearch_cross}
\end{figure}
\begin{table}[htbp]
\centering
\caption{Coefficient summary for the DCN hyperparameter analysis (cross-subject).}
\label{tab:hparam_lm_cross_dcn}
\small
\begin{tabular}{l c}
\toprule
Term & Coef. \\
\midrule
Intercept (D1 = 4, Kernel Size = 2, Learning Rate = 0.001, Dropout = 0.25) & 21.0 \\
D1 = 25 & 0.3 \\
Kernel Size = 10 & 3.0 \\
Learning Rate = 0.01 & 1.0 \\
Dropout = 0.5 & -0.9 \\
\bottomrule
\end{tabular}
\end{table}

\clearpage
\begin{figure}[H]
\centering
    \includegraphics[width=0.95\textwidth]{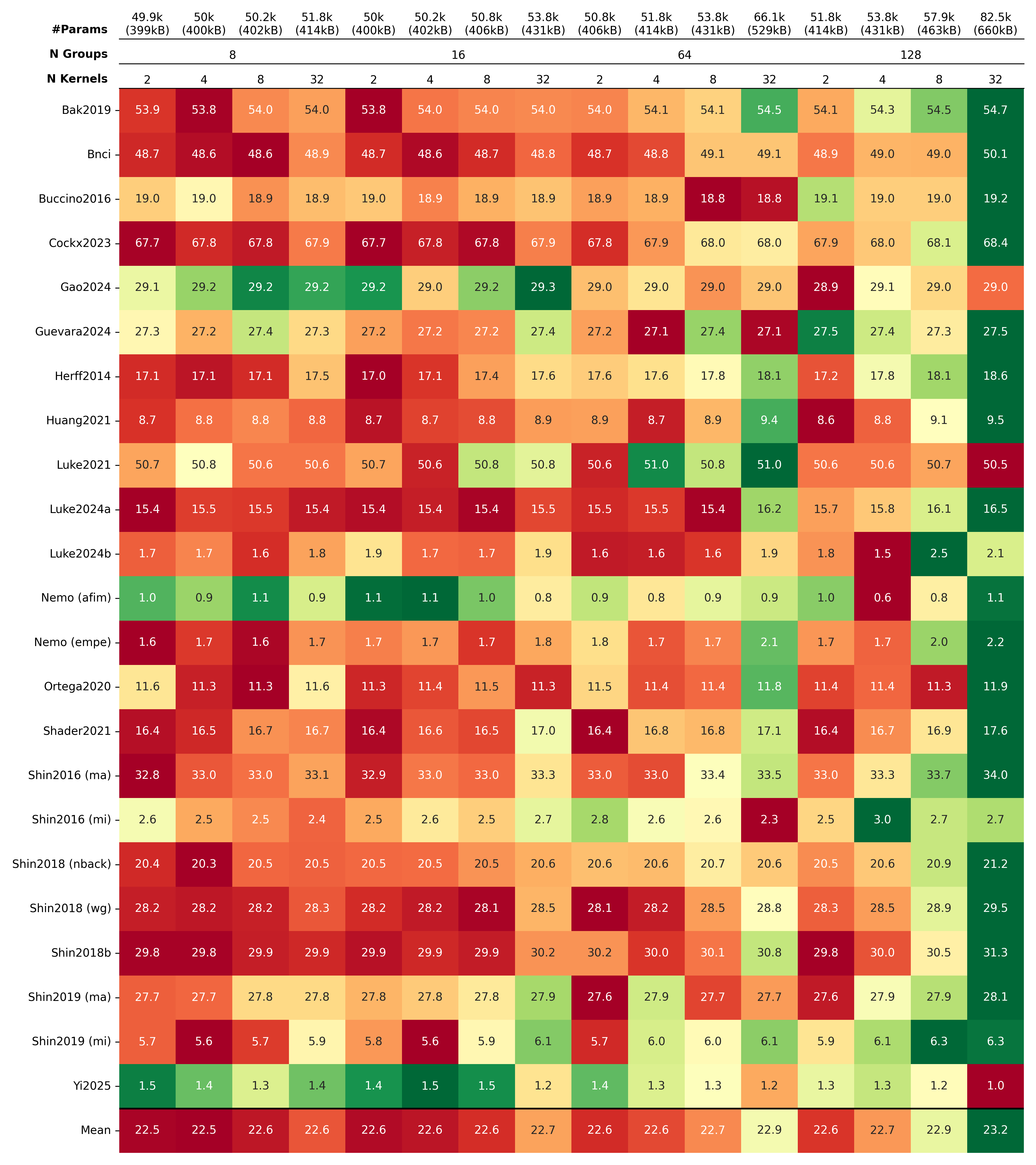}
\caption{Cross-subject validation $\kappa$ across the full hyperparameter grid for MRH.}
    \label{fig:mrh_paramsearch_cross}
\end{figure}
\begin{table}[htbp]
\centering
\caption{Coefficient summary for the MRH hyperparameter analysis (cross-subject).}
\label{tab:hparam_lm_cross_aeon_multirockethydra}
\small
\begin{tabular}{l c}
\toprule
Term & Coef. \\
\midrule
Intercept (N Groups = 8, N Kernels = 2) & 22.5 \\
N Groups = 16 & 0.0 \\
N Groups = 64 & 0.1 \\
N Groups = 128 & 0.3 \\
N Kernels = 4 & 0.0 \\
N Kernels = 8 & 0.1 \\
N Kernels = 32 & 0.3 \\
\bottomrule
\end{tabular}
\end{table}

\clearpage
\begin{figure}[H]
\centering
    \includegraphics[width=0.95\textwidth]{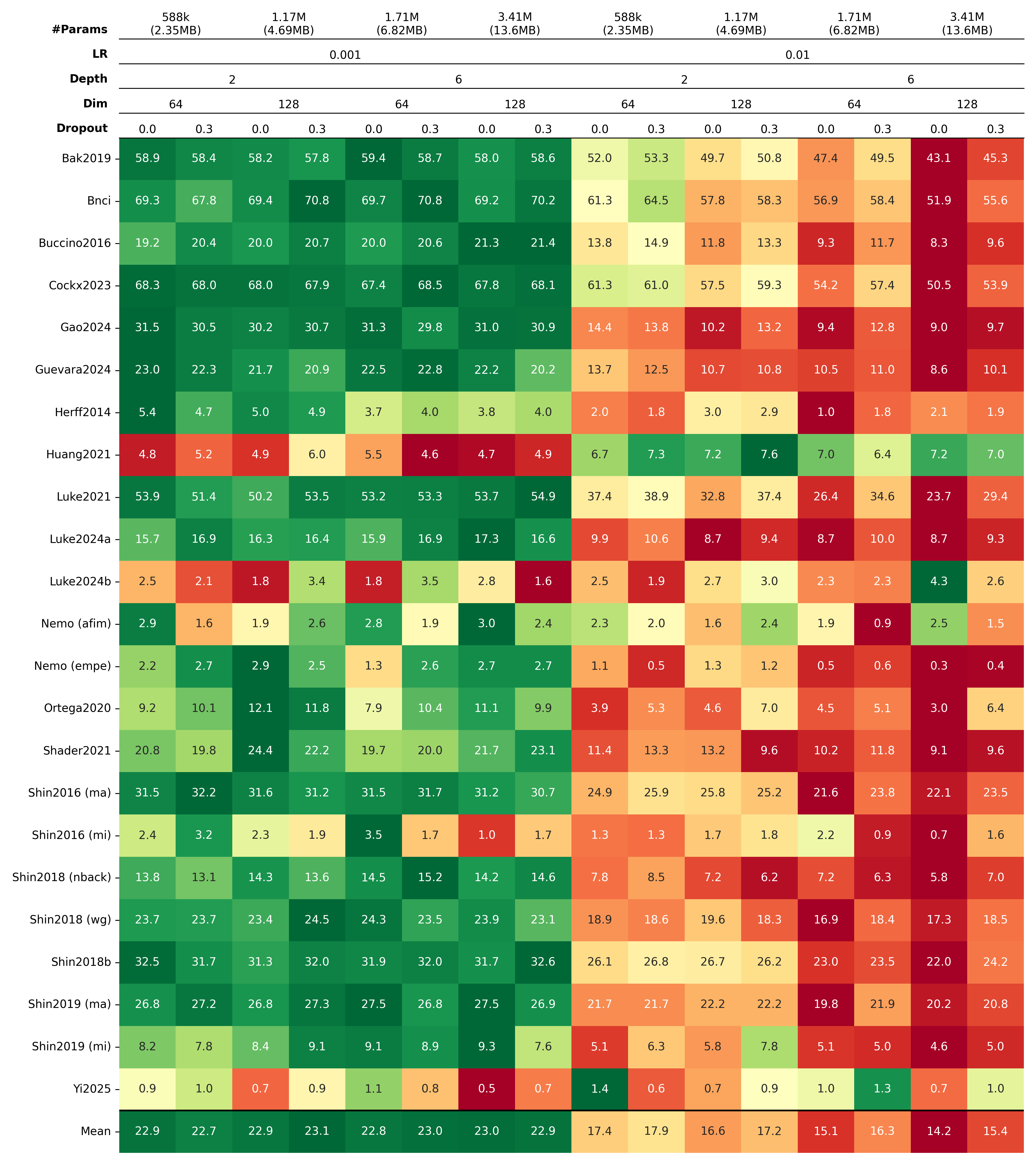}
\caption{Cross-subject validation $\kappa$ across the full hyperparameter grid for fNIRS-T.}
    \label{fig:fnirst_paramsearch_cross}
\end{figure}
\begin{table}[htbp]
\centering
\caption{Coefficient summary for the fNIRS-T hyperparameter analysis (cross-subject).}
\label{tab:hparam_lm_cross_fnirst}
\small
\begin{tabular}{l c}
\toprule
Term & Coef. \\
\midrule
Intercept (Depth = 2, Dim = 64, Dropout = 0.0, Learning Rate = 0.001) & 23.4 \\
Depth = 6 & -1.0 \\
Dim = 128 & -0.4 \\
Dropout = 0.3 & 0.4 \\
Learning Rate = 0.01 & -6.7 \\
\bottomrule
\end{tabular}
\end{table}

\clearpage
\begin{figure}[H]
\centering
    \includegraphics[width=0.95\textwidth]{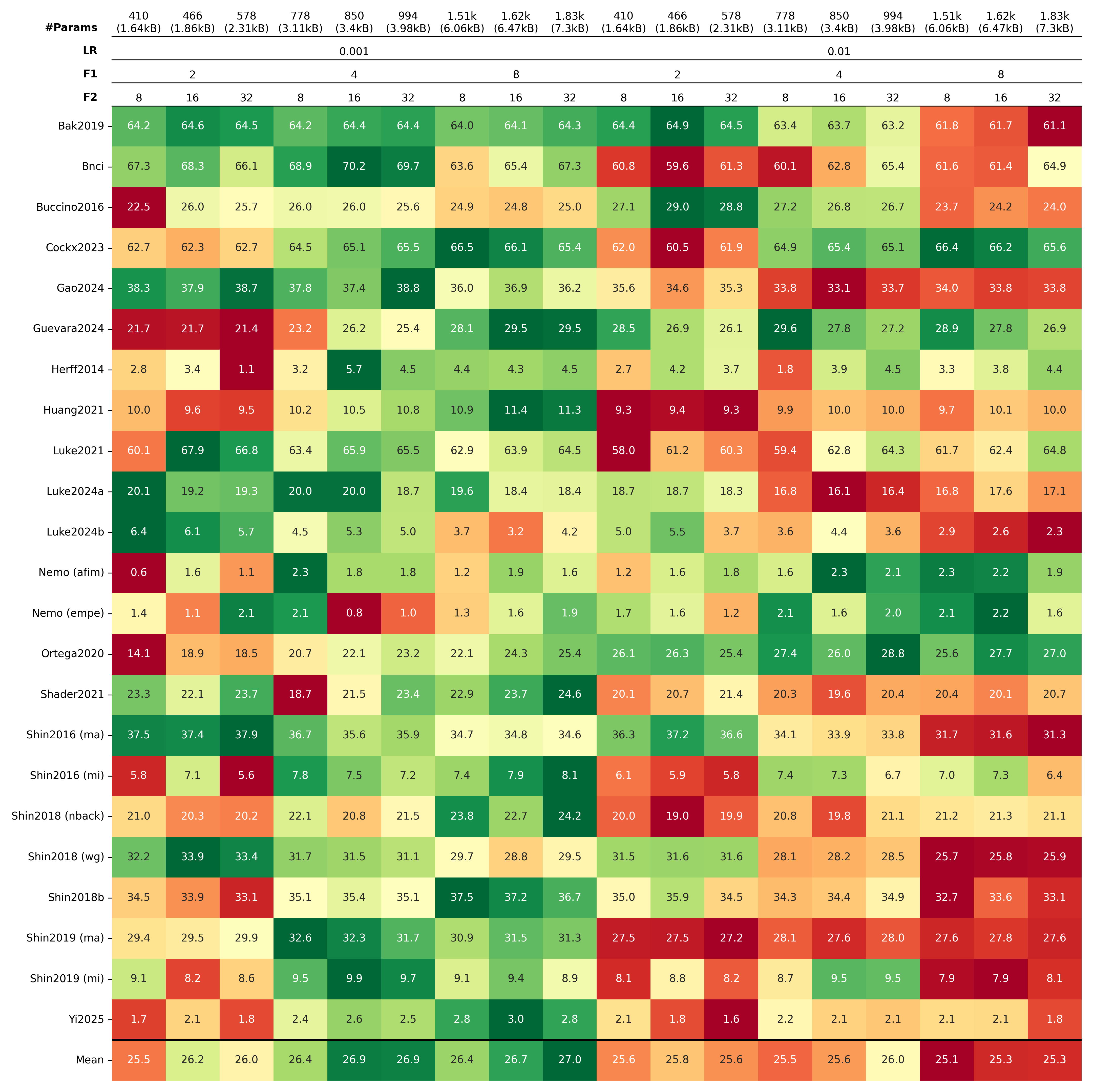}
\caption{Cross-subject validation $\kappa$ across the full hyperparameter grid for fNIRSNet23.}
    \label{fig:fnirsnet23_paramsearch_cross}
\end{figure}
\begin{table}[htbp]
\centering
\caption{Coefficient summary for the fNIRSNet23 hyperparameter analysis (cross-subject).}
\label{tab:hparam_lm_cross_fnirsnet23}
\small
\begin{tabular}{l c}
\toprule
Term & Coef. \\
\midrule
Intercept (F1 = 2, F2 = 8, Learning Rate = 0.001) & 26.0 \\
F1 = 4 & 0.4 \\
F1 = 8 & 0.2 \\
F2 = 16 & 0.3 \\
F2 = 32 & 0.4 \\
Learning Rate = 0.01 & -0.9 \\
\bottomrule
\end{tabular}
\end{table}

\clearpage
\begin{figure}[H]
\centering
    \includegraphics[width=0.95\textwidth]{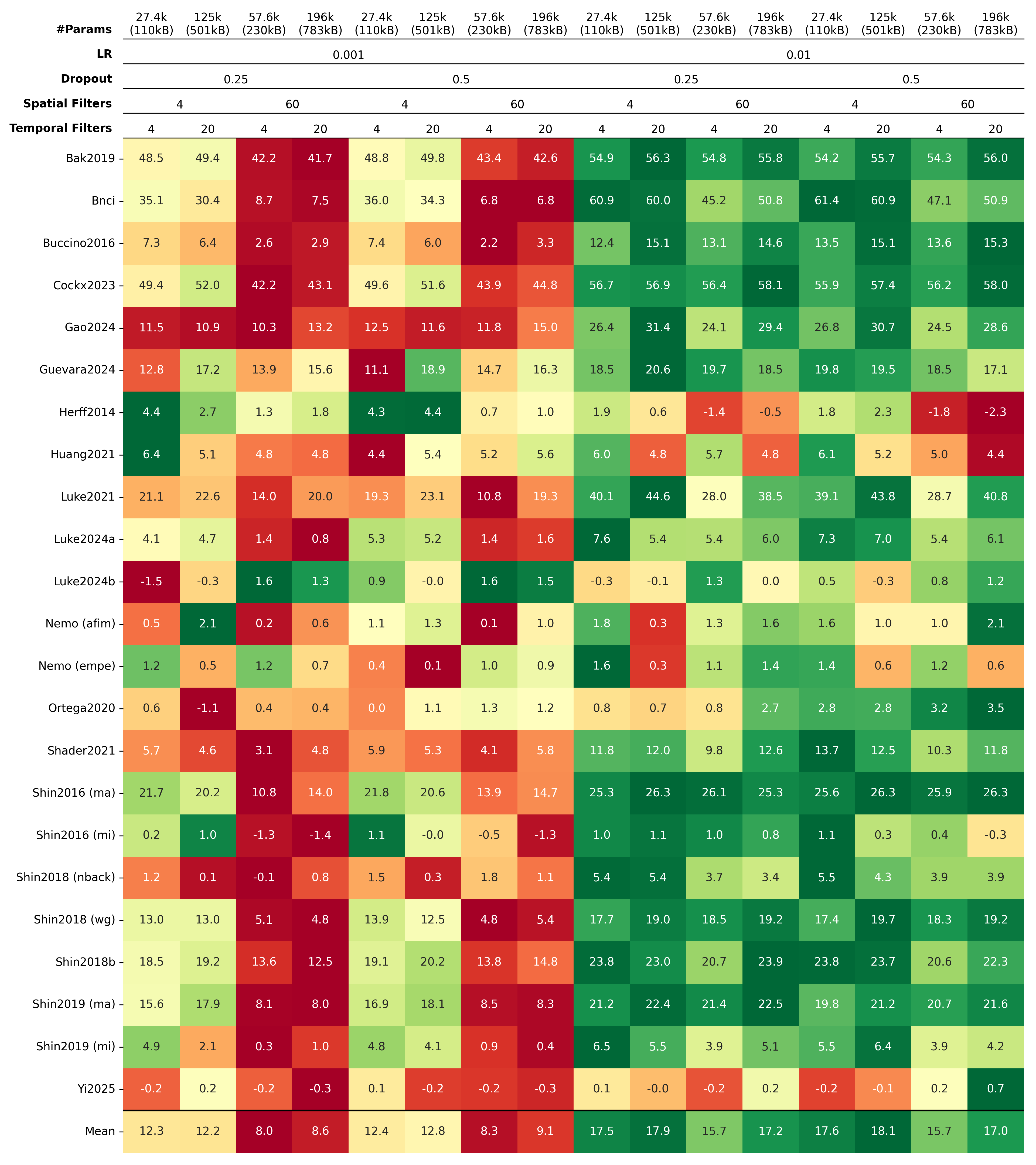}
\caption{Cross-subject validation $\kappa$ across the full hyperparameter grid for fNIRSNet24.}
    \label{fig:fnirsnet24_paramsearch_cross}
\end{figure}
\begin{table}[htbp]
\centering
\caption{Coefficient summary for the fNIRSNet24 hyperparameter analysis (cross-subject).}
\label{tab:hparam_lm_cross_fnirsnet24}
\small
\begin{tabular}{l c}
\toprule
Term & Coef. \\
\midrule
Intercept (Learning Rate = 0.001, Dropout = 0.25, Spatial Filters = 4, Temporal Filters = 4) & 11.3 \\
Learning Rate = 0.01 & 6.6 \\
Dropout = 0.5 & 0.2 \\
Spatial Filters = 60 & -2.6 \\
Temporal Filters = 20 & 0.7 \\
\bottomrule
\end{tabular}
\end{table}

\clearpage
\begin{figure}[H]
\centering
    \includegraphics[width=0.95\textwidth]{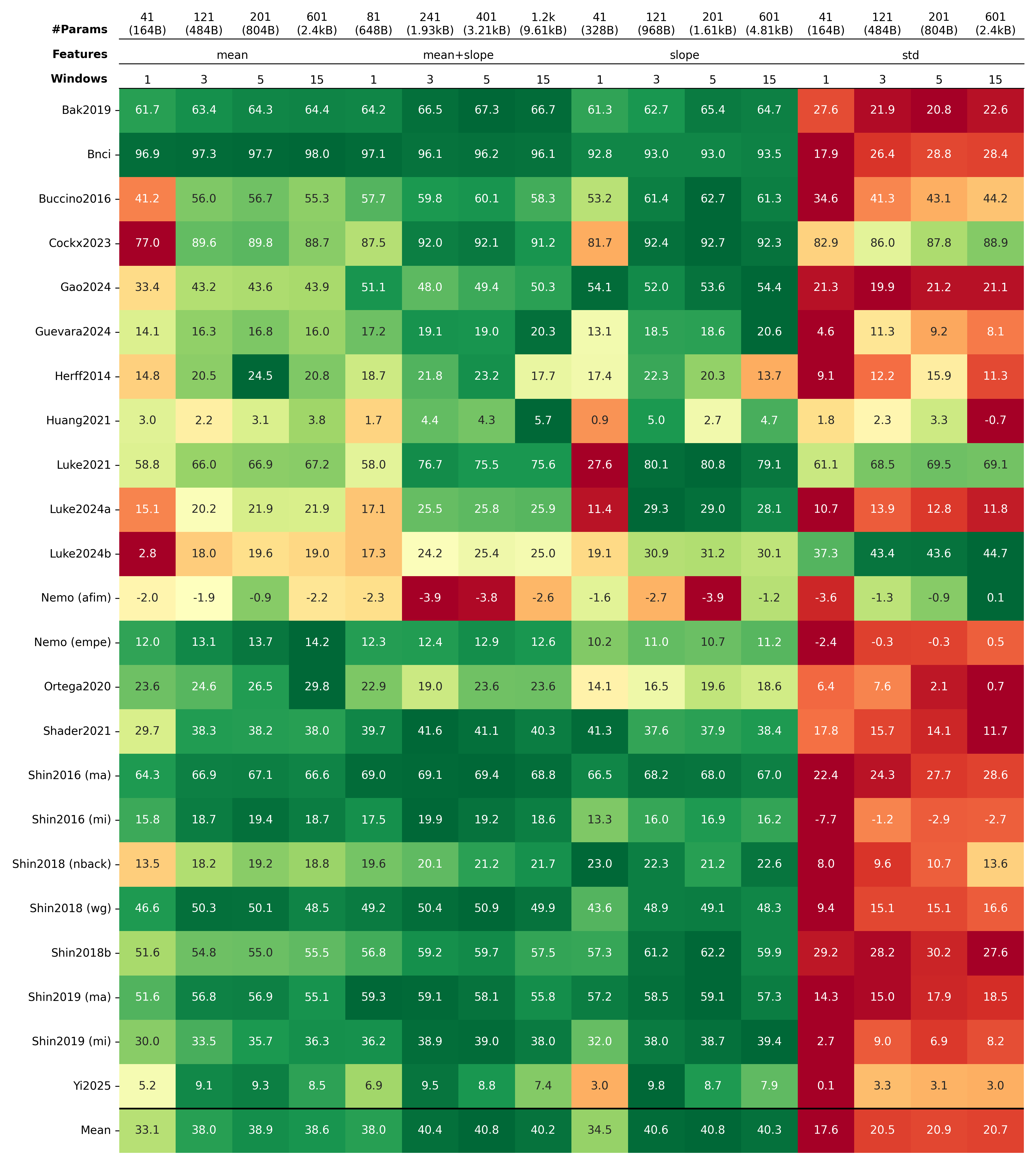}
\caption{Within-subject validation $\kappa$ across the full hyperparameter grid for LDA.}
    \label{fig:lda_paramsearch_within}
\end{figure}
\begin{table}[htbp]
\centering
\caption{Coefficient summary for the LDA hyperparameter analysis (within-subject).}
\label{tab:hparam_lm_within_lda}
\small
\begin{tabular}{l c}
\toprule
Term & Coef. \\
\midrule
Intercept (Features = mean, Windows = 1) & 33.9 \\
Features = mean+slope & 2.7 \\
Features = slope & 1.9 \\
Features = std & -17.2 \\
Windows = 3 & 4.1 \\
Windows = 5 & 4.5 \\
Windows = 15 & 4.1 \\
\bottomrule
\end{tabular}
\end{table}

\clearpage
\begin{figure}[H]
\centering
    \includegraphics[width=0.95\textwidth]{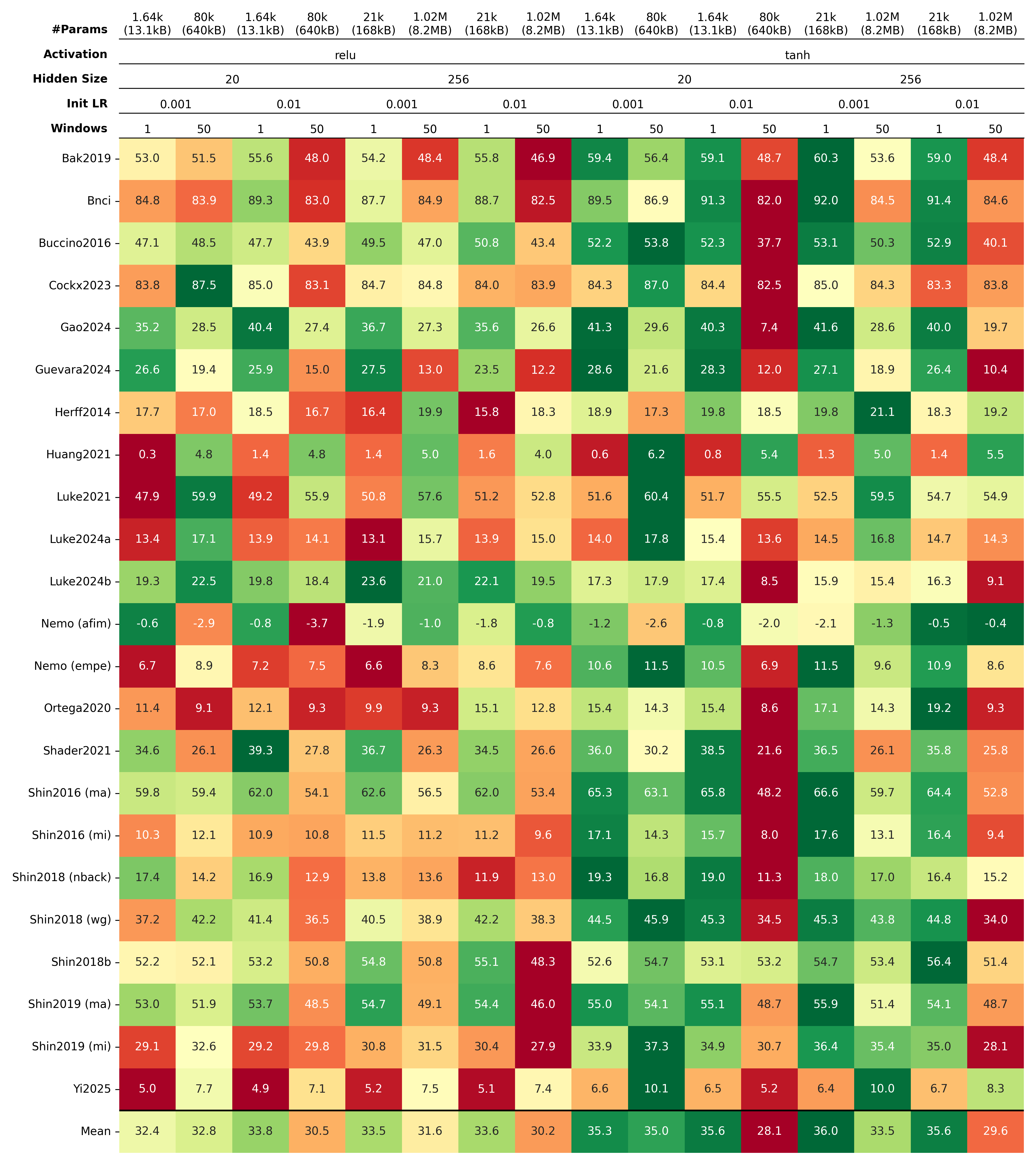}
\caption{Within-subject validation $\kappa$ across the full hyperparameter grid for ANN.}
    \label{fig:ann_paramsearch_within}
\end{figure}
\begin{table}[htbp]
\centering
\caption{Coefficient summary for the ANN hyperparameter analysis (within-subject).}
\label{tab:hparam_lm_within_ann}
\small
\begin{tabular}{l c}
\toprule
Term & Coef. \\
\midrule
Intercept (Activation = relu, Hidden Size = 20, Init LR = 0.001, Windows = 1) & 34.6 \\
Activation = tanh & 1.3 \\
Hidden Size = 256 & 0.0 \\
Init LR = 0.01 & -1.6 \\
Windows = 50 & -3.1 \\
\bottomrule
\end{tabular}
\end{table}

\clearpage
\begin{figure}[H]
\centering
    \includegraphics[width=0.95\textwidth]{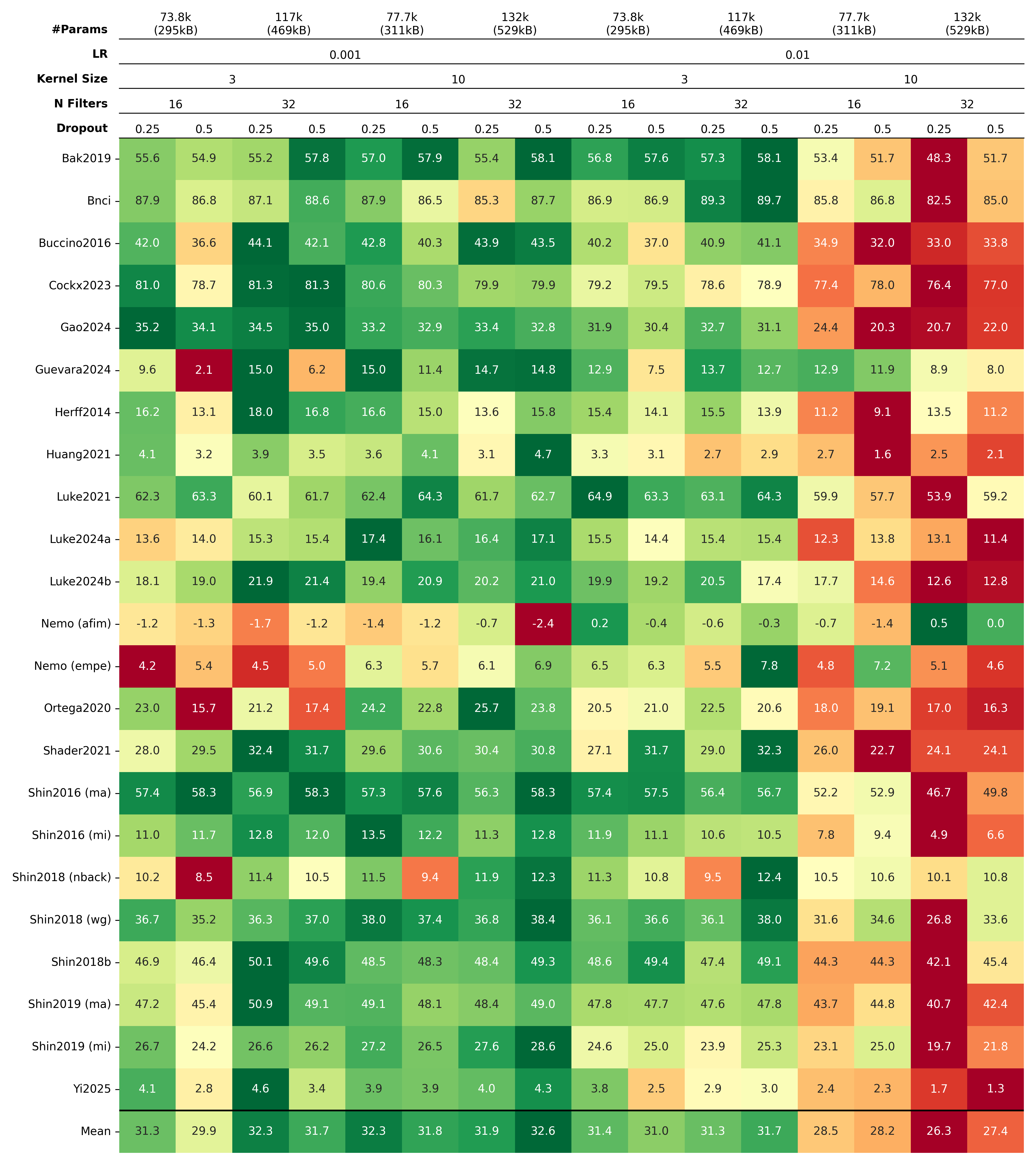}
\caption{Within-subject validation $\kappa$ across the full hyperparameter grid for CNN.}
    \label{fig:cnn_paramsearch_within}
\end{figure}
\begin{table}[htbp]
\centering
\caption{Coefficient summary for the CNN hyperparameter analysis (within-subject).}
\label{tab:hparam_lm_within_cnn2a}
\small
\begin{tabular}{l c}
\toprule
Term & Coef. \\
\midrule
Intercept (Kernel Size = 3, Learning Rate = 0.001, N Filters = 16, Dropout = 0.25) & 32.5 \\
Kernel Size = 10 & -1.4 \\
Learning Rate = 0.01 & -2.2 \\
N Filters = 32 & 0.1 \\
Dropout = 0.5 & -0.1 \\
\bottomrule
\end{tabular}
\end{table}

\clearpage
\begin{figure}[H]
\centering
    \includegraphics[width=0.95\textwidth]{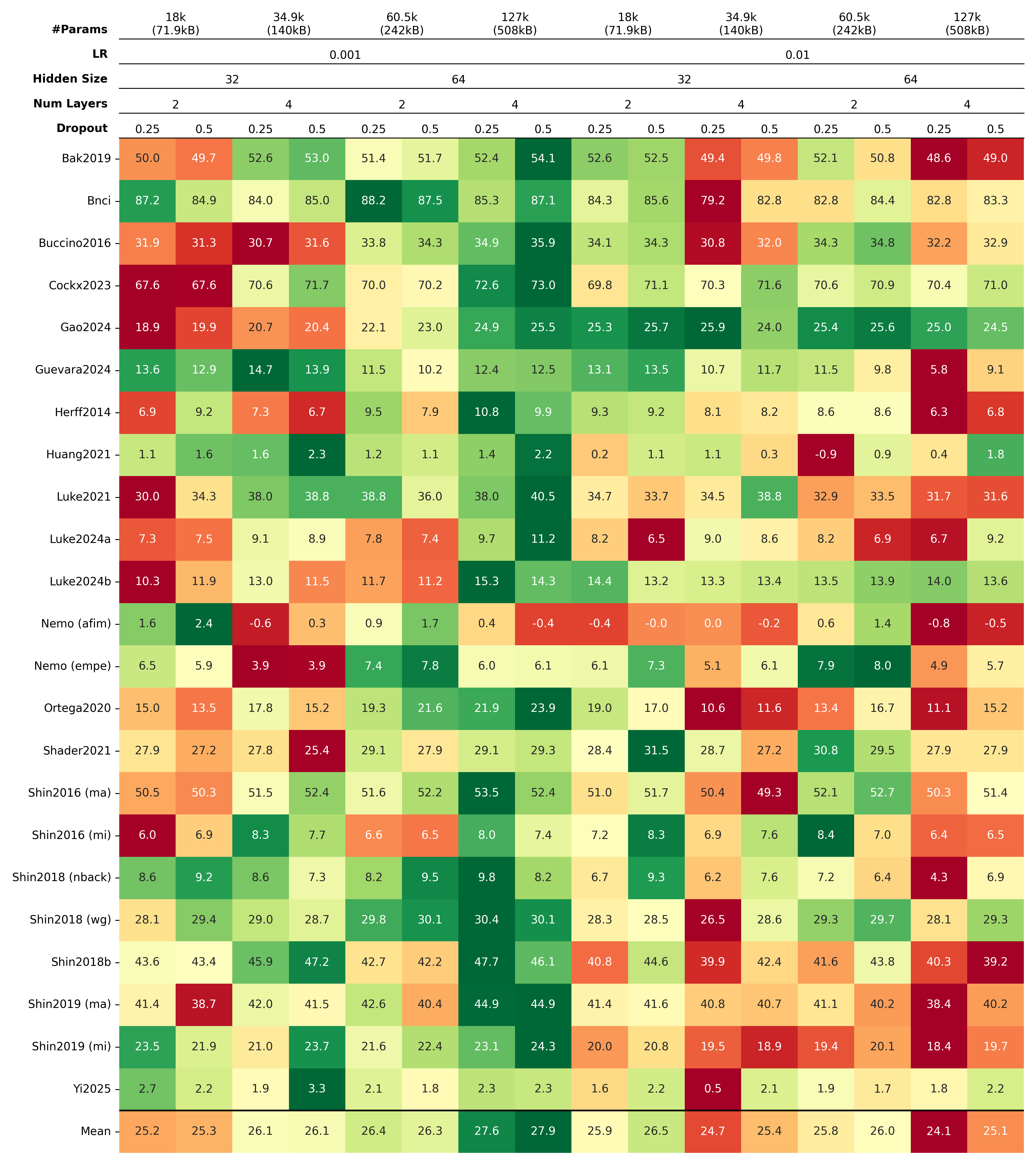}
\caption{Within-subject validation $\kappa$ across the full hyperparameter grid for LSTM.}
    \label{fig:lstm_paramsearch_within}
\end{figure}
\begin{table}[htbp]
\centering
\caption{Coefficient summary for the LSTM hyperparameter analysis (within-subject).}
\label{tab:hparam_lm_within_lstm}
\small
\begin{tabular}{l c}
\toprule
Term & Coef. \\
\midrule
Intercept (Hidden Size = 32, Learning Rate = 0.001, Num Layers = 2, Dropout = 0.25) & 26.0 \\
Hidden Size = 64 & 0.5 \\
Learning Rate = 0.01 & -0.9 \\
Num Layers = 4 & -0.1 \\
Dropout = 0.5 & 0.3 \\
\bottomrule
\end{tabular}
\end{table}

\clearpage
\begin{figure}[H]
\centering
    \includegraphics[width=0.95\textwidth]{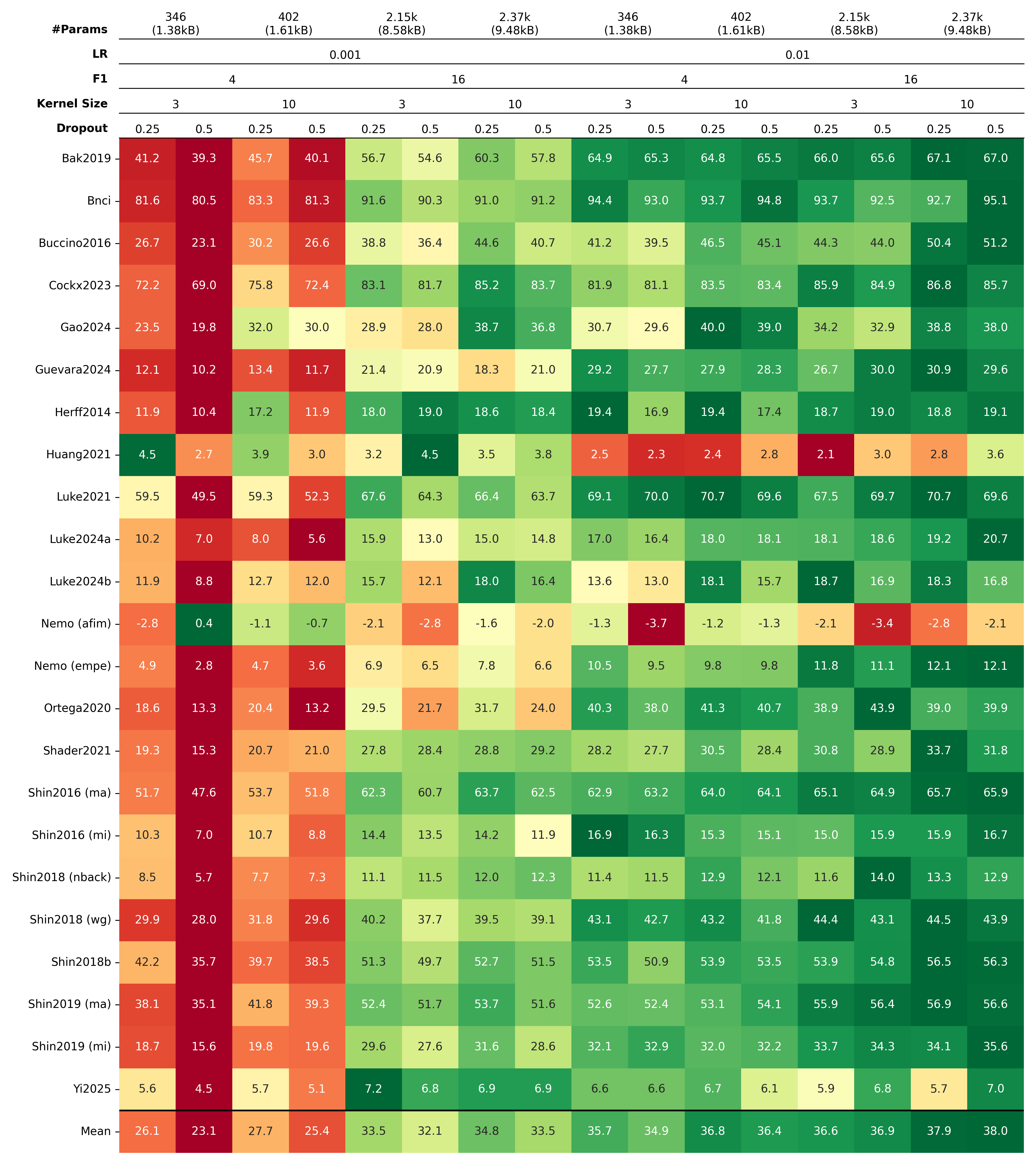}
\caption{Within-subject validation $\kappa$ across the full hyperparameter grid for EEGNet.}
    \label{fig:eegnet_paramsearch_within}
\end{figure}
\begin{table}[htbp]
\centering
\caption{Coefficient summary for the EEGNet hyperparameter analysis (within-subject).}
\label{tab:hparam_lm_within_eegnet}
\small
\begin{tabular}{l c}
\toprule
Term & Coef. \\
\midrule
Intercept (F1 = 4, Kernel Size = 3, Learning Rate = 0.001, Dropout = 0.25) & 27.0 \\
F1 = 16 & 4.6 \\
Kernel Size = 10 & 1.4 \\
Learning Rate = 0.01 & 7.1 \\
Dropout = 0.5 & -1.1 \\
\bottomrule
\end{tabular}
\end{table}

\clearpage
\begin{figure}[H]
\centering
    \includegraphics[width=0.95\textwidth]{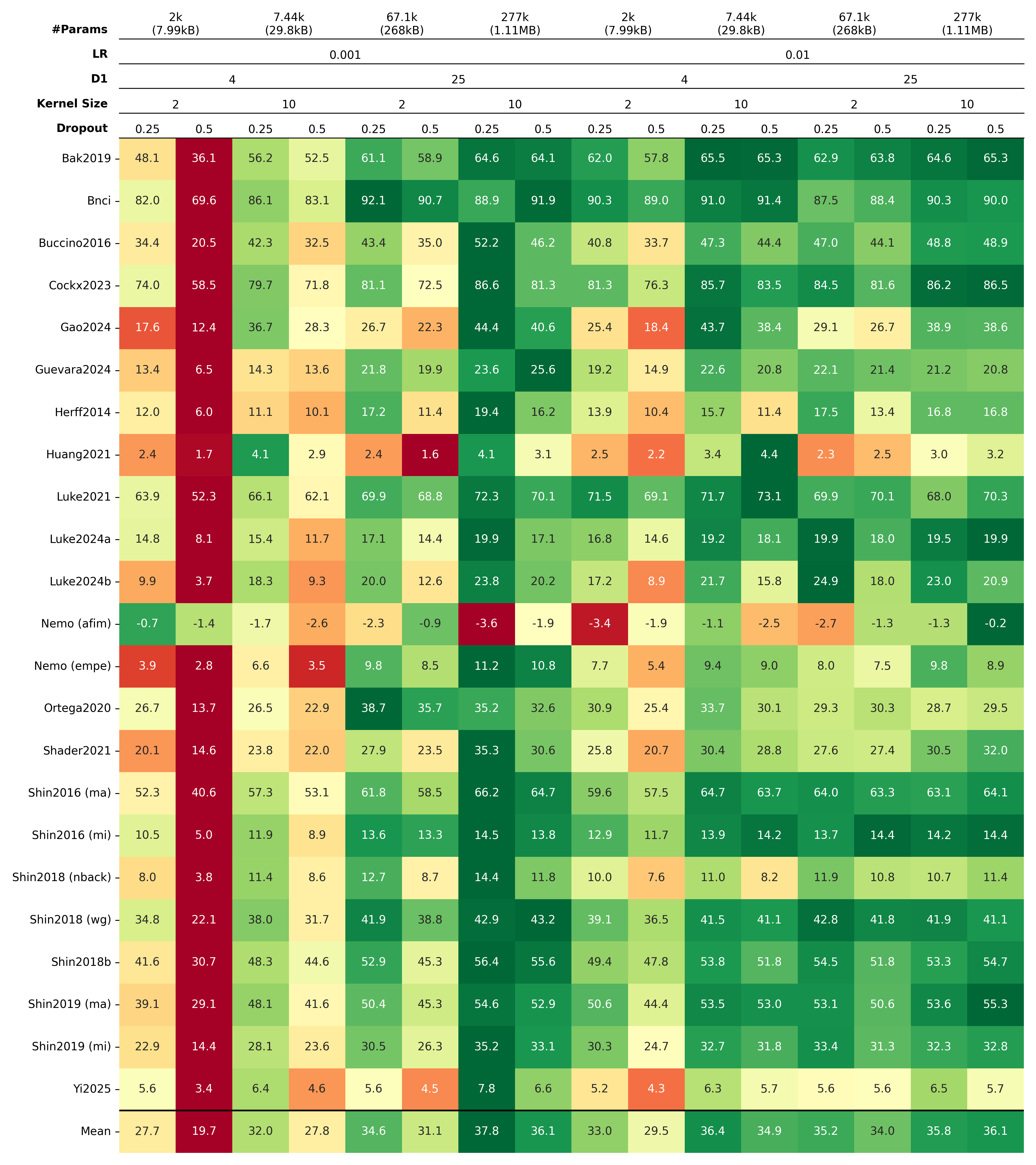}
\caption{Within-subject validation $\kappa$ across the full hyperparameter grid for DCN.}
    \label{fig:dcn_paramsearch_within}
\end{figure}
\begin{table}[htbp]
\centering
\caption{Coefficient summary for the DCN hyperparameter analysis (within-subject).}
\label{tab:hparam_lm_within_dcn}
\small
\begin{tabular}{l c}
\toprule
Term & Coef. \\
\midrule
Intercept (D1 = 4, Kernel Size = 2, Learning Rate = 0.001, Dropout = 0.25) & 27.8 \\
D1 = 25 & 5.0 \\
Kernel Size = 10 & 4.0 \\
Learning Rate = 0.01 & 3.5 \\
Dropout = 0.5 & -2.9 \\
\bottomrule
\end{tabular}
\end{table}

\clearpage
\begin{figure}[H]
\centering
    \includegraphics[width=0.95\textwidth]{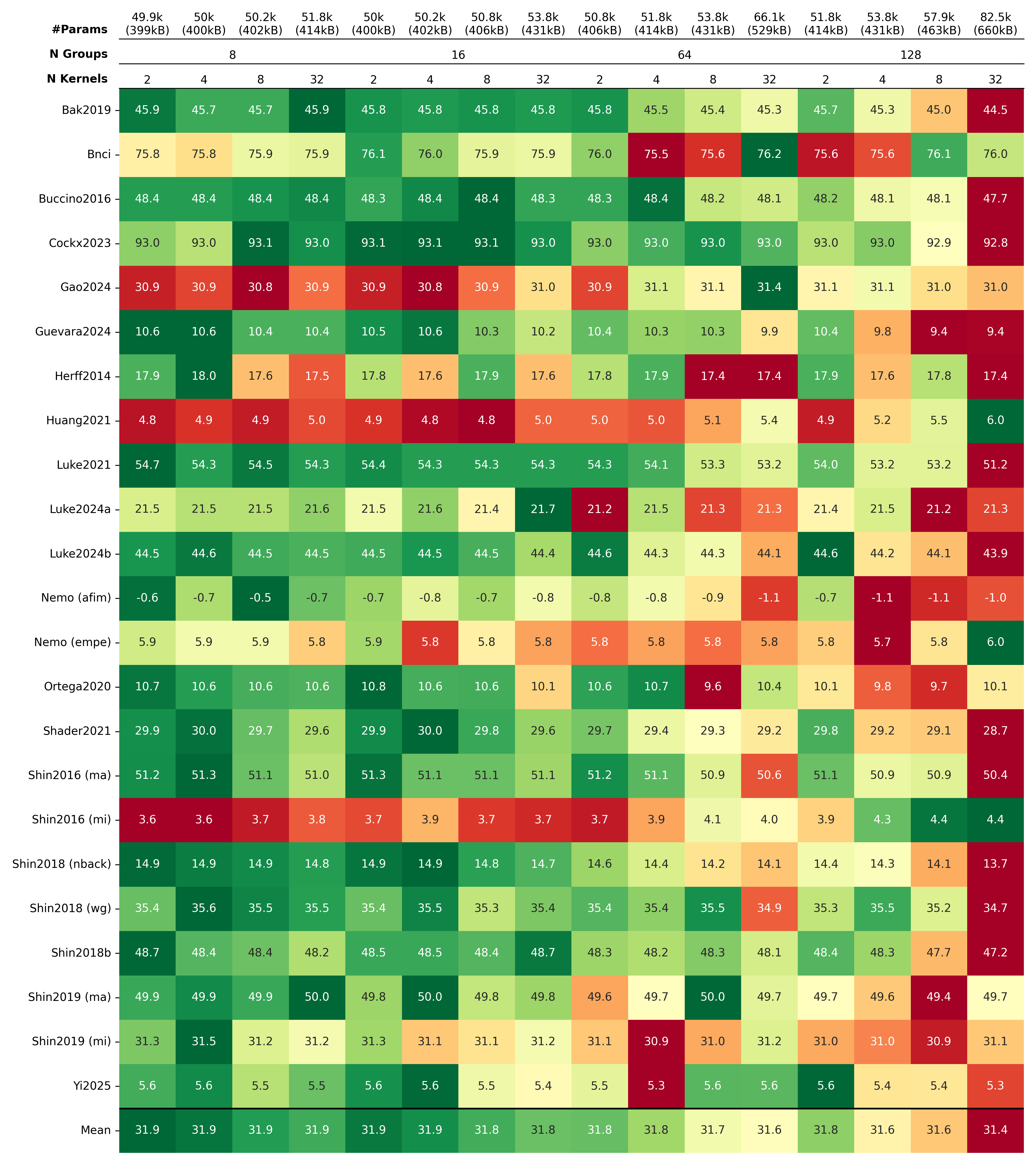}
\caption{Within-subject validation $\kappa$ across the full hyperparameter grid for MRH.}
    \label{fig:mrh_paramsearch_within}
\end{figure}
\begin{table}[htbp]
\centering
\caption{Coefficient summary for the MRH hyperparameter analysis (within-subject).}
\label{tab:hparam_lm_within_aeon_multirockethydra}
\small
\begin{tabular}{l c}
\toprule
Term & Coef. \\
\midrule
Intercept (N Groups = 8, N Kernels = 2) & 32.0 \\
N Groups = 16 & 0.0 \\
N Groups = 64 & -0.2 \\
N Groups = 128 & -0.3 \\
N Kernels = 4 & -0.1 \\
N Kernels = 8 & -0.1 \\
N Kernels = 32 & -0.2 \\
\bottomrule
\end{tabular}
\end{table}

\clearpage
\begin{figure}[H]
\centering
    \includegraphics[width=0.95\textwidth]{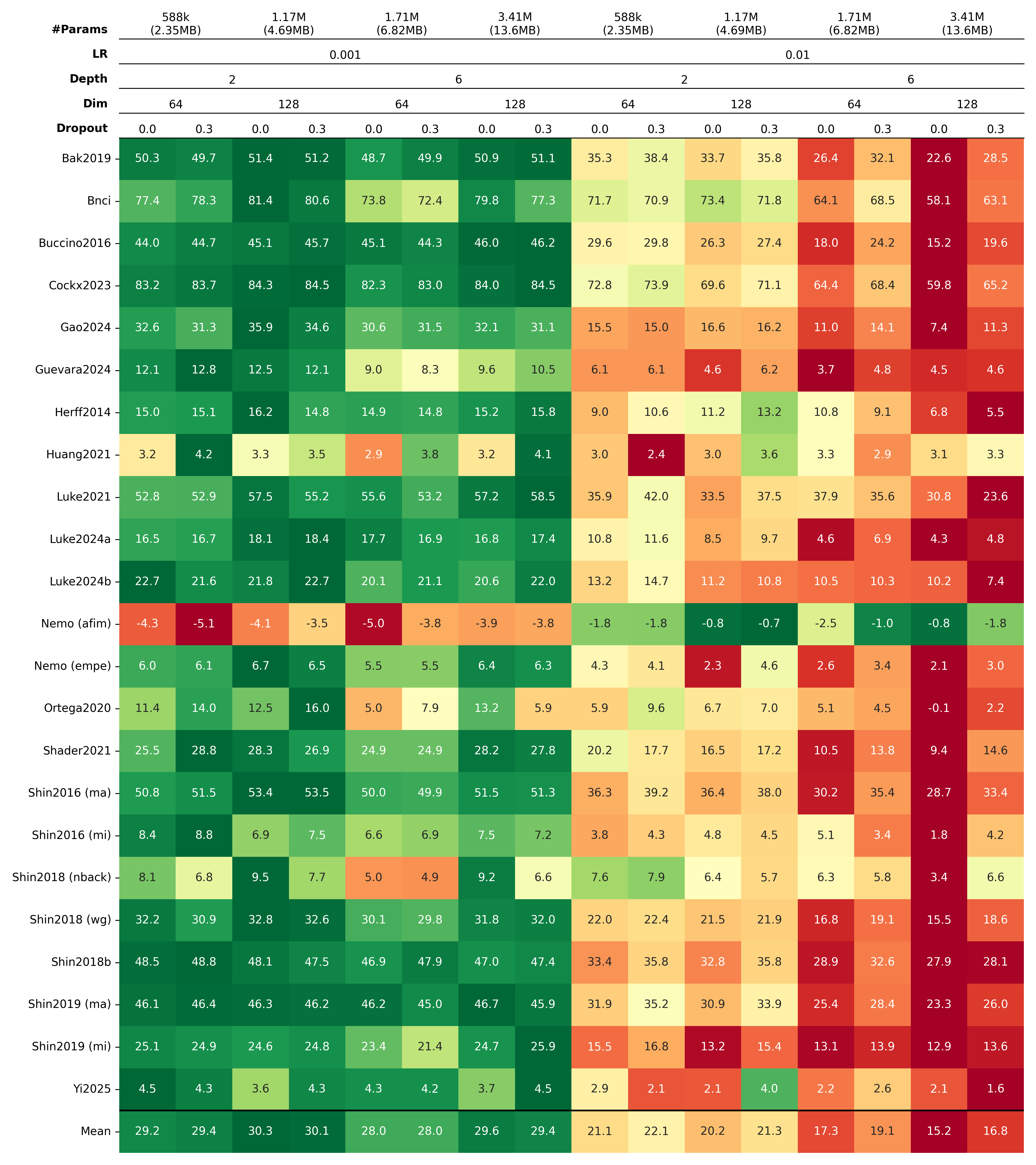}
\caption{Within-subject validation $\kappa$ across the full hyperparameter grid for fNIRS-T.}
    \label{fig:fnirst_paramsearch_within}
\end{figure}
\begin{table}[htbp]
\centering
\caption{Coefficient summary for the fNIRS-T hyperparameter analysis (within-subject).}
\label{tab:hparam_lm_within_fnirst}
\small
\begin{tabular}{l c}
\toprule
Term & Coef. \\
\midrule
Intercept (Depth = 2, Dim = 64, Dropout = 0.0, Learning Rate = 0.001) & 30.3 \\
Depth = 6 & -2.6 \\
Dim = 128 & -0.2 \\
Dropout = 0.3 & 0.7 \\
Learning Rate = 0.01 & -10.1 \\
\bottomrule
\end{tabular}
\end{table}

\clearpage
\begin{figure}[H]
\centering
    \includegraphics[width=0.95\textwidth]{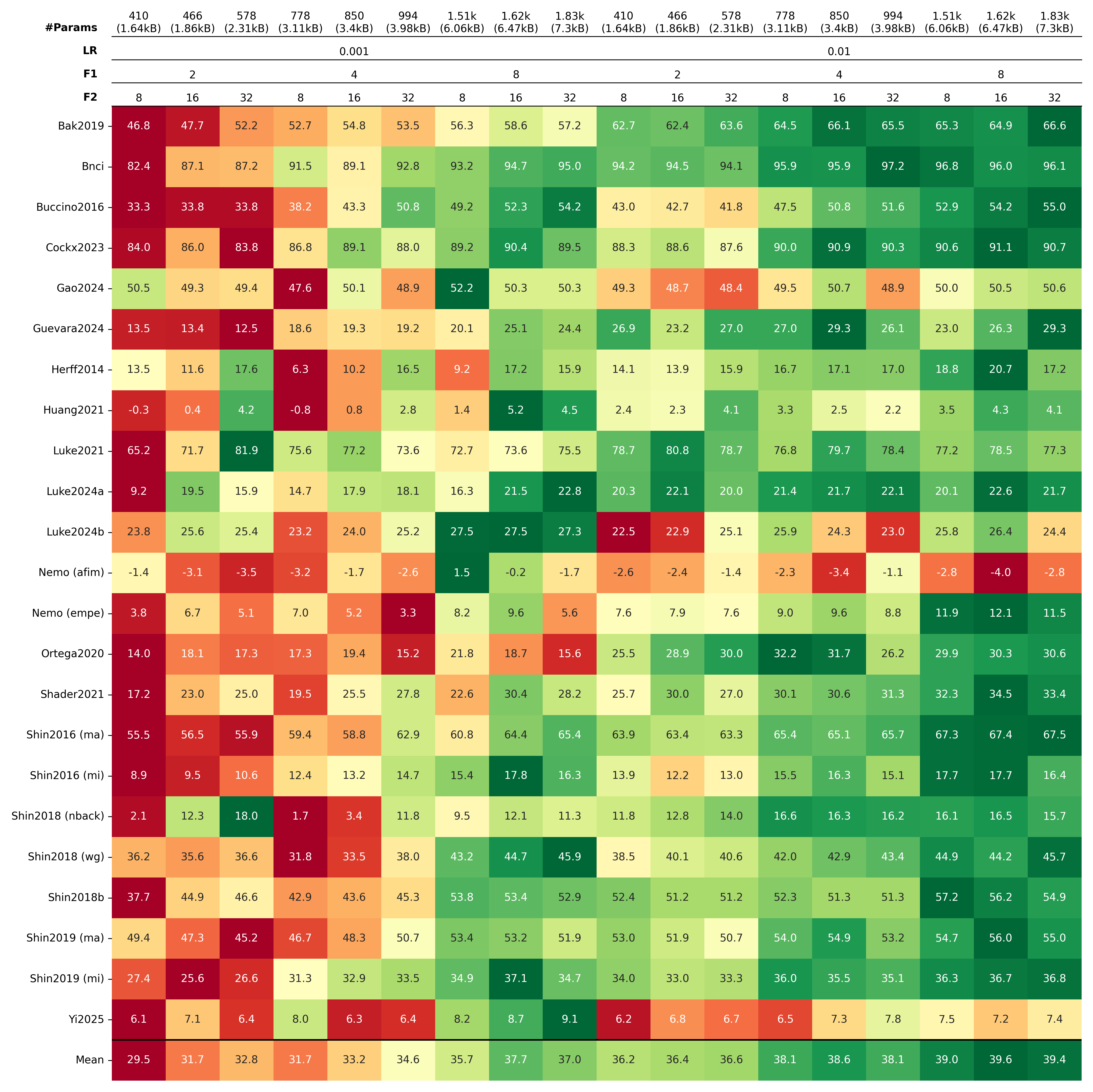}
\caption{Within-subject validation $\kappa$ across the full hyperparameter grid for fNIRSNet23.}
    \label{fig:fnirsnet23_paramsearch_within}
\end{figure}
\begin{table}[htbp]
\centering
\caption{Coefficient summary for the fNIRSNet23 hyperparameter analysis (within-subject).}
\label{tab:hparam_lm_within_fnirsnet23}
\small
\begin{tabular}{l c}
\toprule
Term & Coef. \\
\midrule
Intercept (F1 = 2, F2 = 8, Learning Rate = 0.001) & 30.9 \\
F1 = 4 & 1.8 \\
F1 = 8 & 4.2 \\
F2 = 16 & 1.2 \\
F2 = 32 & 1.4 \\
Learning Rate = 0.01 & 4.2 \\
\bottomrule
\end{tabular}
\end{table}

\clearpage
\begin{figure}[H]
\centering
    \includegraphics[width=0.95\textwidth]{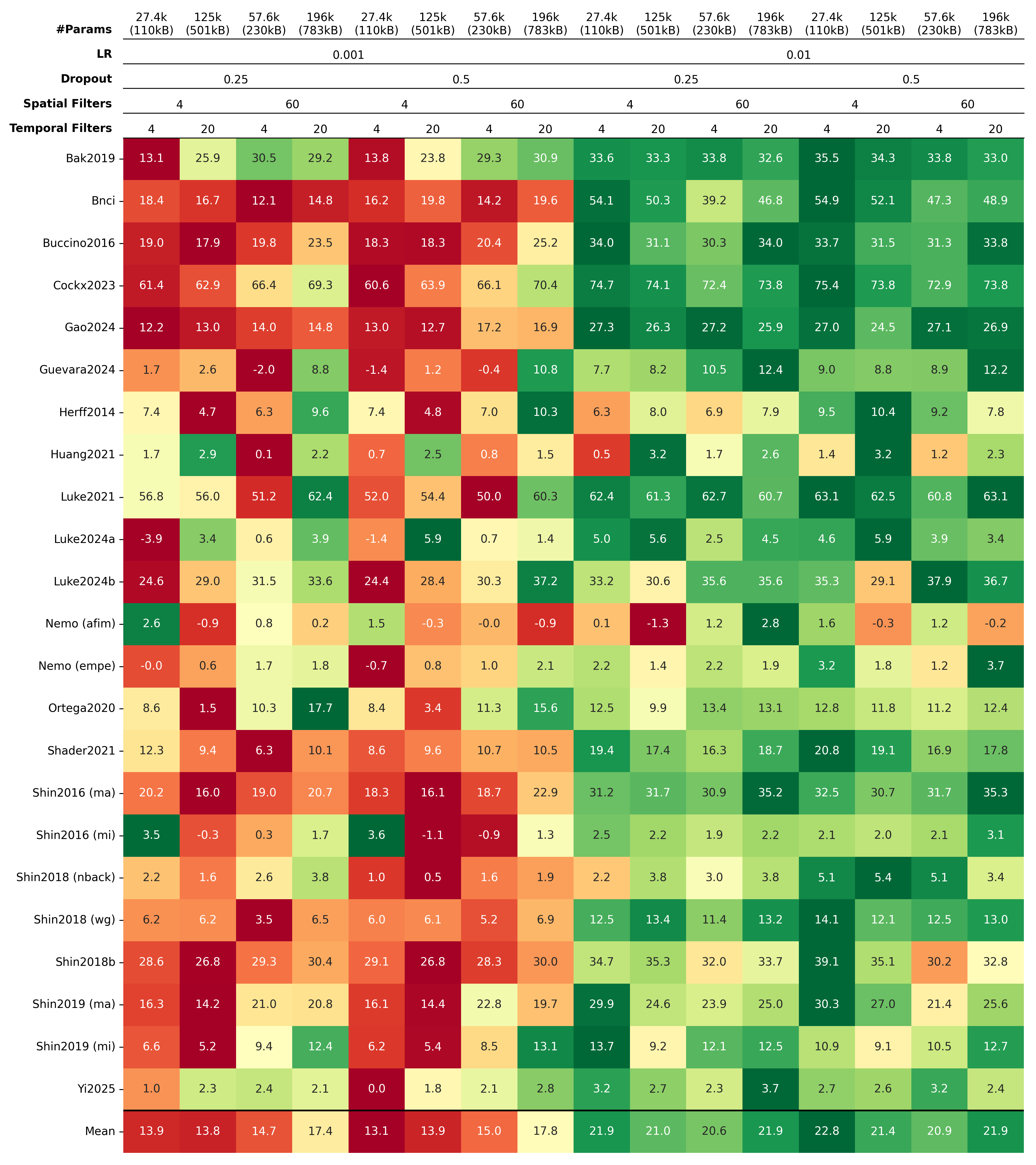}
\caption{Within-subject validation $\kappa$ across the full hyperparameter grid for fNIRSNet24.}
    \label{fig:fnirsnet24_paramsearch_within}
\end{figure}
\begin{table}[htbp]
\centering
\caption{Coefficient summary for the fNIRSNet24 hyperparameter analysis (within-subject).}
\label{tab:hparam_lm_within_fnirsnet24}
\small
\begin{tabular}{l c}
\toprule
Term & Coef. \\
\midrule
Intercept (Learning Rate = 0.001, Dropout = 0.25, Spatial Filters = 4, Temporal Filters = 4) & 13.9 \\
Learning Rate = 0.01 & 6.6 \\
Dropout = 0.5 & 0.2 \\
Spatial Filters = 60 & 1.0 \\
Temporal Filters = 20 & 0.8 \\
\bottomrule
\end{tabular}
\end{table}

\clearpage
\section*{NeurIPS Paper Checklist}

\begin{enumerate}

\item {\bf Claims}
    \item[] Question: Do the main claims made in the abstract and introduction accurately reflect the paper's contributions and scope?
    \item[] Answer: \answerYes{}
    \item[] Justification: The abstract and introduction state the main contributions clearly, including the benchmark of 23 classification tasks from 19 public data collections, the standardized pipeline, and the baseline evaluation. These claims match the scope of the paper in Sections~1--6.

\item {\bf Limitations}
    \item[] Question: Does the paper discuss the limitations of the work performed by the authors?
    \item[] Answer: \answerYes{}
    \item[] Justification: The Discussion section explicitly discusses limitations, including dataset coverage and scale, fixed benchmark defaults, restricted baseline selection, and the gap between benchmark settings and broader practical use. These points are discussed in Section~6.

\item {\bf Theory assumptions and proofs}
    \item[] Question: For each theoretical result, does the paper provide the full set of assumptions and a complete (and correct) proof?
    \item[] Answer: \answerNA{}
    \item[] Justification: The paper is empirical and does not present original theoretical results that require formal assumptions and proofs.

     \item {\bf Experimental result reproducibility}
     \item[] Question: Does the paper fully disclose all the information needed to reproduce the main experimental results of the paper to the extent that it affects the main claims and/or conclusions of the paper (regardless of whether the code and data are provided or not)?
     \item[] Answer: \answerYes{}
     \item[] Justification: The paper describes the preprocessing pipeline, model families, hyperparameter grids, evaluation protocol, repeated cross-validation, statistical testing, software stack, and compute environment in the Methods section and appendices. This provides the information needed to reproduce the reported benchmark setup and main results.

\item {\bf Open access to data and code}
    \item[] Question: Does the paper provide open access to the data and code, with sufficient instructions to faithfully reproduce the main experimental results, as described in supplemental material?
    \item[] Answer: \answerYes{}
    \item[] Justification: The benchmark code is openly released, and the paper provides the information needed to access the datasets used in the benchmark together with the pipeline and evaluation setup needed to reproduce the main results. Reproduction-relevant details are provided in the paper and appendices.

\item {\bf Experimental setting/details}
    \item[] Question: Does the paper specify all the training and test details (e.g., data splits, hyperparameters, how they were chosen, type of optimizer) necessary to understand the results?
    \item[] Answer: \answerYes{}
    \item[] Justification: The Methods section specifies preprocessing defaults, hyperparameter grids, training settings, optimization details, batch sizes, and within-subject and cross-subject evaluation protocols. Additional implementation and preprocessing details are provided in the appendices.

\item {\bf Experiment statistical significance}
    \item[] Question: Does the paper report error bars suitably and correctly defined or other appropriate information about the statistical significance of the experiments?
    \item[] Answer: \answerYes{}
    \item[] Justification: The result tables report mean accuracy with the standard deviation across folds (cross-subject) or subjects (within-subject), and Sections~4--5 report Friedman and Nemenyi tests for model comparisons.

\item {\bf Experiments compute resources}
    \item[] Question: For each experiment, does the paper provide sufficient information on the computer resources (type of compute workers, memory, time of execution) needed to reproduce the experiments?
    \item[] Answer: \answerYes{}
    \item[] Justification: The Compute Resources appendix reports the CPU and GPU node types, memory, total CPU core-hours and GPU-hours, and the model-wise compute breakdown. This information is provided in the Compute Resources appendix.

\item {\bf Code of ethics}
    \item[] Question: Does the research conducted in the paper conform, in every respect, with the NeurIPS Code of Ethics \url{https://neurips.cc/public/EthicsGuidelines}?
    \item[] Answer: \answerYes{}
    \item[] Justification: The work benchmarks previously published public datasets and does not introduce procedures that appear to conflict with the NeurIPS Code of Ethics. No deviations from the code are described in the paper.

\item {\bf Broader impacts}
    \item[] Question: Does the paper discuss both potential positive societal impacts and negative societal impacts of the work performed?
    \item[] Answer: \answerYes{}
    \item[] Justification: We mention the positive benefits of improving reproducibility and comparability in fNIRS classification research. We do not think there are substantial negative societal impacts to discuss for a benchmark paper of this kind.

\item {\bf Safeguards}
    \item[] Question: Does the paper describe safeguards that have been put in place for responsible release of data or models that have a high risk for misuse (e.g., pre-trained language models, image generators, or scraped datasets)?
    \item[] Answer: \answerNA{}
    \item[] Justification: The paper presents a benchmark built from public fNIRS datasets and baseline classifiers rather than high-risk generative models or scraped assets that would require special release safeguards.

\item {\bf Licenses for existing assets}
    \item[] Question: Are the creators or original owners of assets (e.g., code, data, models), used in the paper, properly credited and are the license and terms of use explicitly mentioned and properly respected?
    \item[] Answer: \answerYes{}
    \item[] Justification: The paper credits the source datasets and documents their licenses and access conditions in the dataset license appendix. The released fNIRSAtlas code is provided under the MIT license.

\item {\bf New assets}
    \item[] Question: Are new assets introduced in the paper well documented and is the documentation provided alongside the assets?
    \item[] Answer: \answerYes{}
    \item[] Justification: The paper releases the fNIRSAtlas package as a new code asset, and the package is documented with the information needed to use the benchmark and reproduce the main evaluation setup.

\item {\bf Crowdsourcing and research with human subjects}
    \item[] Question: For crowdsourcing experiments and research with human subjects, does the paper include the full text of instructions given to participants and screenshots, if applicable, as well as details about compensation (if any)? 
    \item[] Answer: \answerNA{}
    \item[] Justification: The paper does not report new crowdsourcing experiments or new data collection with human subjects by the authors.

\item {\bf Institutional review board (IRB) approvals or equivalent for research with human subjects}
    \item[] Question: Does the paper describe potential risks incurred by study participants, whether such risks were disclosed to the subjects, and whether Institutional Review Board (IRB) approvals (or an equivalent approval/review based on the requirements of your country or institution) were obtained?
    \item[] Answer: \answerNA{}
    \item[] Justification: The work uses existing public datasets and does not describe new human-subjects research conducted by the authors.

\item {\bf Declaration of LLM usage}
    \item[] Question: Does the paper describe the usage of LLMs if it is an important, original, or non-standard component of the core methods in this research? Note that if the LLM is used only for writing, editing, or formatting purposes and does \emph{not} impact the core methodology, scientific rigor, or originality of the research, declaration is not required.
    \item[] Answer: \answerNA{}
    \item[] Justification: The core methods and experiments do not use LLMs as an important, original, or non-standard methodological component.

\end{enumerate}

\end{document}